\documentclass[oneside,table,dvipsnames]{arxiv}
\usepackage{tabularx}
\usepackage{float}
\usepackage{bm}
\usepackage{rotating}   
\usepackage{graphicx}
\usepackage{epstopdf}
\usepackage{booktabs}
\usepackage{enumitem}
\usepackage{verbatim}
\usepackage{amsmath}
\usepackage{dblfloatfix}
\usepackage{wrapfig}

\newcommand{\cmark}{\textcolor{teal}{\ding{51}}}   
\newcommand{\xmark}{\textcolor{red!70}{\ding{55}}} 
\usepackage{pifont}       

\makeatletter
\def\els@aparagraph[#1]#2{\elsparagraph[#1]{#2\@addpunct{.}}}
\def\els@bparagraph#1{\elsparagraph*{#1\@addpunct{.}}}
\makeatother

\usepackage{tikz,lipsum,lmodern}
\usepackage{array}
\usepackage[parfill]{parskip}
\usepackage{multirow}
\usepackage{multicol}

\newcommand{\fakesection}[1]{%
  \par\refstepcounter{section}
  \sectionmark{#1}
  \addcontentsline{toc}{section}{\protect\numberline{\thesection}#1}
}
\usepackage{lineno}
\usepackage{orcidlink}

\usepackage[most]{tcolorbox}
\newcounter{reviewboxcounter}
\newtcolorbox[auto counter,
              number freestyle={\noexpand\arabic{reviewboxcounter}}]{reviewbox}[1][]{
  enhanced,
  colback=blue!5,
  colframe=blue!40!black,
  arc=2mm,
  boxrule=1pt,
  title={Box \thetcbcounter: #1},  
  fonttitle=\bfseries,
  coltitle=black!70,
  attach boxed title to top left={xshift=0.5cm, yshift=-2mm},
  boxed title style={
    colback=blue!40!white,
    colframe=blue!40!black,
    sharp corners
  },
  before skip=10pt,
  after skip=10pt
}
\DeclareFloatingEnvironment[name={Video},fileext=lsf,listname={List of Videos}]{video}

\begin{document}

\title{\vspace{-3ex} 
\textbf{Arnold: A multi-task, multi-embodiment muscle transformer policy}} 
\leadauthor{An, Chiappa \& Simos et al.}
\shorttitle{Arnold}

\author{\large 
  Boshi An$^{1,2}$, Alberto Silvio Chiappa$^{1,2}$, Merkourios Simos$^{1,2}$,\\ Chengkun Li$^{1}$, Alexander Mathis\Envelope$^,$}


\affil[1]{Brain Mind Institute, School of Life Sciences, École Polytechnique Fédérale de Lausanne (EPFL), Lausanne, Switzerland}
\affil[2]{These authors contributed equally.}
\affil[\Envelope]{Correspondence: alexander.mathis@epfl.ch}

\onecolumn
\maketitle

\section*{Abstract}

\textbf{Controlling high-dimensional and nonlinear musculoskeletal models of the human body is a foundational scientific challenge.
Recent machine learning breakthroughs have heralded in-silico policies that master individual skills like reaching, object manipulation and locomotion in musculoskeletal systems with many degrees of freedom.
However, these agents are merely “specialists”, achieving high performance for a single skill.
In this work, we develop Arnold, a transformer-based musculoskeletal control policy that masters multiple tasks and embodiments.
Arnold combines behavior cloning and reinforcement learning to address 14 challenging control tasks spanning dexterous object manipulation, reaching, and locomotion, matching or exceeding the performance of single-task specialist policies.
A key innovation is Arnold’s sensorimotor vocabulary, a compositional representation of the semantics of heterogeneous sensory modalities, objectives, and actuators.
Arnold leverages this vocabulary via a transformer architecture to deal with the variable observation and action spaces across tasks.
This framework supports efficient multi-task, multi-embodiment learning and facilitates rapid adaptation to novel tasks, while encouraging universal motor strategies such as action and kinematic smoothness.
Finally, causal probing of the motor output reveals that low-dimensional muscle synergies remain largely task-specific and that variance-based analyses systematically underestimate functional control dimensionality, consistent with biological observations on the limited transferability of such synergies. Code and data are available here: \href{https://github.com/amathislab/arnold}{https://github.com/amathislab/arnold}}



\section{Introduction}

Dexterous and adaptive motor control in humans and other animals serves as a central inspiration for robotics and artificial intelligence.
While most robotics research has traditionally focused on motor-actuated systems~\citep{tassa2018deepmind,makoviychuk2021isaac}, muscle-like control is gaining momentum, supported by emerging hardware platforms~\citep{jacobsen1986design,kumar2013fast,yasa2023overview,shaw2023leap,allegro,christoph2025orca}.
These muscle-based systems offer numerous advantages, including flexible weight distribution, efficient energy storage in tendons, and valuable opportunities for cross-pollination with computational neuroscience~\citep{zhang2019robotic,hansen2022temporal,wochner2023learning,vargas2024taskdriven,chiappa2024acquiring,perez2025deep}. 
In parallel, biomechanics simulators have also flourished~\citep{delp2007opensim, seth2011opensim, geijtenbeek2019scone, rasmussen2002anybody, caggiano2022myosuite}, enabling in-silico experiments across multiple platforms; for a systematic review of predictive musculoskeletal simulation spanning these ecosystems see~\citeauthor{denayer2025prisma} ~\citep{denayer2025prisma}.

Developing musculoskeletal control policies poses unique challenges due to the large number of sensory inputs and actuators~\citep{wolpert2011principles,vargas2024taskdriven}, their over-actuated nature~\citep{bernstein1967coordination}, as well as nonlinear muscle dynamics~\citep{hill1938heat,delp2007opensim,caggiano2022myosuite}.
These technical complications have motivated advances in reinforcement learning (RL), exploration, curriculum learning (CL) and imitation learning (IL)~\citep{schumacher2022dep,la2021ostrichrl,song2021deep,berg2023sar,chiappa2023latent,caggiano2023myochallenge,feng2023musclevae,wochner2023learning,he2024dynsyn,chiappa2024acquiring,schumacher2025natural,simos2026reinforcement}.
By combining these techniques, previous works have succeeded in controlling biologically realistic models of the human body to reach strong performance for complex tasks such as object manipulation and locomotion~\citep{caggiano2023myochallenge,caggiano2024myochallenge,caggiano2024myochallengea, wang2026myochallenge}. 
Unfortunately, these results are limited to one specific problem at a time, and different embodiments or tasks require learning new policies. 
This approach complicates drawing systems-level conclusions about motor control~\citep{bernstein1967coordination,schmidt2018motor}, not least because specialist policies reproduce only a limited repertoire of motor skills.

\begin{figure}[t]
    \centering
    \includegraphics[width=\linewidth]{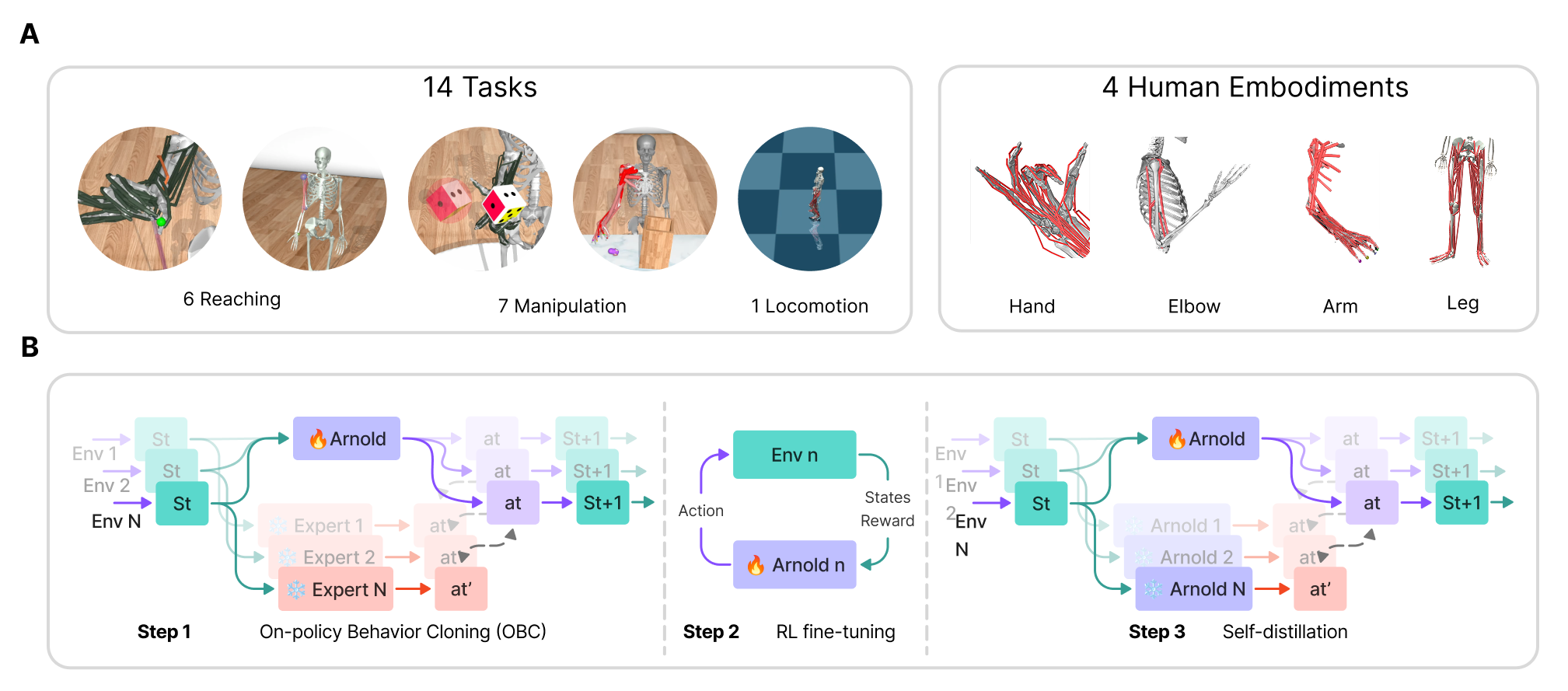}
    \caption{\textbf{Task and training overview for Arnold.} \textbf{A:} Arnold is a transformer-based musculoskeletal policy that controls multiple human embodiments and solves 14 different motor control tasks. \textbf{B:} We train Arnold in three steps. \textbf{Step 1}: Arnold learns to imitate multiple experts in parallel using on-policy behavior cloning (OBC), which trains the shared policy on its own state distribution labeled by expert actions and yields a unified policy that can operate across all tasks. \textbf{Step 2}: Different versions of Arnold are specialized to individual tasks using PPO. These specialists might outperform the original experts. 
    \textbf{Step 3}: The specialized copies of Arnold from Step 2 become the new "experts". The original Arnold model from Step 1 is then updated using OBC to imitate these improved experts. 
    Overall, steps 2 and 3 form a refinement process where Arnold learns from improved versions of itself.
    }
    \label{fig:overview}
\end{figure}

Constructing a musculoskeletal policy that flexibly handles different tasks and embodiments poses distinct but interconnected challenges that do not arise when training specialist single-task policies.
First, the number of muscles and sensory channels varies substantially across embodiments: the observation and action spaces of a finger-control task are fundamentally different from those of locomotion, making fixed-dimensionality architectures incompatible with multi-task deployment.
Second, even when the same muscle appears across tasks, its functional role changes with the behavioral context; a policy must therefore identify not only which effectors it controls but also how each contributes to the current task objective.
Third, training a single policy to simultaneously imitate multiple expert policies amplifies the distributional mismatch inherent in standard behavior cloning: the state distributions encountered by a multi-task learner during deployment diverge from those of any individual expert, and errors in one task's distribution can destabilize learning across all tasks.

In this work, we addressed these challenges by developing \textbf{Arnold}, \textbf{a} t\textbf{r}a\textbf{n}sf\textbf{o}rmer-based mu\textbf{l}ti-task, multi-embo\textbf{d}iment policy for human musculoskeletal control (Figure~\ref{fig:overview}).
Arnold addresses variable observation and action spaces by formulating motor control as a sequence-to-sequence problem, which is solved by using a transformer encoder-decoder architecture.
Functional-role disambiguation is achieved through a \emph{sensorimotor vocabulary}, a set of compositional token embeddings that describe each sensory stream and actuator not only by its raw value but also by its functional role in the sensory-to-motor transformation~\citep{cisek2021evolution,chiappa2022dmap}. This design enables a compact set of representations shared across tasks and embodiments.
The distributional mismatch is addressed with on-policy behavior cloning (OBC)~\citep{ross2011reduction, lee2020learning,lee2024learning,li2026omniclone, chiappa2024autobidding}, which trains the policy exclusively on its own rollouts while the respective experts provide target actions; at the end of this stage, Arnold already handles all 14 tasks successfully under a single unified policy.
A second stage of task-specific PPO fine-tuning followed by self-distillation further refines this unified policy beyond the performance of the individual experts, without requiring any architectural modification.

Arnold enables efficient transfer to tasks previously unseen by the policy. A network pretrained on a subset of tasks learned novel tasks faster than a randomly initialized network, indicating that the training framework encourages the emergence of reusable motor control strategies.
We found that muscle activity smoothness, kinematic smoothness, and correlation of muscle activity to human EMG data improve as a consequence of multi-task training, while self-distillation also improves EMG correlation. Interestingly, causal probing at the level of motor output based on control-subspace inactivation (CSI)~\citep{chiappa2024acquiring} revealed both a lack of muscle synergy transfer across tasks and a systematic underestimation of control space dimensionality by variance-based synergy analyses, indicating that universal motor strategies may exist further upstream in the sensorimotor pathway. We anticipate that musculoskeletal policies of this kind will serve as increasingly powerful platforms for probing the computational principles of motor control across the full breadth of human motor capabilities, opening new avenues for both motor neuroscience and embodied artificial intelligence.

\section{Results}
\label{sec:experiments_arnold}

\subsection*{Problem formulation}

The core objective of this study was to design a controller that solves multiple Partially Observable Markov Decision Processes (POMDP) at once.
Each POMDP represents a distinct motor control task, which in turn is characterized by three components: a musculoskeletal model, a task objective, and (optional) object interactions.

Specifically, we tackled several complex, biologically-realistic motor control tasks from the MyoSuite library~\citep{caggiano2022myosuite}, ranging from single finger control and object manipulation tasks to locomotion (Figure~\ref{fig:overview}A; for task descriptions see Figure~\ref{fig:fulltask_description}).
The 14 tasks span four distinct musculoskeletal models, creating a comprehensive testbed that reflects some of the versatility of human motor control.
At the simplest level, we considered precise joint control for different fingers and the elbow, escalating to the coordination required for object manipulation (e.g. \emph{Pen reorient}) using 39 hand muscles.
The complexity of musculoskeletal control is exemplified in two coordination tasks: in \emph{Object relocation}, agents grasp and relocate differently-shaped objects through 63-muscle control, while in \emph{Walk to point}, the goal is to walk to arbitrary targets using an 80-muscle lower body.
All tasks are described in detail in the Methods section.

Although so far no studies have considered this set of tasks together, each individual task has been studied and, at least partially, solved through specialist single-task policies (hereby denoted as `experts').
These expert policies are based on previous work and constitute formidable baselines~\citep[][see Methods]{caggiano2023myochallenge,caggiano2024myochallenge,chiappa2024acquiring,simos2026reinforcement,wang2026myochallenge}. Each task defines a reward function, which is a measure of a policy's performance. Task-specific reward structures, however, make them poorly interpretable metrics, as an upper bound is not obviously defined. We therefore employ a unified, more interpretable performance metric: the \emph{relative performance}. This metric is defined as the policy's solved fraction, i.e., the fraction of time steps in which the target configuration is achieved, divided by the solved fraction of the corresponding single-task expert. 
Values above 100\% therefore indicate performance above the expert baseline under the same task-specific solved-fraction metric, and should not be interpreted as an absolute probability of task success.

At each simulation step, the policy observes the proprioceptive state of the body in the form of muscle (length, velocity, force and activation) and joint (angular position and velocity) information, the (position and orientation) configuration of objects if applicable, and the task objective. This format of proprioceptive feedback deviates from the original formulation in MyoSuite and MyoChallenges that considers joint states~\citep{caggiano2022myosuite,caggiano2023myochallenge,caggiano2024myochallenge,wang2026myochallenge}, but is more biologically plausible~\citep{proske2012proprioceptive,vargas2024taskdriven}.
In all considered tasks, the objective is defined by one or more target locations or orientations, which the agent needs to reach either with body parts or with objects. The agent can interact with the environment by selecting the activation of each muscle, causing a contraction of variable intensity. The biomechanical simulator~\citep[MuJoCo][]{todorov2012mujoco} transforms activations into physical forces through a Hill-type muscle model~\citep{winters1990hillbased}. Given these different tasks, the policy network needs to handle observation and action spaces of variable size and content.

\begin{figure}[ht!]
    \centering
    \includegraphics[width=\linewidth]{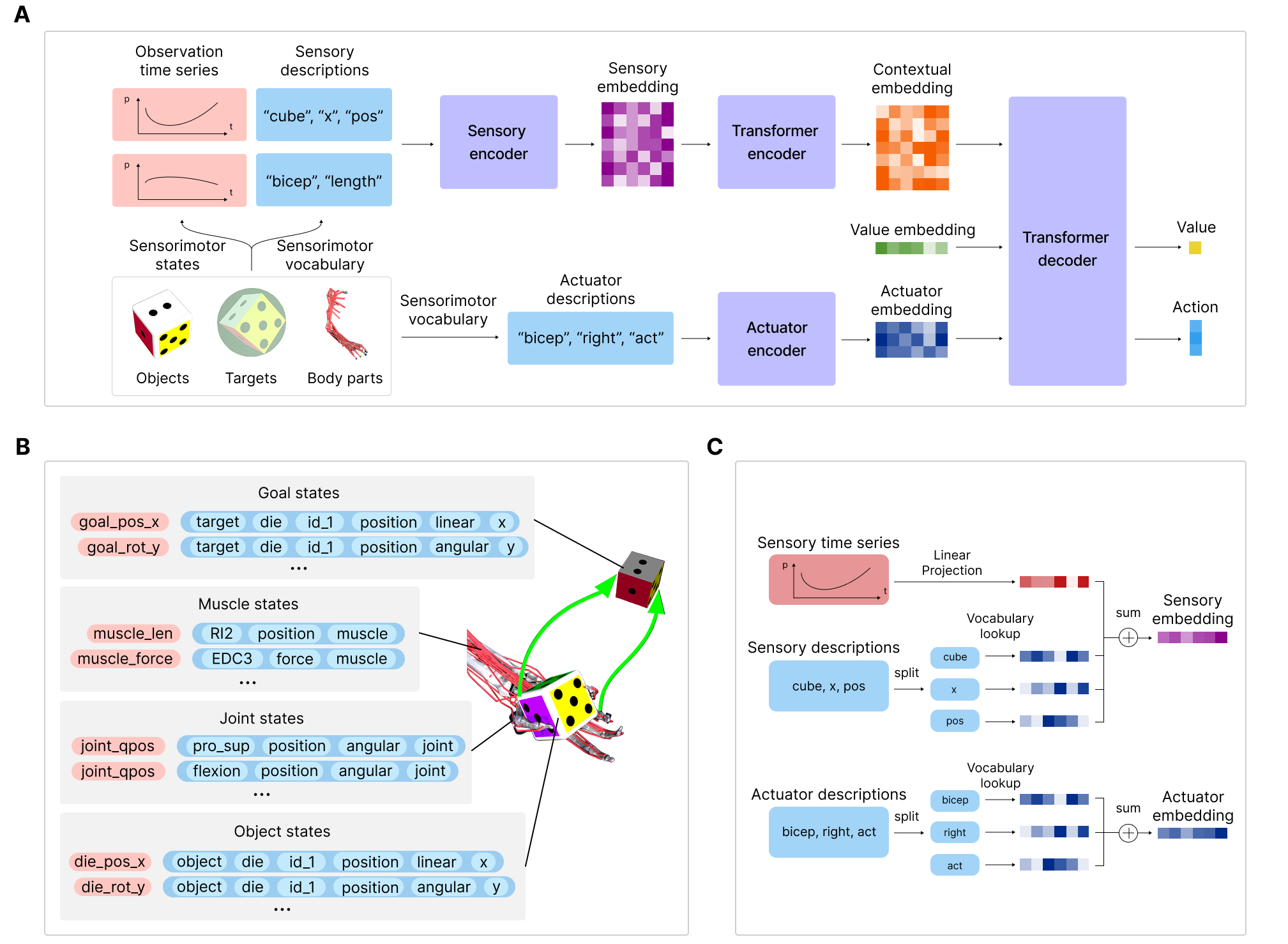}
    \caption{\textbf{Architecture of the muscle transformer policy.} \textbf{A.} \emph{Sensory embeddings} (purple) are computed from raw observations by the \emph{sensory encoder} before they are processed by the transformer encoder-decoder network. \emph{Actuator embeddings} (dark blue) are computed by the \emph{actuator encoder} based on the current task and instruct the model which embodiment should be controlled. Cross-attention is performed in the decoder network to generate muscle actuation and value estimation based on the \emph{contextual embeddings} (orange). \textbf{B.} Illustration of inputs for one example task (Die reorient) in which a model of the human hand (39 muscles) needs to be controlled in order to orient a die towards a goal orientation that changes in every episode. Arnold receives the goal states as well as the muscle, die and joint states. These raw observations of the task are organized semantically with a compositional \emph{sensorimotor vocabulary}. \textbf{C.} In the \emph{sensory encoder}, sensorimotor descriptions are split into vocabulary entries and mapped onto embeddings through a \emph{vocabulary lookup} table. The embeddings from all entries are summed and added to the projected sensorimotor time series to obtain the \emph{sensory embedding}. For actuator descriptions, the same lookup mechanism is used to obtain the \emph{actuator embedding}.}
    \label{fig:architecture}
\end{figure}

\subsection*{A transformer framework with compositional sensorimotor vocabulary addresses diverse state and action spaces}

Multi-task learning is characterized by two fundamental challenges. First, the size of the observation and action spaces may change across tasks.
Second, even the same observation and action dimensions may carry significantly different semantics based on the different task structures.
However, in standard model-free multi-task RL, the policy network is implemented as a multi-layer perceptron (MLP) where both the observation and the action space are fixed.
A common remedy when observation and action spaces differ across tasks is to define a global space as the union of the per-task spaces, zero-padding observations and masking out action dimensions irrelevant to the current task, with a task index appended to disambiguate context~\citep{yu2021metaworld, terry2020revisiting}.
However, this approach scales poorly as the number of heterogeneous tasks grows.
Indeed, we found that policies based on MLP networks trained on all 14 tasks with MT-PPO~\citep{yu2021metaworld,schulman2017proximal} or MT-SAC~\citep{haarnoja2018soft} showed poor performance on most tasks after 50M (20M for MT-SAC) environment interactions (Table~\ref{tab:ablations}, Table~\ref{tab:ablations_per_task}, Figure~\ref{fig:mt_ppo_curves}).

We addressed these two challenges by formulating the motor control problem as sequence-to-sequence modeling: the translation of sensory input into motor commands (Figure~\ref{fig:architecture}A).
Each sensory modality (e.g. muscle length of a specific muscle, object position, and target orientation) from the simulation, as well as the information about controllable muscles, was associated with a descriptive sequence of tokens from a \emph{sensorimotor vocabulary} (Figure~\ref{fig:architecture}B).

The \emph{sensorimotor vocabulary} consists of a set of compositional tokens that can be arranged to describe a wide variety of sensory inputs, target definitions, and motor outputs.
In total, 214 elements are sufficient to characterize all inputs and outputs of the 14 MyoSuite tasks.
These elements are combined to describe complex sensory input by summing them (Figure~\ref{fig:architecture}C).
For example, the embedding of the length of the right soleus muscle is the sum of the "right", "soleus", "muscle", and "length" embeddings (see Methods).
This compositionality of the \emph{sensorimotor vocabulary} allowed us to reduce its size by taking advantage of physiological symmetries and shared semantics (e.g., lateral body symmetry and shared muscle features), promoting reusable representations for the same sensorimotor element across tasks.

To generate muscle activations, the information of the muscle actuators is first embedded into \emph{actuator embeddings} through an \emph{actuator encoder} (Figure~\ref{fig:architecture}A). The \emph{transformer decoder} receives as input one \emph{actuator embedding} per muscle and outputs one vector for each of them through cross-attention with the encoder outputs. Finally, the output of the decoder is linearly mapped into muscle activations. When deploying the muscle transformer with an actor-critic algorithm like PPO~\citep{schulman2017proximal}, we use a separate transformer decoder for predicting the value function, where a \emph{value embedding} is passed to the decoder inputs and maps to a scalar value estimate (see Methods for details on the architecture).

This network architecture can accept any number of input sensory modalities and output control signals for any number of actuators. This flexibility makes the muscle transformer suitable for learning tasks characterized by different embodiments, external objects, and reward structures (Figure~\ref{fig:overview}A and Figure~\ref{fig:fulltask_description}).

Similar to the simple MLP-based method described above, we used a parallel experience collection setup, where instances of each task environment run simultaneously and populate a shared rollout buffer (Figure~\ref{fig:overview}B: Step 1). 
We trained five randomly-initialized instances of the muscle transformer network for 50M environment interactions using PPO~(Figures~\ref{fig:ablations}A~and~\ref{fig:muscle_transfomer_ppo_curves}).
Overall, our proposed architecture led to an increase in performance in most of the 14 tasks~(PPO + T + SV, Table~\ref{tab:ablations}, Table~\ref{tab:ablations_per_task}, Figure~\ref{fig:ablations}, SV = sensorimotor vocabulary).
For these tasks, we also identified that reward standardization and per-component observation standardization played a critical role; ablating either of those processing steps led to a substantial performance drop~(Figure~\ref{fig:ablations}B).
Notably, removing the sensorimotor vocabulary while retaining the transformer architecture led to performance collapse~(PPO + T, Table~\ref{tab:ablations}, Table~\ref{tab:ablations_per_task}).

Nevertheless, the overall performance of the multi-task policy was below expert-level, with the largest differences found in the most challenging tasks, such as \emph{Walk to point} and \emph{Object relocation}.

\subsection*{On-policy behavior cloning distills single-task specialists into a single policy}

The unfavorable comparison of multi-task reinforcement learning against single-task expert policies was largely expected; the more challenging environments had required advanced training strategies beyond standard RL (specifically curriculum learning, mixture of experts, and pretraining with motion imitation~\citep{chiappa2024acquiring,caggiano2024myochallenge,simos2026reinforcement}) to successfully train the original expert policies (see Methods).
Therefore, instead of tackling the challenge of multi-task learning from scratch, we turned to imitation learning, using the specialist expert policies as teachers to train the muscle transformer model.

\begin{figure}[ht!]
    \centering
    \includegraphics[width=0.95\linewidth]{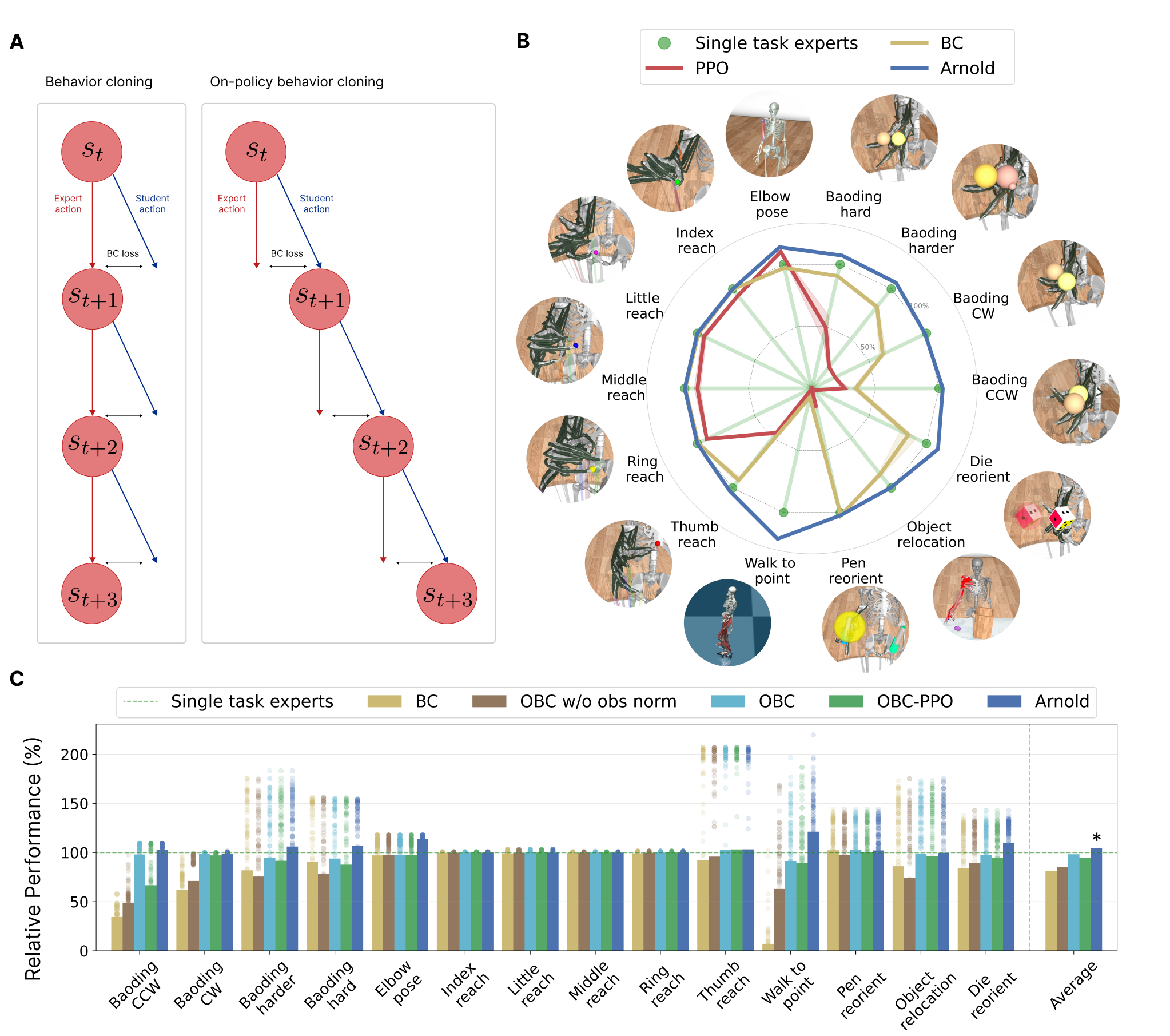}
    \caption{\textbf{Multi-task imitation learning performance.}  \textbf{A:} Difference between standard (top) and on-policy BC (OBC) (bottom). By using on-policy trajectories to train the student policy, we ensure that the transitions adhere to the same distribution as the ones the agent will encounter during deployment.
    \textbf{B:} Multi-task performance comparison between PPO, BC and Arnold, compared to the single-task expert performance (mean $\pm$ s.e.m. across 5 random seeds).
    \textbf{C:} Multi-task performance comparison between BC, variants of OBC, OBC-PPO and Arnold, compared to the single-task expert performance (dots denoting the \emph{relative performance} sampled from 200 episodes). All performances are measured as \emph{relative performance}, defined in the \emph{Problem formulation} section as the policy's solved fraction divided by that of the corresponding single-task expert. Arnold significantly outperformed the single-task experts on aggregate (two-sided Wilcoxon signed-rank test, N=14 tasks, p=0.0166).}
    \label{fig:performance}
\end{figure}

In standard BC~\citep{pomerleau1988alvinn,michie1990cognitive,sammut1992learning}, a student policy learns to mimic the actions of an expert from a dataset of state-action pairs collected from the expert's interaction with the environment. However, the student policy often underperforms relative to the expert, primarily due to imitation error and the generalization gap~\citep{ross2011reduction}. The latter is particularly relevant: since the training data is generated by the expert, it does not reflect the distribution of states encountered by the student during deployment. As a result, even small discrepancies in action selection can compound over time, leading the student into out-of-distribution states and significantly degrading performance. This distributional mismatch can be mitigated when both the expert policy and the environment are accessible. DAgger~\citep{ross2011reduction} iteratively collects trajectories using the current student policy, annotates the visited states with the expert's actions, and \emph{aggregates} these into a growing dataset on which the student is retrained, so that the training distribution progressively converges to the states the student encounters at deployment.

On-policy behavior cloning (OBC) retains DAgger's central idea of training on student-visited states labeled by the expert but differs in one respect: rather than aggregating student rollouts into a growing offline dataset across iterations~\citep{ross2011reduction,bi2018navigation,kelly2019hg,sun2024mega}, OBC discards past data entirely and trains the policy online on the rollouts of its current on-policy rollout buffer~\citep{lee2020learning,caggiano2023myochallenge,li2024learning,he2026viral}, with the expert providing target actions at every step (Figure~\ref{fig:performance}A)~\citep{chiappa2024autobidding}.
This on-policy formulation, which has recently grown in popularity~\citep{lee2020learning,lee2024learning,li2026omniclone}, is particularly important in the first stage of Arnold.
In standard behavior cloning, the student is trained on expert-generated states but is evaluated on states induced by its own actions; errors therefore compound when the student drifts away from the expert distribution.
By labeling the student's own rollouts with expert actions, OBC trains Arnold on the state distributions it will encounter at deployment, providing an initial unified policy that can operate on all tasks without catastrophic failure.
This online formulation confers two practical advantages in our multi-task, high-dimensional setting.
First, it removes the need to store and replay an ever-growing dataset across 14 tasks and four embodiments, which would otherwise scale unfavorably.
Second, and more importantly, by reframing imitation as a fully on-policy procedure, OBC integrates directly with on-policy reinforcement learning methods such as PPO: when a reward function is available, the behavior cloning loss can be combined with the policy gradient objective within a single update, allowing the policy to jointly optimize for imitation and task reward~\citep{wu2026perceptive}.
We deploy this combined approach, which we refer to as OBC-PPO, in the subsequent fine-tuning stage to surpass the original experts.

Across the 14 tasks, OBC succeeded in narrowing the gap to the expert policies (Table~\ref{tab:ablations}, Table~\ref{tab:ablations_per_task}).
Notably, in contrast to OBC, standard BC showed poor performance in the \emph{Walk to point}, and \emph{Object Relocation} tasks, where out-of-distribution error accumulation was expected to have the greatest impact, e.g., by causing a fall or loss of control of the objects.
A task-specific token embedding scheme showed diminished performance on several tasks, suggesting that token sharing promotes effective representations of sensory and motor components (OBC + Task-SV, Table~\ref{tab:ablations}, Table~\ref{tab:ablations_per_task}, Figure~\ref{fig:overview_with_vocabulary}).
OBC-PPO performed largely on par with OBC, with a notable performance drop in the Baoding tasks.
For these comparisons, BC, OBC and OBC-PPO used the same network architecture, hyperparameters, and rebalanced parallel experience collection setup as PPO, isolating the effect of the imitation strategy itself (see Methods).

\subsection*{Super-expert performance with self-distillation}
Expert single-task policies might not be globally optimal, leaving room for further performance improvement.
Inspired by fine-tuning techniques applied to large language models~\citep{deepseek-ai2025deepseekr1} and RL fine-tuning with an augmented loss~\citep{rajeswaran2018learning}, we used the policies trained with OBC as a starting point for further RL training (Figure~\ref{fig:overview}: Step 2).
For a subset of tasks, fine-tuning led to the agent surpassing the performance of the expert~(Table~\ref{tab:arnold-expert-significance}, Figure~\ref{fig:rl_fine_tuning}).
Using these improved policies as new experts, we distilled them back into the unified policy (Figure~\ref{fig:overview}: Step 3), achieving overall super-expert performance (Arnold, Table~\ref{tab:ablations}, Table~\ref{tab:ablations_per_task}, Figure~\ref{fig:performance}C).

Thus, we succeeded in training Arnold on 14 complex musculoskeletal tasks comprising different embodiments with expert or super-expert performance. We qualitatively illustrate successful trials for some behaviors of Arnold (Video~\ref{fig:video_1}) and a grid movie of all tasks followed by some unsuccessful episodes (Video~\ref{fig:video_2}).

\begin{table}[t]
\centering
\small
\setlength{\tabcolsep}{4pt}
\renewcommand{\arraystretch}{1.2}
\resizebox{\textwidth}{!}{%
\begin{tabular}{l cccc | ccccccc}
\toprule
& \multicolumn{4}{c|}{\textbf{Functionality}} & \multicolumn{7}{c}{\textbf{Solved fraction performance (Relative to Experts\%)}} \\
\cmidrule(lr){2-5} \cmidrule(lr){6-12}
\textbf{Method}
& \rotatebox{90}{Transformer}
& \rotatebox{90}{Sensorimotor Vocab.}
& \rotatebox{90}{On-policy BC}
& \rotatebox{90}{Self-distillation}
& \textbf{Finger reach}
& \textbf{Elbow pose}
& \textbf{Baoding}
& \textbf{Reorient}
& \textbf{Object Relocation}
& \textbf{Walk to point}
& \textbf{Avg} \\
\midrule
MT-PPO
& \xmark & \xmark & N/A & N/A
& $35.0 \pm 9.5$ & $113.5 \pm 0.1$ & $28.8 \pm 2.0$ & $14.4 \pm 1.9$ & $0.0 \pm 0.0$ & $0.4 \pm 0.2$ & $30.9 \pm 4.0^{*}$ \\
MT-SAC
& \xmark & \xmark & N/A & N/A
& $0.1 \pm 0.0$ & $20.1 \pm 3.6$ & $4.6 \pm 2.1$ & $0.4 \pm 0.0$ & $0.0 \pm 0.0$ & $0.3 \pm 0.2$ & $2.8 \pm 0.8^{*}$ \\
PPO + T
& \cmark & \xmark & N/A & N/A
& $1.9 \pm 1.8$ & $18.3 \pm 1.2$ & $5.0 \pm 4.4$ & $2.4 \pm 1.6$ & $0.0 \pm 0.0$ & $0.3 \pm 0.2$ & $3.8 \pm 2.2^{*}$\\
PPO + T + SV
& \cmark & \cmark & N/A & N/A
& $82.8 \pm 1.2$ & $110.1 \pm 0.6$ & $29.2 \pm 3.1$ & $9.4 \pm 1.4$ & $0.0 \pm 0.0$ & $0.5 \pm 0.1$ & $47.2 \pm 1.1^{*}$\\
BC (offline)
& \cmark & \cmark & \xmark & \xmark
& $98.0 \pm 0.5$ & $97.2 \pm 0.1$ & $67.2 \pm 1.9$ & $93.3 \pm 3.6$ & $86.1 \pm 1.7$ & $7.1 \pm 0.8$ & $81.1 \pm 1.1^{*}$ \\
OBC + Task-SV
& \cmark & \xmark & \cmark & \xmark
& $88.1 \pm 0.1$ & $96.8 \pm 0.5$ & $53.8 \pm 5.8$ & $70.4 \pm 7.9$ & $46.9 \pm 4.6$ & $16.8 \pm 11.1$ & $68.3 \pm 3.5^{*}$\\
OBC w/o Obs. Norm.
& \cmark & \cmark & \cmark & \xmark
& $99.0 \pm 0.6$ & $97.6 \pm 0.5$ & $68.5 \pm 4.1$ & $93.6 \pm 0.7$ & $74.4 \pm 3.9$ & $63.0 \pm 6.0$ & $85.1 \pm 1.0^{*}$\\
OBC
& \cmark & \cmark & \cmark & \xmark
& $100.5 \pm 0.1$ & $97.3 \pm 0.0$ & $96.2 \pm 1.3$ & $100.1 \pm 0.3$ & $99.1 \pm 1.4$ & $91.4 \pm 3.4$ & $98.2 \pm 0.3^{*}$\\
OBC-PPO
& \cmark & \cmark & \cmark & \xmark
& $100.6 \pm 0.0$ & $97.3 \pm 0.1$ & $85.8 \pm 5.9$ & $97.4 \pm 2.2$ & $96.4 \pm 2.3$ & $89.2 \pm 2.0$ & $94.6 \pm 2.1^{*}$\\
\midrule
\textbf{Arnold}
& \cmark & \cmark & \cmark & \cmark
& $100.7 \pm 0.1$ & $113.9 \pm 0.1$ & $103.8 \pm 0.6$ & $106.1 \pm 0.1$ & $99.6 \pm 1.1$ & $121.3 \pm 1.3$ & $104.7 \pm 0.2$\\
\bottomrule
\end{tabular}
}
\caption{\textbf{Ablation study across task families.}
Each ablation removes one functionality (\cmark{} retained, \xmark{} ablated)
from the full Arnold model or replaces a training stage.
Performance is the solved fraction relative to the single-task expert policies
(\%). For each method we compute, per random seed, the average solved-step
fraction on every task normalised by its expert; cells report the mean
$\pm$ s.e.m.\ of these per-seed values across seeds. Asterisks ($^{*}$) on the
\emph{Avg} column denote a statistically significant decrease in overall
performance relative to the full Arnold model (two-sided Wilcoxon signed-rank
test over the $N=14$ tasks, Holm--Bonferroni corrected across methods,
$p<0.05$; see Methods).
The \emph{Finger reach} family aggregates the five single-finger reaching tasks;
\emph{Baoding} the four Baoding variants; \emph{Reorient} the die and pen reorient tasks.}
\label{tab:ablations}
\end{table}

\subsection*{Multi-task training encourages universal control smoothness}

Distilling multiple expert policies into a single control framework while securing additional performance gains conveys several practical advantages; for instance, this consolidation can lead to more efficient deployment.
However, performance metrics alone are insufficient to address whether Arnold truly captures universal concepts about motor control that transcend individual tasks, or if it simply compresses single-task policies into segregated regions of the network's parameter space.
One such universal concept is action and kinematic smoothness.
Smoothness is among the most robust regularities of biological movement: goal-directed reaches of the human arm and eye follow stereotyped trajectories with single-peaked, near-symmetric velocity profiles~\citep{flash1985coordination, harris1998signal}.
Because smoothness recurs across effectors and tasks, it is a natural candidate for a task-transcending control regularity, prompting us to ask whether it emerges in Arnold as a by-product of multi-task training.

We tested whether OBC under a unified sensorimotor vocabulary leads to smoother muscle activity and kinematics by comparing activity and kinematic trajectories to those generated by the expert policies and by policies trained with the Arnold network architecture, but in a single task at each time.
For consistency across metrics, we selected the 11 tasks that used the MyoHand embodiment. Each policy was deployed for 100 episodes on each task.
We quantified muscle activity smoothness by calculating the mean absolute difference $|\Delta a|$~\citep{mysore2021regularizing}, averaged across episodes.
We also quantified kinematic smoothness by calculating spectral arc length (SPARC), log-dimensionless jerk (LDLJ), and number of velocity peaks (NVP)~\citep{balasubramanian2015analysis, rohrer2002movement, hogan2009sensitivity}.

\begin{figure}[t!]
    \centering
    \includegraphics[width=1\linewidth]{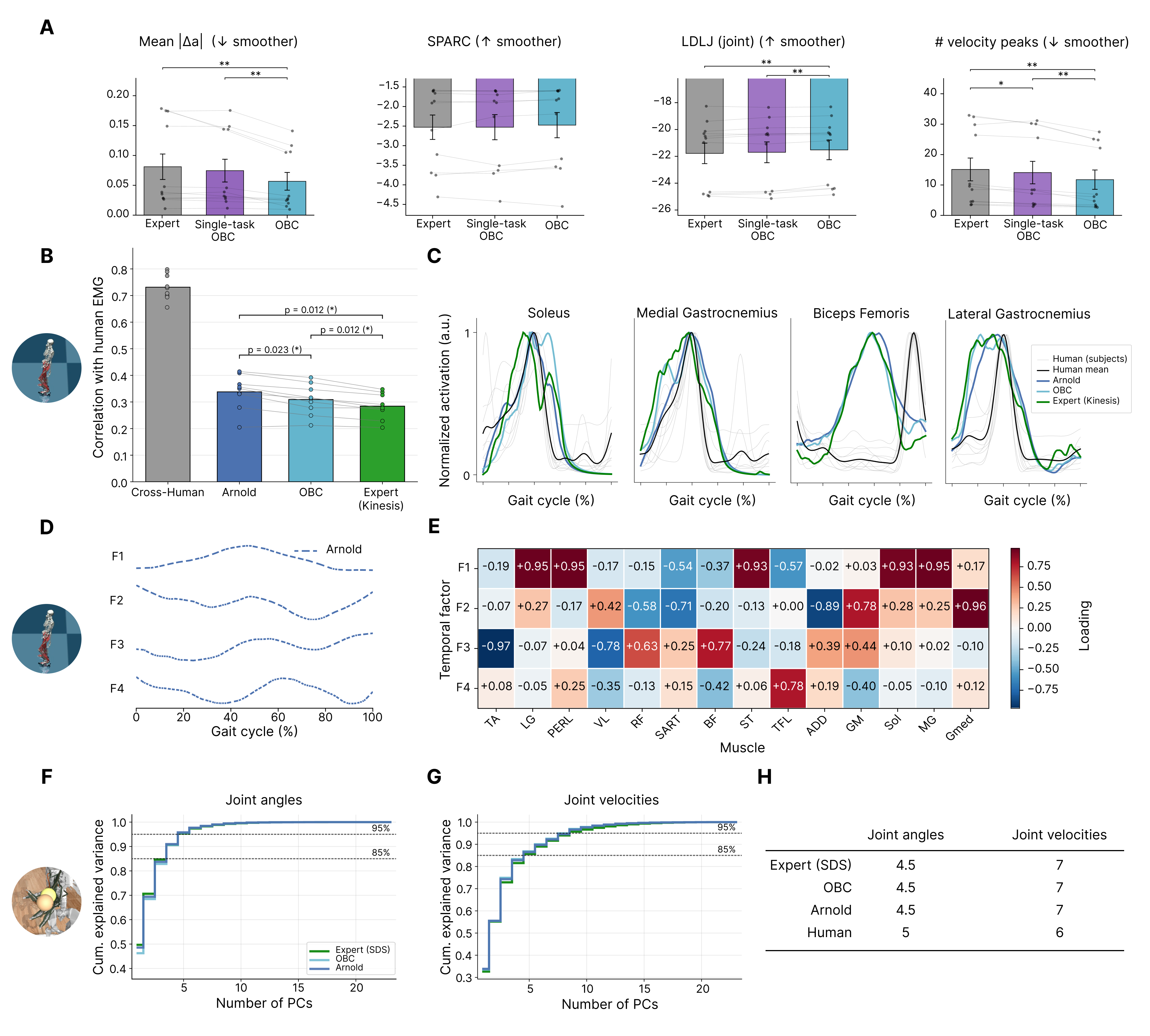}
    \caption{\textbf{Analysis of learned control strategies.} \textbf{A}: Action and kinematic smoothness metrics across 11 MyoHand tasks for three policies; per-task PPO Expert (gray), Single-task OBC (purple), Multi-task OBC (teal). Metrics: Mean |$\Delta$a| (mean absolute step-to-step change in muscle activation, $\downarrow$ smoother), SPARC (spectral arc length, $\uparrow$ smoother), LDLJ joint (log dimensionless jerk of joint kinematics, $\uparrow$ smoother), and number of velocity peaks ($\downarrow$ smoother). Bars are mean across tasks $\pm$ SEM (n = 11 tasks); individual points denote per-task means. Statistics: task-paired two-sided Wilcoxon signed-rank test, Holm-Bonferroni corrected over the 3 pairwise comparisons; * p < 0.05, ** p < 0.01 (only significant pairs shown).
    \textbf{B}: Average correlation between average gait-normalized generated muscle activity patterns in the Walk to Point task and recorded human EMG activity from nine leg muscles during forward gait (N=9 subjects). Gray: Human-human correlation; points indicate 9 leave-one-out comparisons between one subject and the mean of the other eight. Blue: Arnold (after self-distillation). Teal: OBC. Green: Kinesis (expert policy). Individual points indicate correlation with human subjects (averaged across muscles). Statistics: paired two-sided Wilcoxon signed-rank test on Fisher-z-transformed correlations, Bonferroni-corrected over the 3 pairwise policy comparisons; * p < 0.05.
    \textbf{C}: Average gait-normalized generated muscle-activity / EMG patterns for four representative muscles (Soleus, Medial Gastrocnemius, Lateral Gastrocnemius, Biceps Femoris) of the nine analyzed. Thin gray: individual human subjects; black: human subject-mean; blue: Arnold; teal: OBC; green: Kinesis. Each trace is max-normalized to its peak (normalized activation, a.u.) over the gait cycle.
    \textbf{D}: Temporal activation factors F1–F4 obtained by applying varimax-rotated PCA on the muscle × gait-cycle activation matrix generated by Arnold in the Walk to Point task~\citep{ivanenko2004five}. Each factor is a temporal waveform spanning the normalized gait cycle. Four factors were retained (eigenvalue $\geq$ 0.5), accounting for \textasciitilde94.7\% of the variance across the analyzed muscles.
    \textbf{E}: Spatial loadings of each temporal factor (rows F1–F4) onto the 14 analyzed leg muscles present in \citeauthor{ivanenko2004five} (columns: TA = tibialis anterior, LG = lateral gastrocnemius, PERL = peroneus longus, VL = vastus lateralis, RF = rectus femoris, SART = sartorius, BF = biceps femoris long head, ST = semitendinosus, TFL = tensor fasciae latae, ADD = adductor longus, GM = gluteus maximus, Sol = soleus, MG = medial gastrocnemius, Gmed = gluteus medius; multi-compartment glutei aggregated by summing activations).
    \textbf{F,G}: Cumulative explained variance of the PCs of the joint angles (\textbf{F}) and joint velocities (\textbf{G}) for Arnold when deployed on the Baoding CCW task).
    \textbf{H}: Comparison between the number of independent degrees of freedom observed in Arnold, the expert policy (SDS~\citep{chiappa2024acquiring}), and human experimental data~\citep{todorov2004analysis} for the Baoding task. The values are obtained by averaging the number of PCs necessary to explain 85\% and 95\% of the variance of the joint positions and the joint velocities, respectively.}
    \label{fig:smoothness_biomech}
\end{figure}

Across 11 hand tasks, multi-task OBC was significantly smoother than single-task experts and single-task OBC on three out of four metrics ($|\Delta a|$, p < 0.01; LDLJ, p < 0.01; NVP, p < 0.01). Single-task OBC was also significantly smoother than single-task experts in terms of velocity peaks (p < 0.05), indicating that both the network architecture and multi-task training lead to smoother actions and kinematic trajectories (Figure~\ref{fig:smoothness_biomech}A).

\subsection*{Arnold inherits and improves upon biomechanically plausible control strategies}

We further sought to test whether the control strategies of Arnold align with empirical evidence from human motor control. For this analysis, we focused on two challenging task families for which human data exists, namely \emph{Walk to point} and \emph{Baoding}. Prior studies that developed single-task specialist policies (Kinesis and SDS for Walk to point and Baoding, respectively~\citep{simos2026reinforcement, chiappa2024acquiring}) have investigated the alignment between the learned control policy and human behavior, thus providing a relevant baseline.

To test whether Arnold's Walk-to-point behavior reflects human-like muscle recruitment rather than an arbitrary control solution, we compared its simulated activations against surface EMG from nine human subjects walking on a treadmill~\citep{wang2023wearable}.
For nine lower-limb muscles with direct MyoLeg counterparts (soleus, medial/lateral gastrocnemius, tibialis anterior, semimembranosus, biceps femoris, and the vasti and rectus femoris), we correlated each policy's cycle-averaged activation waveform with each subject's EMG over the normalized gait cycle (see Methods).
Across subjects, Arnold reproduced the human activation timing significantly better than both the locomotion specialist (Kinesis expert, p = 0.012) and its behaviorally-cloned distillation (OBC, p = 0.023), although all models remained below the cross-human reference ceiling~(Figure~\ref{fig:smoothness_biomech}B, C).
Thus the multi-task motor prior, despite never being optimized to match EMG, yields the most human-like locomotor recruitment of the policies tested.

To move from individual muscles to coordinated patterns, we applied a temporal factorization approach based on~\citeauthor{ivanenko2004five}~\citep{ivanenko2004five} to Arnold's locomotion rollouts, focusing on the subset of muscles common to the reported human study.
Four temporal factors accounted for \textasciitilde~94.7\% of the activation variance~(Figure~\ref{fig:smoothness_biomech}D, E).
Factor~1 grouped the three triceps surae heads together with peroneus longus, corresponding to the generation of propulsive impulse at push-off.
Factor~3 was anchored by tibialis anterior, the principal ankle dorsiflexor active in terminal swing, consistent with dorsiflexion at the late gait phase.
The two remaining factors were more directly associated with stabilization during gait: factor 2 grouped hip abductors in anti-phase with adductors, consistent with frontal-plane stabilization, while factor 4 was principally linked to tensor fasciae latae, a muscle associated with pelvic balance.
The apparent increased emphasis on stabilization compared to factors derived from human data is expected given the lack of abdominal and upper body musculature which normally aids in stabilization, as well as the more dynamic task of walking to a point compared to treadmill walking.

In the Baoding tasks, we asked whether Arnold recovers biomechanically plausible hand coordination.
Following the kinematic-dimensionality analysis of human subjects~\citeauthor{todorov2004analysis} ~\citep{todorov2004analysis}, we ran PCA on the 23 MyoHand joint angles collected during \emph{Baoding CCW} rollouts (100 episodes per policy) and defined apparent dimensionality as the mean number of principal components required to explain 85\% and 95\% of the joint-position variance~(Figure~\ref{fig:smoothness_biomech}F, G).
Arnold's kinematics spanned the same dimensionality as the per-task expert~\citep{chiappa2024acquiring}. These values broadly align with the reduced dimensionality reported for human hand kinematics during equivalent manipulation~\citep{todorov2004analysis}, suggesting that Arnold recovers the task's intrinsic coordination structure (Figure~\ref{fig:smoothness_biomech}H). 

Overall, we found that Arnold has smoother behavior than the experts, and that its behavior resembles human behavior in terms of EMG and kinematic dimensionality. 

\subsection*{Arnold is more data-efficient for novel tasks}

A multi-task policy should show improved learning capacity in novel motor control tasks.
Whereas the addition of new tasks is incompatible with the fixed observation spaces employed in traditional multi-task approaches, Arnold's \emph{sensorimotor vocabulary} facilitates the expansion of the task space even post-training thanks to its compositional properties.
For instance, in a new reaching task not encountered during training (e.g., using a different finger), Arnold learns to perform the task faster compared to training from scratch, suggesting that it partially reuses the token embeddings learned for the muscles, joints, and objectives that recur across the training tasks.

\begin{wrapfigure}{r}{0.45\textwidth}
    \centering
    \includegraphics[width=\linewidth]{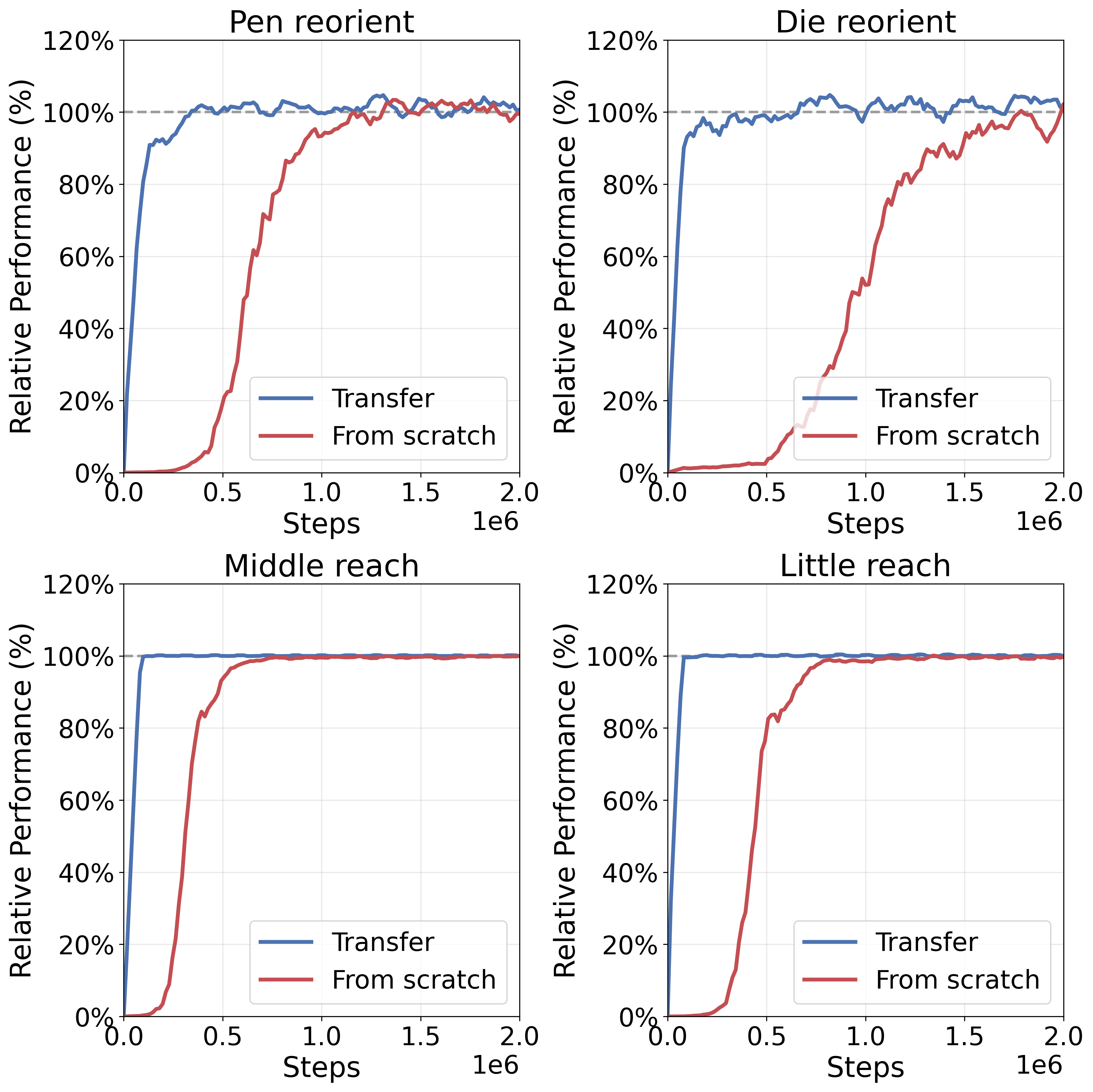}
    \caption{\textbf{Testing generalization.} Learning curves of single-task training experiments, pretrained network (blue) vs randomly initialized network (red). The pretrained network experienced 50M steps of the 10 remaining tasks before being trained on Pen reorient, Die reorient, Middle reach and Little reach with OBC-PPO (one at a time). It learns the unseen tasks considerably faster than a random network, suggesting that it learned transferable representations in the pretraining phase. All the performances are measured with the \emph{relative performance} defined in the \emph{Problem formulation} section}
    \label{fig:few-shot}
    \vspace{-8pt}
\end{wrapfigure}

To assess transferability, we pretrained Arnold on ten tasks with OBC and then resumed training on four held-out tasks with OBC-PPO (Little reach, Middle reach, Pen reorient, Die reorient).
On all four downstream tasks, the pretrained network reached expert performance substantially earlier than a randomly initialized network (Figure~\ref{fig:few-shot}), with the largest gains on the manipulation tasks (Pen reorient, Die reorient), which require the most training from scratch.
While the incorporation of previously unseen token compositions or objects precluded meaningful zero-shot transfer, the pre-trained contextual representations provided a foundation from which new tasks were learned with substantially less data.

\subsection*{Muscle synergies show limited transfer across tasks}

Humans and many other animals are motor generalists, coordinating over-actuated, high-dimensional motor systems across tasks, from watchmaking to sprinting.
A striking empirical regularity emerges from studies of this coordination: motor outputs do not exploit the full space of possible muscle activations, but appear to occupy lower-dimensional subspaces often characterized by muscle synergies~\citep{mussa1994linear,davella2001modularity,d2003combinations,tresch2009case,alessandro2013muscle,loeb2021learning}. Indeed, the low-dimensional kinematics exhibited by both humans and Arnold in the Baoding task (Figure~\ref{fig:smoothness_biomech}F-G) can be interpreted through the lens of synergies. 

\begin{figure}[t]
    \centering
    \includegraphics[width=1\linewidth]{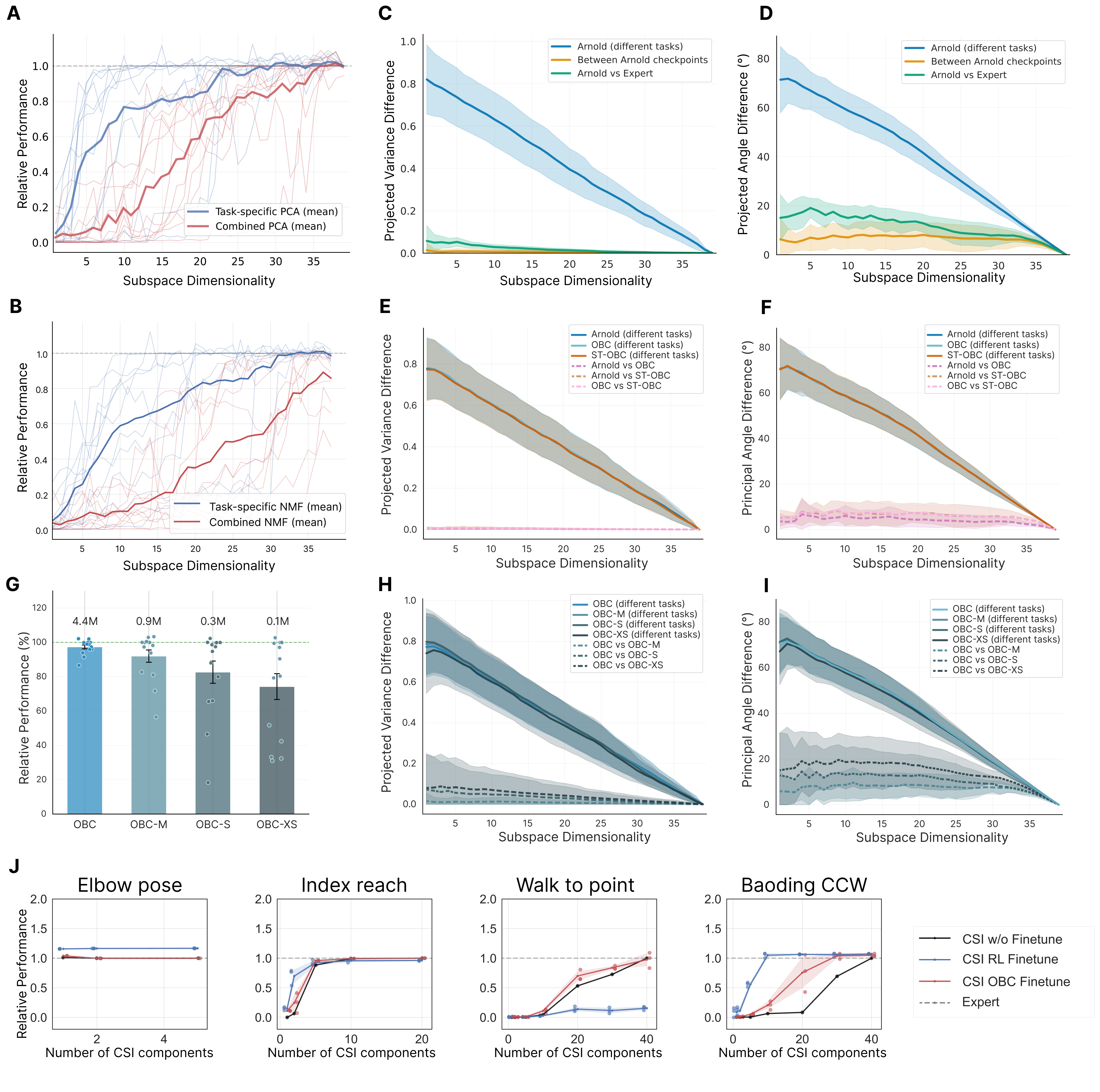}
    \caption{
    \textbf{Dimensionality analysis.}
    \textbf{A, B}: Average relative task performance across the 11 MyoHand tasks, when the control signal is projected on action subspaces generated by progressively more principal components (PCA, A) or non-negative factors (NMF, B) (from 1 to 39 muscles, left to right). Blue: task-specific principal components. Red: shared principal components. Shaded lines: task-specific graphs. Plots per task can be found in Figures~\ref{fig:ev_v_csi}, ~\ref{fig:ev_v_csi_nmf}.
    \textbf{C, D, E, F, H, I}: Projected Variance Difference (PVD) measures how well synergies from one condition can explain variance in the other's muscle activation patterns. Principal Angle Difference (PAD) quantifies angular differences between synergy subspaces, with 90° indicating completely orthogonal (maximally different) patterns. \textbf{C, D}: PVD and PAD measured between different tasks performed by Arnold (blue), between different Arnold checkpoints (orange), and between Arnold and expert policies (green).
    \textbf{E, F}: PVD and PAD measured between Arnold, OBC, and single-task OBC, across tasks and across policies.
    \textbf{G}: Average performance (solved step fraction) relative to single-task expert policies by OBC policies across different network parameter sizes (shown above the bars). Individual points represent task-wise performance for each of the 11 MyoHand tasks.
    \textbf{H, I}: PVD and PAD measured between OBC policies of different network parameter size, across tasks and across policies.
    \textbf{J}: Performance comparison between fine-tuned CSI policies and frozen (black) CSI policies for indicative tasks. MLP policies are trained with OBC for 5M steps and then constrained to a lower-dimensional action space. Fine-tuned policies are further trained on this action space for 5M additional steps, either with OBC (red) or with PPO (blue). Complete results shown in Figure~\ref{fig:csi_all_tasks}.}
    \label{fig:pca_analysis}
\end{figure}

Artificial agents like Arnold offer a unique opportunity for better understanding emerging coordination principles; of course with the caveat that the principles might be different in biology. Unlike classic synergy analyses, which rely on reconstructing kinematic or EMG signals, an embodied policy permits a functional test: whether learned control remains successful when restricted to a low-dimensional subspace, and whether such subspaces transfer across tasks~\citep{chiappa2024acquiring}. To address these questions we performed control subspace inactivation (CSI) analysis~\citep{chiappa2024acquiring} on muscle activations generated by Arnold across the 11 hand (MyoHand) tasks, identifying control subspaces both per task and across all tasks combined, using principal component analysis (PCA) and non-negative matrix factorization (NMF) (see Methods).
Projecting control signals onto task-specific subspaces, relatively few synergies sufficed to reach full task performance, particularly in reaching tasks (Figures~\ref{fig:pca_analysis}A,B~and per task Figure~\ref{fig:ev_v_csi},~\ref{fig:ev_v_csi_nmf}), confirming that Arnold compresses muscle activity into low-dimensional subspaces within individual tasks.
However, substantially more components were required when the subspace was shared across tasks, indicating that these low-dimensional subspaces are highly task-specific.
These results were consistent across the two factorization methods.
Repeating the analysis on the single-task expert policies yielded similar performance profiles (Figure~\ref{fig:arnold_v_expert}). We note that CSI is a functional measure, relating performance to dimensionality, rather than a signal-reconstruction measure such as cumulative explained variance (EV).
EV systematically underestimates the functional dimensionality of the task (Figure~\ref{fig:ev_v_csi}), consistent with prior work on specialist policies for Baoding~\citep{chiappa2024acquiring}.

Next we analyzed if the emerging muscle synergies are universal. To test the robustness of the task-specificity, we quantified subspace differences using Projected Variance Difference (PVD) and Principal Angle Difference (PAD) measures (see Methods).
Across both measures, differences between tasks dominated differences between policies: subspaces differed far more across task pairs than between independent Arnold checkpoints, between Arnold and the single-task experts (Figure~\ref{fig:pca_analysis}C,D), or between Arnold and multi-task or single-task OBC (Figure~\ref{fig:pca_analysis}E,F).

We further asked whether the absence of transferable structure could be an artifact of network capacity, since a sufficiently large network may host independent task-specific solutions without sharing representations.
We trained OBC policies at three additional reduced network sizes (Table~\ref{tab:obc-architectures}). We found that reduced-capacity models retained considerable performance on all but the most complex tasks (Figure~\ref{fig:pca_analysis}G, Figure~\ref{fig:capacity_full_tasks}).
The synergy structure of those models remained task-specific and policy-invariant under PVD and PAD (Figure~\ref{fig:pca_analysis}H,I).
These results suggest that task-specificity reflects the structure of the tasks rather than spare capacity.

A potential confound of CSI is that projecting activations onto a restricted subspace can push the policy out of its training distribution, producing performance drops unrelated to true functional dimensionality.
We addressed this by re-optimizing CSI-constrained MLP specialists within the restricted subspace using either PPO or OBC (see Methods).
For tasks where PPO trains well from scratch (\emph{Elbow pose}, \emph{Reach}) or from a strong pretrained initialization (\emph{Baoding CW/CCW}), PPO fine-tuning recovered performance at lower dimensionalities than the frozen CSI curves~(Figure~\ref{fig:pca_analysis}J).
For the most challenging tasks (\emph{Walk to point}, \emph{Object relocation}, \emph{Baoding hard/harder}), PPO fine-tuning instead degraded performance, consistent with optimization difficulty rather than a dimensionality constraint.
OBC fine-tuning mostly preserved the frozen CSI curves across tasks, indicating that the projections were not severely out-of-distribution.
CSI thus provides a reasonable upper bound on functional dimensionality, which PPO can tighten where optimization remains tractable.

Together, these analyses support three conclusions. First, the control structure as observed through muscle synergies is consistent across Arnold checkpoints, training stages, and the original experts, and is preserved across network capacities. Second, this structure is task-specific: Arnold learns appropriate per-task solutions rather than a universal motor representation, even after self-distillation. Third, signal-reconstruction analyses underestimate the dimensionality of control required for successful performance, so the low dimensionality reported by variance-based methods may not by itself indicate a simplified control strategy.

\section{Discussion}
\label{sec:discussion}

We introduced Arnold, a multi-task, multi-embodiment policy for musculoskeletal motor control. A single agent achieved expert or super-expert performance across diverse tasks involving different body parts, objects, and objectives. This versatility rested on four key components: a compositional \emph{sensorimotor vocabulary}, on-policy behavior cloning, self-distillation, and the careful design of a multi-task learning framework.
In the following paragraphs, we situate Arnold within the broader landscape of research on muscle-based control, multi-task learning, and transformer architectures, and we discuss the potential implications about biological motor control.

\paragraph{Muscle control.}
Controlling models of the human body through muscles has long been an active research topic, spurred by the introduction of biomechanical simulators like OpenSim~\citep{delp2007opensim,seth2011opensim}, more recently, MyoSuite~\citep{caggiano2022myosuite}. Both of those simulators featured in almost a decade of NeurIPS competitions~\citep{kidzinski2018learning,kidzinski2018learninga,kidzinski2019artificial,kidzinski2020artificial,caggiano2023myochallenge,caggiano2024myochallenge,caggiano2024myochallengea, wang2026myochallenge}.
Musculoskeletal control has also been pursued on complementary simulation platforms~\citep{geijtenbeek2019scone, rasmussen2002anybody}. While a direct cross-platform comparison falls outside the scope of this work, the core algorithmic challenge of controlling high-dimensional, over-actuated systems is shared across simulators. Recent contributions outside the MyoSuite framework have advanced biologically plausible reward design for locomotion and scalable exploration for high-dimensional musculoskeletal control~\citep{schumacher2025natural, wei2026scalable}. Crucially, however, these works, like their MyoSuite-based counterparts, each address a single task or embodiment at a time; Arnold is, to our knowledge, the first policy to simultaneously master dexterous object manipulation, reaching, and locomotion within a unified framework spanning multiple musculoskeletal embodiments.

In the domain of motion imitation from reference motion capture, techniques such as PPO~\citep{schulman2017proximal} and model-based learning~\citep{yao2022controlvae} have been extended to musculoskeletal control~\citep{feng2023musclevae,simos2026reinforcement,li2026towards}.
However, significant challenges remain in RL tasks that either lack reference motion data, or involve dexterous object manipulation, motivating work on exploration~\citep{schumacher2022dep,chiappa2023latent}, learned low-dimensional action spaces~\citep{berg2023sar,he2024dynsyn}, curricula~\citep{caggiano2023myochallenge,chiappa2024acquiring}, world models~\citep{hansen2024td} and reward shaping~\citep{schumacher2025natural}. In contrast, here we leverage on-policy behavior cloning (OBC) to combine multiple single-task experts into a unified policy.

\paragraph{From single-task to multi-task policies.} 

Policy distillation is commonly used to tackle the challenge of multi-task learning, transferring knowledge from teacher policies into a single student~\citep{rusu2016policy,parisotto2016actormimic}, or task-specific policies coordinate around a central distilled policy~\citep{teh2017distral}.
Subsequent work on multi-task policies extended this idea by leveraging demonstrations or specialist policies to guide and improve the fine-tuning of multi-task policies.
These demonstrations can either be actively selected~\citep{bagatella2024active}, generated by RL rollouts~\citep{rajeswaran2018learning,xu2024rldg}, derived from diverse offline data sources~\citep{furuta2023system}, or focused on difficult tasks~\citep{jia2022improving}.
While most research has focused on robotic systems, Arnold tackles multi-embodiment, multi-task learning in the inherently non-linear and high-dimensional domain of musculoskeletal dynamic simulation. 
Notably, Arnold operates on an anatomically structured sensorimotor representation that overlaps with the biological motor system, including muscle-state variables, joint kinematics, task goals, and object states. Unlike the specialists that receive high-level proprioceptive signals such as joint angles, Arnold operates on muscle-state variables, which are more closely related to what the proprioceptive system needs to operate on~\citep{proske2012proprioceptive,vargas2024taskdriven,perez2025deep}.

\paragraph{Transformers in motor control and robotic foundation models.}
Transformers have found application in reinforcement learning by modeling decision-making as a sequence-to-sequence problem~\citep{chen2021decision}, and, more relevant to our work, by flexibly processing a variable number of sensory inputs and motor commands~\citep{reed2022generalist,zargarbashi2024robotkeyframing,team2024octo} across variable embodiments~\citep{chiappa2022dmap,trabucco2022anymorph,wang2024scaling,sferrazza2024body}.
To our knowledge, Arnold is the first work applying this idea to musculoskeletal models, providing a recipe to distill expert policies from multiple embodiments and tasks into a single agent and to surpass them through RL fine-tuning. 


Arnold also connects to the emerging literature on robotic transformer and foundation policies. Recent generalist robot models such as RT-1~\citep{brohan2023rt1}, Octo~\citep{team2024octo}, and OpenVLA~\citep{kim2024openvla} use transformer-based architectures and large-scale pretraining to map sensory inputs, language goals, and robot trajectories to actions. Arnold addresses a complementary bottleneck: how to represent and control heterogeneous bodies whose observation and action spaces differ in both dimensionality and semantics. Arnold provides a recipe for extending foundation-model ideas from vision-language-action manipulation to morphology-variable control. However, unlike current robotics models, Arnold is trained in simulation, does not use internet-scale vision-language pretraining, and likely due to the comparatively smaller scale does not solve novel tasks zero-shot. 

\paragraph{Relevance for artificial embodied intelligence.}
Discovering viable policies in realistic musculoskeletal environments requires specialized techniques~\citep{caggiano2023myochallenge,caggiano2024myochallengea,caggiano2024myochallenge,schumacher2022dep,he2024dynsyn,berg2023sar,chiappa2023latent,simos2026reinforcement,feng2023musclevae}.
We sidestepped this problem by using several of those specialized policies, trained with curriculum learning and motion imitation as teachers, reducing the challenge to distilling multiple specialists into a single unified network.
Pure behavior cloning failed in environments where precision and dexterity are paramount, whereas OBC reached near-expert performance across all tasks, and task-specific PPO fine-tuning followed by self-distillation matched or surpassed the original experts.
Arnold also learned novel tasks substantially faster than randomly initialized networks, indicating that multi-task training encapsulates transferable principles of motor control.
We expect this framework to scale to richer task sets and more complete biomechanical models.

\paragraph{Analysis of learned control strategies.}

A hallmark of natural embodied intelligence is the ability to extract universal control strategies and transfer them to new tasks~\citep{bernstein1967coordination,mathis2024adaptive}.
We used dimensionality analysis to characterize whether such strategies translate into generalizable low-dimensional control components broadly known as muscle synergies~\citep{mussa1994linear,davella2001modularity,d2003combinations,tresch2009case,alessandro2013muscle,loeb2021learning}.
Within individual tasks, Arnold's control occupied low-dimensional subspaces, but these subspaces did not transfer across tasks; the synergies extracted from Arnold were closer to those of the single-task experts than the tasks were to one another.
A natural concern is that this task-specificity could be an artifact of network capacity, since a sufficiently large network may host independent, per-task solutions without sharing representations.
Training OBC policies across a range of network sizes ruled this out: smaller models largely retained performance on all but the most complex tasks while exhibiting the same task-specific, policy-invariant synergy structure.
The absence of transferable synergies may therefore be a property of the tasks and the training paradigm rather than of spare capacity, and it raises the question of what conditions (more extensive task sets, or different training objectives) are required for universal motor representations to emerge in artificial systems. That being said, the emergence of increased action and kinematic smoothness under multi-task training, as well as the increased efficiency in learning new tasks indicate that some universal motor principles already exist in Arnold. This might suggest universality at different levels of the control hierarchy. Smoothness is a property of the output over time, while muscle-coordination is a property across space.  

A central contribution of this analysis is that functional and signal-based measures of control dimensionality diverge.
Control-subspace inactivation (CSI) shows that variance-based analyses systematically underestimate the dimensionality required for successful task performance: components contributing little to signal reconstruction can be essential for control.
This intervention has no direct experimental counterpart, as selective inactivation of a candidate synergy subspace is not achievable \emph{in vivo} with current techniques.
In-silico models such as Arnold may therefore offer access to a class of causal intervention that biology does not yet afford, generating testable predictions about functional control dimensionality. This result is consistent with the observations of \citeauthor{yan2020unexpected}~\citep{yan2020unexpected}, who showed evidence for task-relevance even in low-variance PCs for hand kinematics.


Where biological data permit direct comparison, Arnold's control is consistent with reported human motor strategies.
On the locomotion task, applying the factor-analysis pipeline of~\citet{ivanenko2004five} to Arnold's muscle activations recovered temporal components whose muscle compositions correspond to established gait modules, while EMG analysis showed that generated muscle activity correlates well with human data recordings from several leg muscles; divergences in the remaining factor and in other muscle activity patterns are interpretable in terms of the task demands and the partial musculature of MyoLeg.
On the Baoding tasks, Arnold matched the expert policy in reproducing the low-dimensional kinematic structure of human hand manipulation~\citep{chiappa2024acquiring}. More broadly, embodied multi-task frameworks of this kind can extend, in the context of motor control, recent investigations of how compositional representations emerge in networks trained on multiple tasks~\citep{johnston2023abstract,driscoll2024flexible,riveland2024natural}.

\paragraph{Limitations and future work.}

While Arnold surpassed expert performance and learned novel tasks faster, it did not discover new control policies without demonstrations.
PPO from scratch succeeded only on the simpler tasks, confirming that \textit{Object relocation} and \textit{Walk to point} require curricula or reference motions~\citep{chiappa2024acquiring,caggiano2024myochallenge,simos2026reinforcement}, highlighting the need for future work on exploration and skill discovery.
In addition, although our benchmark includes arm--hand object relocation and legged locomotion, we do not demonstrate control of a full-body musculoskeletal model requiring simultaneous coordination of arms, legs, trunk, and balance.
Extending Arnold to richer full-body and highly dynamic behaviors is therefore an important direction for future work. Recent musculoskeletal controllers have demonstrated impressive whole-body dynamic behaviors including running, jumping, and acrobatic maneuvers~\citep{wei2026scalable, li2026towards}; these results currently rely on single-motion-imitation policies that do not generalize across movement repertoires, and developing a unified policy that combines such dynamism with Arnold's multi-task breadth represents a compelling open challenge.

Similarly, Arnold did not solve novel tasks zero-shot (Figure~\ref{fig:few-shot}), although pretraining on ten tasks provided a better starting point for unseen tasks than random initialization.
The observed transfer may thus reflect improved initialization in the form of informative prior representations and general control principles rather than the rapid retargeting of learned skills characteristic of humans and
animals~\citep{wolpert2011principles}.
Whether this gap stems from prior experience or from specific biological mechanisms remains an open question. Notably, all core behavior cloning results were obtained with roughly 50M environment interactions (\textasciitilde5 days on a current hardware), highlighting both the efficiency of the approach and the potential for further gains at larger scale.

Lastly, although our proposed method is simulator-agnostic in principle, future work should investigate its transferability as new embodiments and simulating frameworks emerge.

Overall, Arnold establishes a foundation for musculoskeletal control that extends far beyond single tasks and embodiments.
Future work will aim to bring Arnold closer to the experimental neuroscience community by facilitating interactions with it, e.g., via natural language~\citep{simos2026reinforcement}, and to contrast its computation with neural dynamics~\citep{shenoy2013cortical}, pairing motor control studies with simulation experiments that require no machine-learning expertise and open new avenues for neuroscientific research.

\textbf{Acknowledgments:} We thank Andrea d'Avella for additional encouragement to carry out the CSI retraining experiments. We are grateful to Siebe Geurts, Crystal Li as well as other members of the Mathis Group for Computational Neuroscience and AI for helpful feedback. This work was funded by the EPFL School of Life Sciences Summer Research Program (B.A.), Swiss SNF grant (310030$\_$212516) and the Simons Foundation (SFI-AN-NC-SCN-00007276-14).

\newpage

\section{Methods}
\label{sec:methods}

Here, we formalize the multi-task control problem and describe the implementation of the muscle transformer policy. In particular, we detail how Arnold can deal with variable observation and action spaces and outline the implementation of OBC as well as other learning algorithms for multiple tasks.

\subsection*{Musculoskeletal tasks and experts}

In all our experiments, we considered the musculoskeletal models of the MyoSuite library~\citep{caggiano2022myosuite}. MyoSuite is implemented in MuJoCo~\citep{todorov2012mujoco} and features biologically-realistic models of multiple body parts ported from OpenSim~\citep{delp2007opensim,seth2011opensim}. The advantage of MyoSuite compared to OpenSim is its speed, as it simulates these models three orders of magnitude faster, enabling our large-scale experiments.

The 14 tasks featured in this work are based on four different musculoskeletal models (Figure~\ref{fig:overview}):

\textbf{MyoElbow} is the simplest model of the four, with six muscles and one joint. The only task posed for this model involves reaching a target elbow angle (\emph{Elbow pose}). The expert for this task was trained with Lattice-PPO by~\citet{chiappa2023latent}. 

\textbf{MyoHand} comprises a detailed model of a human hand, with 39 muscles and 23 joints, featured in eleven tasks. Five tasks consist of reaching a randomly sampled point with a finger. We trained single-task experts using Lattice-PPO~\citep{chiappa2023latent,schulman2017proximal}. The remaining six are object manipulation tasks. In \emph{Pen reorient} and \emph{Die reorient}, the hand must rotate a pen or a die, respectively, to reach a target pose. The experts for these tasks are from \citet{chiappa2023latent}. The last four tasks involve the rotation of two Baoding balls, clockwise and counter-clockwise, with progressively more randomness in the prescribed trajectories and in the physical properties of the balls ({\it hard} and {\it harder}). Domain randomization via object mass and size is included in the task to enforce robustness against possible modeling inconsistencies. The experts for the Baoding tasks are the winning solutions to the 2022 NeurIPS MyoChallenge~\citep{caggiano2023myochallenge}. A targeted curriculum learning strategy resembling part-to-whole practice was developed for this purpose~\citep{chiappa2024acquiring}. 

\textbf{MyoArm} is an extension of the MyoHand that includes the upper arm, pectoral and shoulder muscles. It has 63 muscles and 38 joints. \emph{Object relocation} is implemented using this model. The agent needs to grasp an object of a dynamically generated shape, lift it and place it inside a box. Extensive domain randomization via object geometry, mass, pose, and friction is included. The expert of \emph{Object relocation} is the winning solution of the 2023 NeurIPS MyoChallenge~\citep{caggiano2024myochallenge}. In training this expert, the authors leveraged Lattice-PPO as well as a curriculum learning strategy. 

\textbf{MyoLeg} is a model of the human lower body with 80 muscles and 28 joints. It is used in the \emph{Walk to point} task, in which the agent has to walk towards a random target location. Training a policy involved imitation of human motion data and a mixture-of-experts architecture, in addition to RL fine-tuning~\citep{simos2026reinforcement}.

We provide additional details on each task and environment, including task descriptions (Figure~\ref{fig:fulltask_description}) and specific task and reward parameters (Table~\ref{tab:env_params_1} and Table~\ref{tab:env_params_2}).

\begin{table}[H]
    \centering
    \resizebox{\textwidth}{!}{
        \begin{tabular}{lllllllllll}
        \toprule
        Task & \multicolumn{2}{l}{Elbow pose} & \multicolumn{2}{l}{Finger reach} & \multicolumn{2}{l}{Pen reorient} & \multicolumn{2}{l}{Die reorient} & \multicolumn{2}{l}{Object relocation} \\
        {} &       Parameter &        Value &       Parameter &         Value &       Parameter &          Value &      Parameter &         Value &     Parameter &         Value \\
        \midrule
        Specification      & Max steps         & 100       & Max steps         & 100   & Max steps & 100    & Max steps         & 150       & Max steps         & 150\\
        {}      & Pose th.      & .175      & Pose th.      & .35  & Frame skip    &  5         & Pos. th.       & $\infty$       & Pos. th.       & .1\\
        {}      & Target dist.   & 1         &               &       & {}            & {}        & Rot. th.       & .262      & Rot. th.       & $\infty$ \\
        {}      &               &           &               &       & {}            & {}        & Goal pos.     & (0, 0)    & Qpos noise    & .01\\
        {}      &               &           &               &       & {}            & {}        & Goal rot.     & $-.785\sim .785$ & Tgt. xyz  & $(0,-.45,.9)\sim(.3, -.1, 1.05)$ \\
        {}      &               &           &               &       & {}            & {}        & Frame skip    & 5         & Tgt. rpy      & $(-.2, -.2, -.2)\sim(.2, .2, .2)$ \\
        {}      &               &           &               &       & {}            & {}        &               &           & Obj. xyz      & $(-.1, -.35, 1)\sim(.1, -.15, 1)$ \\
        {}      &               &           &               &       & {}            & {}        &               &           & Obj. geom.    & $(.015, .015, .015)\sim(.025, .025, .025)$ \\
        {}      &               &           &               &       & {}            & {}        &               &           & Obj. mass     & $.05\sim .2$ \\
        {}      &               &           &               &       & {}            & {}        &               &           & Obj. fric.    & $(.8,.004,.00008)\sim(1.2,.006,.00012)$ \\
        \midrule
        Reward  & Pose          & 1     & Reach         & 1     & Pos. align diff& 100           & Pos. dist.     & 2     & Solved        & 20 \\
                & Solved        & 1     & Solved        & 1     & Rot. align diff& 100   & Rot. dist.     & .2    & Pos. dist.     & 10 \\
                & Action reg.   & 1     & Action reg.   & 1     & Alive         & 1     & Pos. align diff& 100   & Palm dist.    & 1\\
                & {}            & {}    &               &       & Solved        & 1     & Rot. align diff& 10    & Tip dist.     & .1\\
                & {}            & {}    &               &       &               &       & Alive         & 1 \\
                & {}            & {}    &               &       &               &       & Solved        & 2 \\
        \bottomrule
        \end{tabular}
    }
    \caption{\textbf{Task and reward parameters.} Parameters of \emph{Elbow pose}, \emph{Finger reach}, \emph{Pen reorient}, \emph{Die reorient} and \emph{Object relocation}. The notations \emph{th, geom, fric, dist, pos, rot, reg, tgt, obj} are abbreviations for threshold, geometry, friction, distance, position, rotation, regularization, target and object.}
    \label{tab:env_params_1}
\end{table}

\begin{table}[H]
    \centering
    \resizebox{\textwidth}{!}{
        \begin{tabular}{lllllllllll}
        \toprule
        Task & \multicolumn{2}{l}{Baoding CW} & \multicolumn{2}{l}{Baoding CCW} & \multicolumn{2}{l}{Baoding hard} & \multicolumn{2}{l}{Baoding harder} & \multicolumn{2}{l}{Walk to point} \\
        {} &       Parameter &        Value &       Parameter &         Value &       Parameter &          Value &      Parameter &         Value   & Parameter & Value \\
        \midrule
        Specification      & Max steps         & 200       & Max steps         & 200   & Max steps & 200   & Max steps         & 200        & Max steps         & 150 \\
        {}      & Goal period       & 5         & Goal period       & 5         & Goal period       & $4\sim 6$     & Goal period       & $4\sim 6$     &       & \\
        {}      & Init. phase       & 1.57      & Init. phase       & 1.57      & Init. phase  & 1.57          & Init. phase        & 1.57          &             &    \\
        {}      & Lim. init. angle  & 0         & Lim. init. angle  & 0         & Lim. init. angle  & 0             & Lim. init. angle  & 3.14             \\
        {}      &                   &           &                   &           & Goal x            & $.02\sim.03$              & Goal x            & $.02\sim.03$            &  \\
        {}      &                   &           &                   &           & Goal y            & $.022\sim.032$            & Goal y            & $.022\sim.032$          &  \\
        {}      &                   &           &                   &           & Obj. size         & $.018\sim.024$            & Obj. size         & $.018\sim.024$          &  \\
        {}      &                   &           &                   &           & Obj. mass         & $.03\sim.3$               & Obj. mass         & $.03\sim.3$             &  \\
        {}      &                   &           &                   &           & Obj. fric.        & $(.2, .001, .00002)$      & Obj. fric.        & $(.2, .001, .00002)$    &  \\
        \midrule
        Reward  & Pos. dist. 1    & 1     & Pos. dist. 1    & 1     & Pos. dist. 1   & 1              & Pos. dist. 1    & 1     & Pos. dist. & .8     \\
                & Pos. dist. 2    & 1     & Pos. dist. 2    & 1     & Pos. dist. 2    & 1     & Pos. dist. 2    & 1     & W-upright     & .1\\
                & Alive          & 1     & Alive          & 1     & Alive          & 1     & Alive          & 1     & K-energy      & .05\\
                & Solved         & 5     & Solved         & 5     & Solved         & 5     & Solved         & 5     &               W-energy & .1 \\
                & & & & & & & & & Solved & 1 \\
        \bottomrule
        \end{tabular}
    }
    \caption{\textbf{Task and reward parameters (continued).} Task and reward parameters of \emph{Baoding CW}, \emph{Baoding CCW}, \emph{Baoding hard}, \emph{Baoding harder} and \emph{Walk to point}. The notations \emph{lim, init, obj, fric, dist} and \emph{pos} are abbreviations for limit, initialization, object, friction, distance and position.}
    \label{tab:env_params_2}
\end{table}

\subsection*{Muscle transformer architecture}

Arnold's architecture consists of a transformer encoder-decoder network. While the encoder network is shared, two separate decoders output the action and the state value (Table~\ref{tab:hyperparameters}).

Each sensory modality (e.g., muscle length and velocity, object position and orientation) is represented by a short time series (six time steps), mapped into a sensory embedding of a fixed size by a linear encoder.

We found that longer time horizons did not convey an advantage over this simple approach. Specifically, we tested modified versions of Arnold (Figure~\ref{fig:bilateral}), where an additional layer of attention over time is added to the original model, allowing the model to process a history of $96$ time steps. The results showed that, although the modified version did better on some of the tasks for randomized experts (Gaussian noise is added to expert actions), it is not comparable with the original method when the noise is not added. Our conjecture is that with a larger history, the model can learn to better handle noisy input, resulting in better performance compared to the original network; when  noise is absent, the added attention layer does not provide any extra benefit compared to the original short horizon implementation.

\begin{table}[h]
\vspace{10pt}
\centering
\begin{tabular}{lccc}
\toprule
\textbf{Hyperparameter}       & \textbf{Encoder} & \textbf{Action decoder} & \textbf{Value decoder} \\
\midrule
Embedding size                & 128              & 128                     & 128                    \\
Feedforward dimension         & 512              & 512                     & 512                    \\
Number of heads               & 4                & 4                       & 4                      \\
Number of layers              & 6                & 6                       & 6                      \\
Dropout                       & 0                & 0                       & 0                      \\
Activation                    & ReLU             & ReLU                    & ReLU                   \\
LayerNorm $\varepsilon$       & $10^{-5}$        & $10^{-5}$               & $10^{-5}$              \\
Pre-norm (norm\_first)        & True             & True                    & True                   \\
\midrule
\#Parameters                  & 1,189,888        & 1,587,712               & 1,587,712              \\
\midrule
\textbf{Additional components} & \multicolumn{3}{c}{} \\
\midrule
Tokenizer (Linear) $\times 3$ & \multicolumn{3}{c}{2,688} \\
Positional encoder (Embedding) $\times 3$ & \multicolumn{3}{c}{82,176} \\
Action net                    & \multicolumn{3}{c}{129} \\
Value net                     & \multicolumn{3}{c}{129} \\
Log std net                   & \multicolumn{3}{c}{129} \\
Log std (scalar)              & \multicolumn{3}{c}{1} \\
Total additional parameters   & \multicolumn{3}{c}{85,252} \\
\midrule
\textbf{Total parameters}     & \multicolumn{3}{c}{4,450,564} \\
\bottomrule
\end{tabular}
\caption{\textbf{Model hyperparameters and parameter counts.} Hyperparameters and parameter counts for the Arnold network}
\label{tab:hyperparameters}
\end{table}

When deploying the muscle transformer with an actor-critic algorithm like PPO~\citep{schulman2017proximal}, the decoder network also receives as input a \emph{value embedding}, which is associated with a scalar output. We remark that this network architecture can accept any number of input sensory modalities and output a control signal for any number of actuators, making it suitable for multi-task learning of different embodiments and environments (Figure~\ref{fig:overview}). We detail the network's hyperparameters in Table~\ref{tab:hyperparameters}. 

We note that we did not implement Q-value estimation, because it is not required by PPO, which we used in our experiments. However, we hypothesize that it could be implemented in a similar way, either by running the encoder with additional action input (corresponding to the action whose Q-value needs to be estimated), or directly feeding it to the decoder. In discrete action spaces, a method such as Q-learning could be implemented by providing tokens for each admissible action as input to the decoder.

\subsection*{Sensorimotor vocabulary}

Transformer networks are permutation equivariant without positional encoding, meaning that the order of the input vectors does not change the output values.
Typically, positional encoding is used to break positional invariance~\citep{vaswani2017attention,devlin2019bert,shaw2018selfattention,su2023roformer}.
In our setting, the observation values denote sensory elements which vary in number and ordering across tasks (Figure~\ref{fig:overview}). The policy network needs to distinguish between these elements, which represent different sensory modalities and anatomical locations. To this end, we propose the use of a compositional token embedding strategy to encode the identity of the data source. As highlighted in the main text, we created a \emph{sensorimotor vocabulary} (Figure~\ref{fig:architecture}) that is organized into semantically meaningful categories (Figure~\ref{fig:vocabulary_explained}). The vocabulary is used in the \emph{sensory encoder} and the \emph{actuator encoder}, where the embedding for any sensorimotor input is calculated as the sum of the embeddings of all the tokens involved in the phrase that describes that input. The input time series is projected linearly and added to the embeddings, as illustrated in Figure~\ref{fig:architecture}A.

\subsection*{Training details}
\label{app:training_details}

All training experiments were carried out on a GPU cluster featuring A100 GPUs. We ran two training experiments in parallel on a single node, to maximally utilize the GPU memory. 
\begin{itemize}
    \item We ran PPO, BC, OBC and OBC-PPO using the same network architecture and hyperparameters (Table~\ref{tab:training_details}). 
    \item OBC-PPO combines imitation loss (L2), the surrogate policy gradient loss of PPO, the value loss, and an entropy loss to encourage exploration. 
    \item PPO, instead, sets the imitation loss coefficient to 0, while OBC sets the policy gradient and entropy coefficients to 0. 
    \item For BC, we used the same setup as OBC, but collected the rollouts using the expert policy's actions.
    \item For MT-PPO and MT-SAC, we used the union observation space of all 14 tasks, with 0 paddings and an additional task index input to the policy. The policy is a MLP network with 2 hidden layers of size 512.
    \item For PPO + T, we used the same transformer policy and pure policy-gradient setup as PPO. Instead of the compositional sensorimotor vocabulary, each observation and action element was assigned its own independent, learnable token embedding (the ``atomic'' vocabulary).
    \item For OBC + Task-SV, we used the same transformer policy and imitation setup as OBC. The sensorimotor vocabulary was made task-specific; each of the 14 tasks was associated with an independent set of learnable token embeddings, removing all cross-task sharing of the sensorimotor representation.
\end{itemize}

All training runs consisted of 50M steps (environment interactions) at fixed learning rate unless specified otherwise. The policy gradient objective does not pair well with a large learning rate, thus we reduced it for PPO. After the first 50M steps, we ran BC, OBC and OBC-PPO for an additional 5M steps with a smaller learning rate, which boosted the performance in certain tasks. We did not achieve the same improvement with PPO. Due to the low performance of MT-SAC~\citep{haarnoja2018soft} after 20M steps, we did not continue the training. All main experiments were performed for five different random seeds. Ablation experiments (MT-PPO, MT-SAC, PPO + T, and OBC + Task-SV) were performed for three different random seeds. One full PPO run required approximately 4 days, while a run of OBC, OBC-PPO or BC required approximately 5 days. The overhead was due to the computation of the expert actions. 

\begin{table}[ht!]
\vspace{5pt}
\centering
\resizebox{0.8\textwidth}{!}{
\begin{tabular}{lccccc}
\toprule
\textbf{Hyperparameter} & \textbf{PPO} & \textbf{BC} & \textbf{OBC} & \textbf{OBC-PPO} & \textbf{RL fine-tuning} \\
\midrule
Initial standard deviation & 1.0 & 1.0 & 1.0 & 1.0 & 1e-3 \\
Batch size & 128 & 128 & 128 & 128 & 128 \\
Entropy coefficient & 1e-6 & 0.0 & 0.0 & 1e-6 & 1e-6 \\
Value function coefficient & 0.5 & 0.5 & 0.5 & 0.5 & 0.5 \\
Policy gradient coefficient & 1.0 & 0.0 & 0.0 & 1.0 & 1.0 \\
Imitation coefficient & 0.0 & 1.0 & 1.0 & 1.0 & 0.0 \\
Initial learning rate & 2e-5 & 1e-3 & 1e-3 & 1e-3 & 2e-6 \\
Reduced learning rate & n/a & 1e-5 & 1e-5 & 1e-5 & n/a \\
Rollout steps & 512 & 512 & 512 & 512 & 512 \\
Training epochs & 3 & 3 & 3 & 3 & 3 \\
Expert rollout & n/a & Yes & No & No & n/a \\
Discount factor (gamma) & 0.99 & 0.99 & 0.99 & 0.99 & 0.99 \\
GAE lambda & 0.95 & 0.95 & 0.95 & 0.95 & 0.95 \\
Clip range & 0.2 & 0.2 & 0.2 & 0.2 & 0.2 \\
Max gradient norm & 0.5 & 0.5 & 0.5 & 0.5 & 0.5 \\
Standardize advantage & True & True & True & True & True \\
Standardize observations & True & True & True & True & True \\
Imitation loss & n/a & MSE & MSE & MSE & n/a \\
\bottomrule
\end{tabular}
}
\caption{\textbf{Hyperparameters for different training methods.} Hyperparameters for PPO, BC, OBC, OBC-PPO, RL fine-tuning with the Arnold network.}
\label{tab:training_details}
\end{table}

\subsection*{Multi-task learning}

Training a policy across diverse environments poses unique challenges, requiring a flexible architecture and careful handling of observations, rewards, and action variability. Below, we outline the key components of our approach.

\textbf{Experience collection.} 
To expose the policy to diverse transitions at each training step, we execute all tasks in parallel (Figure~\ref{fig:overview}B: Step 1). Each task is assigned to a separate process, which returns one observation per step. Observations are padded to enable batching, passed through the policy, and the resulting actions are routed back to their respective environments. Transitions are stored in a shared rollout buffer, containing exclusively data generated by the current policy.

\textbf{Resource allocation.}
Some tasks are more challenging than others. Therefore, the policy might require relatively more environment interactions to learn those tasks. To address this, we allocated more environment instances to challenging tasks during experience collection (Table~\ref{tab:env_counts}). The relative difficulty of a task was assessed based on the training steps required for single-task imitation learning (Figure~\ref{fig:single_task_lc}).

\textbf{Per-component observation standardization.}
Standardizing inputs (zero mean, unit variance) often accelerates learning. In RL, this requires maintaining running statistics over observed states, as no fixed dataset exists. In a multi-task setting, the meaning of observation components varies across tasks, so we maintain normalization statistics per \emph{observation signature}, i.e., a descriptor of the sensory modality. For example, all ["right", "soleus", "muscle", "length"] inputs share the same statistics regardless of task. This method improves both learning efficiency and final performance (Figures~\ref{fig:ablations} and~\ref{fig:performance}C), while preventing the number of standardization coefficients from increasing excessively.

\textbf{Reward standardization.}
Different reward ranges across tasks complicate learning a unified value function. We normalize returns per task by using task-specific mean and standard deviation of cumulative rewards, thus ensuring stable value targets. While a task-specific reward standardization requires task identifiers at training time, the policy remains agnostic to task identity during deployment, where rewards are not provided. Removing this normalization significantly impairs PPO training (Figure~\ref{fig:ablations}). Although OBC does not require a value function, learning it facilitates subsequent fine-tuning with PPO.

\textbf{Task-specific exploratory noise.}
PPO uses a stochastic policy where each action component is sampled from a Gaussian distribution~\citep{schulman2017proximal}. Typically, standard deviations are learned via backpropagation, to adapt their magnitude to the sensitivity of each component to noise and to the component's scale. However, in multi-task settings, the action semantics vary. To address this, the policy outputs each action component's standard deviation as $\bm{\sigma} = \tilde{\sigma}\,\operatorname{softmax}(\mathbf{x}^\top E)N_A$, where $\tilde{\sigma} \in \mathbb{R}$ is a global learnable scalar, $E \in \mathbb{R}^{N_T \times N_A}$ are the transformer decoder embeddings and $\mathbf{x} \in \mathbb{R}^{N_T}$ maps each embedding to a noise magnitude. Here, $N_T$ denotes the embedding size and $N_A$ the number of actuators. This formulation allows the policy to scale exploratory noise per component according to the task-specific action semantics.

\subsection*{RL fine-tuning and self-distillation}
We used a policy trained with OBC as a starting point and performed additional RL training individually per task (Figure~\ref{fig:overview}B Step 2). This fine-tuning phase generated multiple new policies, some featuring super-expert performance in specific tasks (Figure~\ref{fig:rl_fine_tuning}). We used these new policies as experts for multi-task learning with OBC, repeating the previous distillation procedure into the original Arnold policy, thus achieving overall super-expert performance with a single policy (Figure~\ref{fig:overview}B Step 3).

For the fine-tuning phase of Arnold, we resumed the training of OBC after 55M steps, replacing the imitation loss with the policy gradient and entropy losses (the value loss was active at all times). In the fine-tuning experiments, OBC was trained on a single task at a time, generating one new expert per task. Fine-tuning improved the performance compared to the expert in \emph{Walk to point}, \emph{Elbow pose}, \emph{Baoding hard}, \emph{Baoding harder}, and \emph{Die reorient} (Figure~\ref{fig:rl_fine_tuning}). Crucially, in order to achieve a performance improvement we had to reset the standard deviation of the action distribution to $10^{-3}$ (during the pretraining phase it had approached 0) and reduce the learning rate to $2\times 10^{-6}$. We then replaced the previous experts with these improved versions and ran additional 10M steps of OBC, leading to the final Arnold policy.

\subsection*{Smoothness metrics}
\label{smoothness_metrics}
For each policy and task we collected 100 rollout episodes, and from every episode extracted two signals: the per-step muscle activations applied by the policy (the actions, $a_t$), and the trajectory of the 23 hand joint angles, $q_t$, recovered from the recorded observation tensor and truncated to the executed episode length.
From these we computed four complementary smoothness measures, with all temporal derivatives obtained by finite differencing at unit timestep (one simulation step).
The first, the mean activation increment ($\overline{|\Delta a|}$), is defined as the mean absolute step-to-step change in muscle activation, $\langle |a_{t+1}-a_t| \rangle$, averaged over all muscles and time steps; lower values indicate a less jittery control signal.
The remaining three are standard kinematic smoothness metrics applied to the joint-angle trajectories.
The spectral arc length (SPARC)~\citep{balasubramanian2015analysis} was computed on the joint-velocity magnitude $\|\dot q_t\|_2$: we took the normalized magnitude spectrum of its zero-padded Fourier transform, restricted to frequencies up to a 10 Hz adaptive cut-off (trimming the trailing tail below 5\% of peak amplitude), and summed the arc length of the resulting curve; SPARC is negative, with values closer to zero denoting smoother movement.
The log dimensionless jerk (LDLJ) was computed per joint as $-\ln\big(\tfrac{T^5}{q_{\text{peak}}^2}\int \dddot q^2 dt \big)$, where $T$ is the movement duration, $\dddot q$ the third time derivative of the joint angle, and $q_{\text{peak}}$ the peak excursion of that joint about its mean~\citep{hogan2009sensitivity}; this dimensionless quantity penalizes large, sustained jerk while normalizing for amplitude and duration, and was averaged across joints (excluding near-static joints to avoid divergence), with higher (less negative) values indicating smoother motion.
Finally, the number of velocity peaks (NVP) quantifies movement segmentation~\citep{rohrer2002movement}: for each joint we counted local maxima in the angular speed $|\dot q_t|$ exceeding 5\% of that joint's peak speed and averaged the count across joints. Each metric was computed per episode and then aggregated to a per-task mean before comparison across policy types.

\subsection*{Comparison of simulated and human muscle activations}
To assess how closely each policy's muscle activations resemble human electromyography (EMG) during walking, we compared simulated activation profiles against EMG recorded from nine human subjects (five male, four female) walking at 4.5 km/h. EMG recordings were acquired from a publicly available dataset~\citep{wang2023wearable}. Gait cycles were segmented from each policy's rollouts at left heel strike, detected as the rising edge of the low-pass-filtered (5 Hz) left-foot contact force crossing an absolute threshold; a cycle was retained only if a right heel strike occurred between two successive left strikes (enforcing a left–right–left stance pattern), and cycles falling outside the central 80\% of the duration distribution or with a mean forward velocity below 0.2 m/s were discarded. The retained cycles were low-pass filtered (15 Hz), resampled to a common time base, and averaged across cycles to yield a mean activation profile per muscle for each policy (Arnold, OBC, and Kinesis). Nine simulated muscles were mapped to their human EMG counterparts (soleus, medial and lateral gastrocnemius, tibialis anterior, semimembranosus, biceps femoris, and vastus lateralis, vastus medialis and rectus femoris), and the mean simulated profiles were linearly resampled onto the human time grid. For each muscle we computed the Pearson correlation between the simulated profile and each subject's averaged EMG, giving a human-to-simulation correlation per subject and muscle. As a reference ceiling, we computed cross-human correlations between every pair of subjects (leave-one-out: each subject versus the mean of the other eight), quantifying the intrinsic inter-subject variability of human EMG.

To examine whether the overall control strategy beyond individual muscle activations is compatible with reported human motor strategies, we applied factor analysis to Arnold's outputs, following~\citeauthor{ivanenko2004five}~\citep{ivanenko2004five}. We restricted the muscle set to the 14 MyoLeg muscles with direct counterparts in ~\citeauthor{ivanenko2004five}'s recording set, extracted gait cycles from Arnold's Walk-to-point rollouts using right-foot heel-strike events, time-normalized to a 200-point gait cycle, and applied varimax-rotated PCA with an eigenvalue $\geq$ 0.5 retention criterion.
The Bartlett test of sphericity was highly significant (df = 91, $\chi^2 = 8082$, $p < 10^{-6}$) and the Kaiser-Meyer-Olkin measure was 0.554, within the reported range for human EMG. Four factors satisfied the eigenvalue $\geq$ 0.5 criterion, jointly explaining 94.7\% of the variance in muscle activations across the gait cycle.

\subsection*{Muscle synergy analysis and control subspace inactivation (CSI)}
\label{pca_analysis}

To investigate whether Arnold learned a strategy compatible with the muscle-synergies hypothesis, we performed control subspace inactivation (CSI) analysis~(Figure~\ref{fig:ev_v_csi}). CSI was developed in recent work by us~\citep{chiappa2024acquiring}. Specifically, we collected muscle activations generated by Arnold while solving the 11 tasks involving the MyoHand model, and we used this dataset to find the principal components of the muscle activation signal. We performed principal component analysis (PCA) both on the muscle activations of the 11 tasks together and on the activations of individual tasks, calculating the cumulative explained variance (EV) achieved by the task-specific and combined PC subspaces. We additionally performed non-negative matrix factorization (NMF, Figure~\ref{fig:ev_v_csi_nmf}) under the same task-specific and combined settings, using reconstruction R² as the quality metric; NMF enforces non-negativity and thus more directly operationalizes the additive-combination assumption of the synergy hypothesis. We then deployed Arnold on 100 episodes per task, projecting the control signal on a reduced subspace, spanned by the N most important components for N up to 39, the number of muscles of MyoHand. We compared the average relative performance that Arnold achieved under these projections while using the task-specific PCs versus the combined PCs (Figure~\ref{fig:ev_v_csi}). Consistent with~\citet{chiappa2024acquiring}, we found that an analysis using EV tends to underestimate the control space dimensionality for individual tasks. Furthermore, we extended this finding to the combined-task case.

Beyond this difference, both EV-based and CSI-based analyses showed an increase in dimensionality under the combined-task components compared to the task-specific components. To compare the similarity between pairs of subspaces, we used Projected Variance Difference (PVD) and Principal Angle Difference (PAD) measures, which we adopted from~\citet{todorov2004analysis}.

One potential limitation of CSI is that due to the perturbed control command, the policy might be pushed off its training distribution and thus generalize poorly. This lack of transferability could thus be the bottleneck rather than the (constrained) subspaces per se. To address this concern, we developed a CSI fine-tuning analysis to investigate the functional dimensionality of the learned control policies. We did this experiment with specialist MLP policies per task, as training Arnold for 3 seeds on 14 tasks while constraining it to subspaces of varying dimensionality would be even more computationally prohibitive. After training an MLP policy (2 hidden layers, 256 dimensional hidden space) on each task (N=14), we constrained its action space to a learnable low-dimensional subspace and continued training using two fine-tuning strategies: reinforcement learning (RL) and on-policy behavior cloning (OBC). We trained three seeds for 5M steps (Figure~\ref{fig:csi_learning_curves}). This analysis reveals whether the control policy truly operates in a low-dimensional manifold and provides complementary evidence to the PCA-based synergy analysis. Specifically, if a policy can regain high performance under CSI fine-tuning with an action dimensionality smaller than that suggested by PCA (which only constrains the dimensionality without further fine-tuning), it indicates that the model may not necessarily learn a globally low-dimensional control strategy.

Our results show that, for some tasks (notably \emph{Baoding CCW} and \emph{Baoding CW}), CSI fine-tuning with RL achieved successful task performance at lower action dimensionalities than predicted by PCA. However, in most tasks, RL-based fine-tuning destabilized the policy and led to degraded performance. In contrast, OBC fine-tuning generally maintained performance comparable to PCA-constrained results, though the same \emph{Baoding CCW} and \emph{Baoding CW} tasks again deviated from PCA predictions. Both results show task-specific differences in control structure.

\subsection*{Statistical analysis}

All analyses were performed in Python using scipy.stats~\citep{2020SciPy-NMeth}. Group comparisons used non-parametric tests, except where specified otherwise. Summary values are reported as mean $\pm$ s.e.m. unless stated otherwise.

\textbf{Performance metric and normalization.} For each policy and task, performance is the solved-step fraction — the fraction of timesteps in an episode during which the task's "solved" criterion was met — evaluated over 200 stochastic rollouts per task. Each method's per-task performance is expressed relative to that task's specialist expert (the per-task teacher): relative performance (\%) = (method mean ÷ expert mean) × 100. Values can exceed 100\% when the student policy outperforms the single-task expert.

\textbf{Ablation study (Tables~\ref{tab:ablations}, \ref{tab:ablations_per_task}).} For each task column, relative performance was aggregated across seeds. Each cell reports the mean $\pm$ s.e.m. For task families, because every task contributes the same number of episodes (200), the values represent the unweighted mean of per-task relative means. The Avg column pools all 14 tasks. Overall performance was compared to full Arnold with a two-sided Wilcoxon signed-rank test with Holm--Bonferroni correction across methods, pairing each method's per-task relative performance against Arnold's (statistical metrics in Table~\ref{tab:ablation-significance}).

\textbf{Multi-task versus expert performance (Figure~\ref{fig:performance}B and~Table \ref{tab:arnold-expert-significance}).} The \emph{relative performance} of each policy was compared against the single-task expert baselines. The overall improvement over experts was tested using a two-sided Wilcoxon signed-rank test with * p < 0.05. We also studied per-task improvement and reported the p values from Holm-Bonferroni-corrected one-sided one-sample t-test (Table~\ref{tab:arnold-expert-significance}).

\textbf{Action and kinematic smoothness (Figure~\ref{fig:smoothness_biomech}A).} For each of the four smoothness metrics (mean |$\Delta$a|, SPARC, LDLJ, number of velocity peaks), the three policies (per-task expert, single-task OBC, multi-task OBC) were compared with a two-sided task-paired Wilcoxon signed-rank test over the n = 11 MyoHand tasks, Holm-Bonferroni corrected across policies, using the per-task mean (100 episodes per task) as the paired unit (* p < 0.05, ** p < 0.01, statistical metrics in Table~\ref{tab:smoothness-significance}).

\textbf{Correlation with human EMG (Figure~\ref{fig:smoothness_biomech}B).} For each policy (Arnold, OBC, Kinesis expert), the Pearson correlation between its mean gait-cycle activation and each human subject's EMG was computed separately for each of the nine analyzed muscles, then averaged across muscles to yield one correlation per subject per policy (N = 9 subjects). Correlations were Fisher-z-transformed, and the three pairwise comparisons between policies (Arnold vs. OBC, Arnold vs. Kinesis, OBC vs. Kinesis) were assessed with a two-sided Wilcoxon signed-rank test paired across the nine subjects, Bonferroni-corrected over the three comparisons (* p < 0.05). The cross-human reference distribution (cross-human leave-one-out comparison) is shown for reference and was not subjected to statistical comparison.

\textbf{Subspace-similarity curves (PVD/PAD).} Projected Variance Difference and Principal Angle Difference are reported as descriptive measures. Shaded regions denote mean $\pm$ standard deviation, where the spread is taken across the set of task-pair comparisons at each number of PCA components.

\newpage

\section*{References}
{

\begin{thebibliography}{112}
\providecommand{\natexlab}[1]{#1}
\providecommand{\url}[1]{\texttt{#1}}
\expandafter\ifx\csname urlstyle\endcsname\relax
  \providecommand{\doi}[1]{doi: #1}\else
  \providecommand{\doi}{doi: \begingroup \urlstyle{rm}\Url}\fi

\bibitem[Tassa et~al.(2018)Tassa, Doron, Muldal, Erez, Li, Casas, Budden,
  Abdolmaleki, Merel, Lefrancq, et~al.]{tassa2018deepmind}
Yuval Tassa, Yotam Doron, Alistair Muldal, Tom Erez, Yazhe Li, Diego de~Las
  Casas, David Budden, Abbas Abdolmaleki, Josh Merel, Andrew Lefrancq, et~al.
\newblock Deepmind control suite.
\newblock \emph{arXiv preprint arXiv:1801.00690}, 2018.

\bibitem[Makoviychuk et~al.(2021)Makoviychuk, Wawrzyniak, Guo, Lu, Storey,
  Macklin, Hoeller, Rudin, Allshire, Handa, and State]{makoviychuk2021isaac}
Viktor Makoviychuk, Lukasz Wawrzyniak, Yunrong Guo, Michelle Lu, Kier Storey,
  Miles Macklin, David Hoeller, Nikita Rudin, Arthur Allshire, Ankur Handa, and
  Gavriel State.
\newblock Isaac gym: High performance gpu-based physics simulation for robot
  learning.
\newblock \emph{arXiv preprint arXiv:2108.10470}, 2021.

\bibitem[Jacobsen et~al.(1986)Jacobsen, Iversen, Knutti, Johnson, and
  Biggers]{jacobsen1986design}
S.~Jacobsen, E.~Iversen, D.~Knutti, R.~Johnson, and K.~Biggers.
\newblock Design of the {{Utah}}/{{M}}.{{I}}.{{T}}. {{Dextrous Hand}}.
\newblock In \emph{1986 {{IEEE International Conference}} on {{Robotics}} and
  {{Automation Proceedings}}}, volume~3, pages 1520--1532, April 1986.
\newblock \doi{10.1109/ROBOT.1986.1087395}.

\bibitem[Kumar et~al.(2013)Kumar, Xu, and Todorov]{kumar2013fast}
Vikash Kumar, Zhe Xu, and Emanuel Todorov.
\newblock Fast, strong and compliant pneumatic actuation for dexterous
  tendon-driven hands.
\newblock In \emph{2013 IEEE International Conference on Robotics and
  Automation}, pages 1512--1519, 2013.
\newblock \doi{10.1109/ICRA.2013.6630771}.

\bibitem[Yasa et~al.(2023)Yasa, Toshimitsu, Michelis, Jones, Filippi, Buchner,
  and Katzschmann]{yasa2023overview}
Oncay Yasa, Yasunori Toshimitsu, Mike~Y. Michelis, Lewis~S. Jones, Miriam
  Filippi, Thomas Buchner, and Robert~K. Katzschmann.
\newblock An {{Overview}} of {{Soft Robotics}}.
\newblock \emph{Annual Review of Control, Robotics, and Autonomous Systems},
  6\penalty0 (Volume 6, 2023):\penalty0 1--29, May 2023.
\newblock ISSN 2573-5144.
\newblock \doi{10.1146/annurev-control-062322-100607}.

\bibitem[Shaw et~al.(2023)Shaw, Agarwal, and Pathak]{shaw2023leap}
Kenneth Shaw, Ananye Agarwal, and Deepak Pathak.
\newblock {LEAP Hand: Low-cost, efficient, and anthropomorphic hand for robot
  learning}.
\newblock \emph{arXiv preprint arXiv:2309.06440}, 2023.

\bibitem[all()]{allegro}
Allegro {{Hand}} {\textbar} robot hand.
\newblock https://www.allegrohand.com.

\bibitem[Christoph et~al.(2025)Christoph, Eberlein, Katsimalis, Roberti,
  Sympetheros, Vogt, Liconti, Yang, Cangan, Hinchet, and
  Katzschmann]{christoph2025orca}
Clemens~C. Christoph, Maximilian Eberlein, Filippos Katsimalis, Arturo Roberti,
  Aristotelis Sympetheros, Michel~R. Vogt, Davide Liconti, Chenyu Yang,
  Barnabas~Gavin Cangan, Ronan~J. Hinchet, and Robert~K. Katzschmann.
\newblock Orca: An open-source, reliable, cost-effective, anthropomorphic
  robotic hand for uninterrupted dexterous task learning.
\newblock \emph{arXiv preprint arXiv:2504.04259}, 2025.

\bibitem[Zhang et~al.(2019)Zhang, Sheng, O’Neill, Walsh, Wood, Ryu, Desai,
  and Yip]{zhang2019robotic}
Jun Zhang, Jun Sheng, Ciar{\'a}n~T O’Neill, Conor~J Walsh, Robert~J Wood,
  Jee-Hwan Ryu, Jaydev~P Desai, and Michael~C Yip.
\newblock Robotic artificial muscles: Current progress and future perspectives.
\newblock \emph{IEEE transactions on robotics}, 35\penalty0 (3):\penalty0
  761--781, 2019.

\bibitem[Hansen et~al.(2022)Hansen, Wang, and Su]{hansen2022temporal}
Nicklas Hansen, Xiaolong Wang, and Hao Su.
\newblock Temporal difference learning for model predictive control.
\newblock \emph{arXiv preprint arXiv:2203.04955}, 2022.

\bibitem[Wochner et~al.(2023)Wochner, Schumacher, Martius, B{\"u}chler,
  Schmitt, and Haeufle]{wochner2023learning}
Isabell Wochner, Pierre Schumacher, Georg Martius, Dieter B{\"u}chler, Syn
  Schmitt, and Daniel Haeufle.
\newblock Learning with {{Muscles}}: {{Benefits}} for {{Data-Efficiency}} and
  {{Robustness}} in {{Anthropomorphic Tasks}}.
\newblock In \emph{Proceedings of {{The}} 6th {{Conference}} on {{Robot
  Learning}}}, pages 1178--1188. PMLR, March 2023.

\bibitem[{Marin Vargas, Alessandro} et~al.(2024){Marin Vargas, Alessandro},
  Bisi, Chiappa, Versteeg, Miller, and Mathis]{vargas2024taskdriven}
{Marin Vargas, Alessandro}, Axel Bisi, Alberto~S. Chiappa, Chris Versteeg,
  Lee~E. Miller, and Alexander Mathis.
\newblock Task-driven neural network models predict neural dynamics of
  proprioception.
\newblock \emph{Cell}, 187\penalty0 (7):\penalty0 1745--1761.e19, March 2024.
\newblock ISSN 0092-8674, 1097-4172.
\newblock \doi{10.1016/j.cell.2024.02.036}.

\bibitem[Chiappa et~al.(2024{\natexlab{a}})Chiappa, Tano, Patel, Ingster,
  Pouget, and Mathis]{chiappa2024acquiring}
Alberto~Silvio Chiappa, Pablo Tano, Nisheet Patel, Abiga{\"i}l Ingster,
  Alexandre Pouget, and Alexander Mathis.
\newblock Acquiring musculoskeletal skills with curriculum-based reinforcement
  learning.
\newblock \emph{Neuron}, 112\penalty0 (23):\penalty0 3969--3983.e5, December
  2024{\natexlab{a}}.
\newblock ISSN 0896-6273.
\newblock \doi{10.1016/j.neuron.2024.09.002}.

\bibitem[Perez~Rotondo et~al.(2025)Perez~Rotondo, Simos, David, Pigeon, Blanke,
  and Mathis]{perez2025deep}
Adriana Perez~Rotondo, Merkourios Simos, Florian David, Sebastian Pigeon, Olaf
  Blanke, and Alexander Mathis.
\newblock Deep-learning models of the ascending proprioceptive pathway are
  subject to illusions.
\newblock \emph{Experimental Physiology}, 2025.

\bibitem[Delp et~al.(2007)Delp, Anderson, Arnold, Loan, Habib, John,
  Guendelman, and Thelen]{delp2007opensim}
Scott~L Delp, Frank~C Anderson, Allison~S Arnold, Peter Loan, Ayman Habib,
  Chand~T John, Eran Guendelman, and Darryl~G Thelen.
\newblock {{OpenSim}}: Open-source software to create and analyze dynamic
  simulations of movement.
\newblock \emph{IEEE transactions on biomedical engineering}, 54\penalty0
  (11):\penalty0 1940--1950, 2007.

\bibitem[Seth et~al.(2011)Seth, Sherman, Reinbolt, and Delp]{seth2011opensim}
Ajay Seth, Michael Sherman, Jeffrey~A Reinbolt, and Scott~L Delp.
\newblock {{OpenSim}}: A musculoskeletal modeling and simulation framework for
  in silico investigations and exchange.
\newblock \emph{Procedia Iutam}, 2:\penalty0 212--232, 2011.

\bibitem[Geijtenbeek(2019)]{geijtenbeek2019scone}
Thomas Geijtenbeek.
\newblock Scone: Open source software for predictive simulation of biological
  motion.
\newblock \emph{Journal of Open Source Software}, 4\penalty0 (38):\penalty0
  1421, 2019.

\bibitem[Rasmussen et~al.(2002)Rasmussen, Vondrak, Damsgaard, De~Zee,
  Christensen, and Dostal]{rasmussen2002anybody}
John Rasmussen, Vit Vondrak, Michael Damsgaard, Mark De~Zee, S{\o}ren~T
  Christensen, and Zdenek Dostal.
\newblock The anybody project--computer analysis of the human body.
\newblock \emph{Biomechanics of Man}, 1, 2002.

\bibitem[Caggiano et~al.(2022{\natexlab{a}})Caggiano, Wang, Durandau, Sartori,
  and Kumar]{caggiano2022myosuite}
Vittorio Caggiano, Huawei Wang, Guillaume Durandau, Massimo Sartori, and Vikash
  Kumar.
\newblock {{MyoSuite}}: A contact-rich simulation suite for musculoskeletal
  motor control.
\newblock In \emph{Learning for Dynamics and Control Conference}, pages
  492--507. PMLR, 2022{\natexlab{a}}.

\bibitem[Denayer et~al.(2025)Denayer, Alfio, D{\'\i}az, Sartori, De~Groote,
  De~Pauw, and Verstraten]{denayer2025prisma}
Menthy Denayer, Eligia Alfio, Mar{\'\i}a~Alejandra D{\'\i}az, Massimo Sartori,
  Friedl De~Groote, Kevin De~Pauw, and Tom Verstraten.
\newblock A prisma systematic review through time on predictive musculoskeletal
  simulations.
\newblock \emph{Journal of NeuroEngineering and Rehabilitation}, 22\penalty0
  (1):\penalty0 149, 2025.

\bibitem[Wolpert et~al.(2011)Wolpert, Diedrichsen, and
  Flanagan]{wolpert2011principles}
Daniel~M. Wolpert, J{\"o}rn Diedrichsen, and J.~Randall Flanagan.
\newblock Principles of sensorimotor learning.
\newblock \emph{Nature reviews. Neuroscience}, 12\penalty0 (12):\penalty0
  739--751, December 2011.
\newblock ISSN 1471-0048.
\newblock \doi{10.1038/nrn3112}.

\bibitem[Bernstein(1967)]{bernstein1967coordination}
Nikolai~A. Bernstein.
\newblock \emph{The Co-ordination and Regulation of Movements}.
\newblock Pergamon Press, Oxford, 1967.

\bibitem[Hill(1938)]{hill1938heat}
A.~V. Hill.
\newblock The heat of shortening and the dynamic constants of muscle.
\newblock \emph{Proceedings of the Royal Society of London. Series B,
  Biological Sciences}, 126\penalty0 (843):\penalty0 136--195, 1938.
\newblock \doi{10.1098/rspb.1938.0050}.

\bibitem[Schumacher et~al.(2022)Schumacher, Haeufle, B{\"u}chler, Schmitt, and
  Martius]{schumacher2022dep}
Pierre Schumacher, Daniel Haeufle, Dieter B{\"u}chler, Syn Schmitt, and Georg
  Martius.
\newblock {{DEP-RL}}: {{Embodied}} exploration for reinforcement learning in
  overactuated and musculoskeletal systems.
\newblock In \emph{The Eleventh International Conference on Learning
  Representations}, 2022.

\bibitem[La~Barbera et~al.(2021)La~Barbera, Pardo, Tassa, Daley, Richards,
  Kormushev, and Hutchinson]{la2021ostrichrl}
Vittorio La~Barbera, Fabio Pardo, Yuval Tassa, Monica Daley, Christopher
  Richards, Petar Kormushev, and John Hutchinson.
\newblock Ostrichrl: A musculoskeletal ostrich simulation to study
  bio-mechanical locomotion.
\newblock \emph{arXiv preprint arXiv:2112.06061}, 2021.

\bibitem[Song et~al.(2021)Song, Kidzi{\'n}ski, Peng, Ong, Hicks, Levine,
  Atkeson, and Delp]{song2021deep}
Seungmoon Song, {\L}ukasz Kidzi{\'n}ski, Xue~Bin Peng, Carmichael Ong, Jennifer
  Hicks, Sergey Levine, Christopher~G Atkeson, and Scott~L Delp.
\newblock Deep reinforcement learning for modeling human locomotion control in
  neuromechanical simulation.
\newblock \emph{Journal of neuroengineering and rehabilitation}, 18\penalty0
  (1):\penalty0 126, 2021.

\bibitem[Berg et~al.(2023)Berg, Caggiano, and Kumar]{berg2023sar}
C.~Berg, V.~Caggiano, and Vikash Kumar.
\newblock Sar: Generalization of physiological agility and dexterity via
  synergistic action representation.
\newblock \emph{arXiv preprint arXiv:2307.03716}, 2023.

\bibitem[Chiappa et~al.(2023)Chiappa, Marin~Vargas, Huang, and
  Mathis]{chiappa2023latent}
Alberto~Silvio Chiappa, Alessandro Marin~Vargas, Ann Huang, and Alexander
  Mathis.
\newblock Latent exploration for {{Reinforcement Learning}}.
\newblock \emph{Advances in Neural Information Processing Systems},
  36:\penalty0 56508--56530, December 2023.

\bibitem[Caggiano et~al.(2022{\natexlab{b}})Caggiano, Durandau, Wang, Chiappa,
  Mathis, Tano, Patel, Pouget, Schumacher, Martius, Haeufle, Geng, An, Zhong,
  Ji, Chen, Dong, Yang, Siripurapu, Ferro~Diez, Kopp, Patil, Hochreiter, Tassa,
  Merel, Schultheis, Song, Sartori, and Kumar]{caggiano2023myochallenge}
Vittorio Caggiano, Guillaume Durandau, Huwawei Wang, Alberto Chiappa, Alexander
  Mathis, Pablo Tano, Nisheet Patel, Alexandre Pouget, Pierre Schumacher, Georg
  Martius, Daniel Haeufle, Yiran Geng, Boshi An, Yifan Zhong, Jiaming Ji,
  Yuanpei Chen, Hao Dong, Yaodong Yang, Rahul Siripurapu, Luis~Eduardo
  Ferro~Diez, Michael Kopp, Vihang Patil, Sepp Hochreiter, Yuval Tassa, Josh
  Merel, Randy Schultheis, Seungmoon Song, Massimo Sartori, and Vikash Kumar.
\newblock Myochallenge 2022: Learning contact-rich manipulation using a
  musculoskeletal hand.
\newblock In Marco Ciccone, Gustavo Stolovitzky, and Jacob Albrecht, editors,
  \emph{Proceedings of the NeurIPS 2022 Competitions Track}, volume 220 of
  \emph{Proceedings of Machine Learning Research}, pages 233--250. PMLR, 28
  Nov--09 Dec 2022{\natexlab{b}}.

\bibitem[Feng et~al.(2023)Feng, Xu, and Liu]{feng2023musclevae}
Yusen Feng, Xiyan Xu, and Libin Liu.
\newblock {MuscleVAE: Model-based controllers of muscle-actuated characters}.
\newblock \emph{arXiv preprint arXiv:2312.07340}, 2023.

\bibitem[He et~al.(2024)He, Zuo, Ma, and Sui]{he2024dynsyn}
Kaibo He, Chenhui Zuo, Chengtian Ma, and Yanan Sui.
\newblock {DynSyn: Dynamical synergistic representation for efficient learning
  and control in overactuated embodied systems}.
\newblock \emph{arXiv preprint arXiv:2407.11472}, 2024.

\bibitem[Schumacher et~al.(2025)Schumacher, Geijtenbeek, Caggiano, Kumar,
  Schmitt, Martius, and Haeufle]{schumacher2025natural}
Pierre Schumacher, Thomas Geijtenbeek, Vittorio Caggiano, Vikash Kumar, Syn
  Schmitt, Georg Martius, and Daniel~FB Haeufle.
\newblock Emergence of natural and robust bipedal walking by learning from
  biologically plausible objectives.
\newblock \emph{{iScience}}, 2025.

\bibitem[Simos et~al.(2026)Simos, Chiappa, and Mathis]{simos2026reinforcement}
Merkourios Simos, Alberto~Silvio Chiappa, and Alexander Mathis.
\newblock {KINESIS}: Motion imitation for human musculoskeletal locomotion.
\newblock In \emph{2026 IEEE International Conference on Robotics and
  Automation (ICRA)}, Vienna, Austria, 2026. IEEE.

\bibitem[Caggiano et~al.(2024{\natexlab{a}})Caggiano, Durandau, Wang, Tan,
  Schumacher, Wang, Chiappa, Vargas, Mathis, Won, Park, Park, Shin, Kim, Koo,
  Yang, Dang, Cai, Song, Song, Sartori, and Kumar]{caggiano2024myochallenge}
Vittorio Caggiano, Guillaume Durandau, Huiyi Wang, Chun~Kwang Tan, Pierre
  Schumacher, Huawei Wang, Alberto~Silvio Chiappa, Alessandro~Marin Vargas,
  Alexander Mathis, Jungdam Won, Jungnam Park, Gunwoo Park, Beomsoo Shin,
  Minsueng Kim, Seungbum Koo, Zhuo Yang, Wei Dang, Heng Cai, Jianfei Song,
  Seungmoon Song, Massimo Sartori, and Vikash Kumar.
\newblock Myochallenge 2023: {{Towards}} human-level dexterity and agility.
\newblock 2024{\natexlab{a}}.

\bibitem[Caggiano et~al.(2024{\natexlab{b}})Caggiano, Durandau, Song, Tan,
  Wang, Hodossy, Schumacher, Gionfrida, Sartori, and
  Kumar]{caggiano2024myochallengea}
Vittorio Caggiano, Guillaume Durandau, Seungmoon Song, Chun~Kwang Tan, Huiyi
  Wang, Balint Hodossy, Pierre Schumacher, Letizia Gionfrida, Massimo Sartori,
  and Vikash Kumar.
\newblock {{MyoChallenge}} 2024: {{Physiological Dexterity}} and {{Agility}} in
  {{Bionic Humans}}.
\newblock In \emph{{{NeurIPS}} 2024 {{Competition Track}}}, April
  2024{\natexlab{b}}.

\bibitem[Wang et~al.(2026)Wang, Tan, Hodossy, Lyu, Guo, Zhao, Liu, Li, Simos,
  Ziliotto, et~al.]{wang2026myochallenge}
Cheryl Wang, Chun~Kwang Tan, Balint~K Hodossy, Eric Lyu, Jun Guo, Wentao Zhao,
  Huaping Liu, Chengkun Li, Merkourios Simos, Bianca Ziliotto, et~al.
\newblock Myochallenge 2025: A new benchmark for human athletic intelligence.
\newblock \emph{arXiv preprint arXiv:2605.15650}, 2026.

\bibitem[Schmidt et~al.(2018)Schmidt, Lee, Winstein, Wulf, and
  Zelaznik]{schmidt2018motor}
Richard~A. Schmidt, Timothy~D. Lee, Carolee~J. Winstein, Gabriele Wulf, and
  Howard~N. Zelaznik.
\newblock \emph{Motor Control and Learning: A Behavioral Emphasis}.
\newblock Human Kinetics, Champaign, IL, 6 edition, 2018.
\newblock ISBN 9781492547754.
\newblock With Web Resource.

\bibitem[Cisek(2021)]{cisek2021evolution}
Paul Cisek.
\newblock Evolution of behavioural control from chordates to primates.
\newblock \emph{Philosophical Transactions of the Royal Society B: Biological
  Sciences}, 377\penalty0 (1844):\penalty0 20200522, December 2021.
\newblock \doi{10.1098/rstb.2020.0522}.

\bibitem[Chiappa et~al.(2022)Chiappa, Marin~Vargas, and
  Mathis]{chiappa2022dmap}
Alberto~Silvio Chiappa, Alessandro Marin~Vargas, and Alexander Mathis.
\newblock {{DMAP}}: A {{Distributed Morphological Attention Policy}} for
  learning to locomote with a changing body.
\newblock \emph{Advances in Neural Information Processing Systems},
  35:\penalty0 37214--37227, December 2022.

\bibitem[Ross et~al.(2011)Ross, Gordon, and Bagnell]{ross2011reduction}
Stephane Ross, Geoffrey~J. Gordon, and J.~Andrew Bagnell.
\newblock A reduction of imitation learning and structured prediction to
  no-regret online learning.
\newblock \emph{arXiv preprint arXiv:1011.0686}, 2011.

\bibitem[Lee et~al.(2020)Lee, Hwangbo, Wellhausen, Koltun, and
  Hutter]{lee2020learning}
Joonho Lee, Jemin Hwangbo, Lorenz Wellhausen, Vladlen Koltun, and Marco Hutter.
\newblock Learning quadrupedal locomotion over challenging terrain.
\newblock \emph{Science robotics}, 5\penalty0 (47):\penalty0 eabc5986, 2020.

\bibitem[Lee et~al.(2024)Lee, Bjelonic, Reske, Wellhausen, Miki, and
  Hutter]{lee2024learning}
Joonho Lee, Marko Bjelonic, Alexander Reske, Lorenz Wellhausen, Takahiro Miki,
  and Marco Hutter.
\newblock Learning robust autonomous navigation and locomotion for
  wheeled-legged robots.
\newblock \emph{Science Robotics}, 9\penalty0 (89):\penalty0 eadi9641, 2024.

\bibitem[Li et~al.(2026{\natexlab{a}})Li, Ma, Lin, Du, Liu, Hu, Cui, Zhu,
  Liang, Jia, et~al.]{li2026omniclone}
Yixuan Li, Le~Ma, Yutang Lin, Yushi Du, Mengya Liu, Kaizhe Hu, Jieming Cui,
  Yixin Zhu, Wei Liang, Baoxiong Jia, et~al.
\newblock Omniclone: Engineering a robust, all-rounder whole-body humanoid
  teleoperation system.
\newblock \emph{arXiv preprint arXiv:2603.14327}, 2026{\natexlab{a}}.

\bibitem[Chiappa et~al.(2024{\natexlab{b}})Chiappa, Gangopadhyay, Wang, and
  Takamatsu]{chiappa2024autobidding}
Alberto~Silvio Chiappa, Briti Gangopadhyay, Zhao Wang, and Shingo Takamatsu.
\newblock Auto-bidding in real-time auctions via oracle imitation learning
  (oil).
\newblock \emph{arXiv preprint arXiv:2412.11434}, 2024{\natexlab{b}}.

\bibitem[Proske and Gandevia(2012)]{proske2012proprioceptive}
Uwe Proske and Simon~C Gandevia.
\newblock The proprioceptive senses: Their roles in signaling body shape, body
  position and movement, and muscle force.
\newblock \emph{Physiological reviews}, 2012.

\bibitem[Todorov et~al.(2012)Todorov, Erez, and Tassa]{todorov2012mujoco}
Emanuel Todorov, Tom Erez, and Yuval Tassa.
\newblock Mujoco: {{A}} physics engine for model-based control.
\newblock In \emph{2012 {{IEEE}}/{{RSJ}} International Conference on
  Intelligent Robots and Systems}, pages 5026--5033. IEEE, 2012.

\bibitem[Winters(1990)]{winters1990hillbased}
Jack~M. Winters.
\newblock Hill-{{Based Muscle Models}}: {{A Systems Engineering Perspective}}.
\newblock In Jack~M. Winters and Savio L-Y. Woo, editors, \emph{Multiple
  {{Muscle Systems}}: {{Biomechanics}} and {{Movement Organization}}}, pages
  69--93. Springer, New York, NY, 1990.
\newblock ISBN 978-1-4613-9030-5.
\newblock \doi{10.1007/978-1-4613-9030-5_5}.

\bibitem[Yu et~al.(2019)Yu, Quillen, He, Julian, Hausman, Finn, and
  Levine]{yu2021metaworld}
Tianhe Yu, Deirdre Quillen, Zhanpeng He, Ryan~C. Julian, Karol Hausman, Chelsea
  Finn, and S.~Levine.
\newblock Meta-world: A benchmark and evaluation for multi-task and meta
  reinforcement learning.
\newblock \emph{Conference on Robot Learning}, 2019.

\bibitem[Terry et~al.(2020)Terry, Grammel, Son, Black, and
  Agrawal]{terry2020revisiting}
Justin~K Terry, Nathaniel Grammel, Sanghyun Son, Benjamin Black, and Aakriti
  Agrawal.
\newblock Revisiting parameter sharing in multi-agent deep reinforcement
  learning.
\newblock \emph{arXiv preprint arXiv:2005.13625}, 2020.

\bibitem[Schulman et~al.(2017)Schulman, Wolski, Dhariwal, Radford, and
  Klimov]{schulman2017proximal}
John Schulman, Filip Wolski, Prafulla Dhariwal, Alec Radford, and Oleg Klimov.
\newblock Proximal policy optimization algorithms.
\newblock \emph{arXiv preprint arXiv:1707.06347}, 2017.

\bibitem[Haarnoja et~al.(2018)Haarnoja, Zhou, Abbeel, and
  Levine]{haarnoja2018soft}
Tuomas Haarnoja, Aurick Zhou, Pieter Abbeel, and Sergey Levine.
\newblock Soft actor-critic: {{Off-policy}} maximum entropy deep reinforcement
  learning with a stochastic actor.
\newblock In \emph{International Conference on Machine Learning}, pages
  1861--1870. PMLR, 2018.

\bibitem[Pomerleau(1988)]{pomerleau1988alvinn}
Dean~A. Pomerleau.
\newblock {{ALVINN}}: {{An Autonomous Land Vehicle}} in a {{Neural Network}}.
\newblock In \emph{Advances in {{Neural Information Processing Systems}}},
  volume~1. Morgan-Kaufmann, 1988.

\bibitem[Michie et~al.(1990)Michie, Bain, and
  Hayes-Michie]{michie1990cognitive}
Donald Michie, Michael Bain, and Jean~E. Hayes-Michie.
\newblock Cognitive models from subcognitive skills.
\newblock In M.~Grimble, S.~McGhee, and P.~Mowforth, editors,
  \emph{Knowledge-Based Systems in Industrial Control}. Peter Peregrinus,
  Stevenage, 1990.

\bibitem[Sammut et~al.(1992)Sammut, Hurst, Kedzier, and
  Michie]{sammut1992learning}
Claude Sammut, Scott Hurst, Dana Kedzier, and Donald Michie.
\newblock Learning to fly.
\newblock In D.~Sleeman and P.~Edwards, editors, \emph{Proceedings of the Ninth
  International Conference on Machine Learning}, pages 385--393, San Francisco,
  1992. Morgan Kaufmann.

\bibitem[Bi et~al.(2018)Bi, Xiao, Sun, and Xu]{bi2018navigation}
Jing Bi, Tianyou Xiao, Qiuyue Sun, and Chenliang Xu.
\newblock Navigation by imitation in a pedestrian-rich environment.
\newblock \emph{arXiv preprint arXiv:1811.00506}, 2018.

\bibitem[Kelly et~al.(2019)Kelly, Sidrane, Driggs-Campbell, and
  Kochenderfer]{kelly2019hg}
Michael Kelly, Chelsea Sidrane, Katherine Driggs-Campbell, and Mykel~J
  Kochenderfer.
\newblock Hg-dagger: Interactive imitation learning with human experts.
\newblock In \emph{2019 International Conference on Robotics and Automation
  (ICRA)}, pages 8077--8083. IEEE, 2019.

\bibitem[Sun et~al.(2024)Sun, Yang, Zhou, Liu, and Mangharam]{sun2024mega}
Xiatao Sun, Shuo Yang, Mingyan Zhou, Kunpeng Liu, and Rahul Mangharam.
\newblock Mega-dagger: Imitation learning with multiple imperfect experts.
\newblock In \emph{2024 IEEE 27th International Conference on Intelligent
  Transportation Systems (ITSC)}, pages 1--8. IEEE, 2024.

\bibitem[Li et~al.(2024)Li, Wang, Pang, Bai, Hu, Liu, Wang, and
  Li]{li2024learning}
Sicen Li, Gang Wang, Yiming Pang, Panju Bai, Shihao Hu, Zhaojin Liu, Liquan
  Wang, and Jiawei Li.
\newblock Learning agility and adaptive legged locomotion via curricular
  hindsight reinforcement learning.
\newblock \emph{Scientific Reports}, 14\penalty0 (1):\penalty0 28089, 2024.

\bibitem[He et~al.(2026)He, Wang, Xue, Ben, Luo, Xiao, Yuan, Da, Casta{\~n}eda,
  Sastry, et~al.]{he2026viral}
Tairan He, Zi~Wang, Haoru Xue, Qingwei Ben, Zhengyi Luo, Wenli Xiao, Ye~Yuan,
  Xingye Da, Fernando Casta{\~n}eda, Shankar Sastry, et~al.
\newblock Viral: Visual sim-to-real at scale for humanoid loco-manipulation.
\newblock In \emph{Proceedings of the IEEE/CVF Conference on Computer Vision
  and Pattern Recognition}, pages 13430--13441, 2026.

\bibitem[Wu et~al.(2026)Wu, Huang, Yang, Zhang, Chen, Abbeel, Duan, Kanazawa,
  Sferrazza, Shi, et~al.]{wu2026perceptive}
Zhen Wu, Xiaoyu Huang, Lujie Yang, Yuanhang Zhang, Xi~Chen, Pieter Abbeel,
  Rocky Duan, Angjoo Kanazawa, Carmelo Sferrazza, Guanya Shi, et~al.
\newblock Perceptive humanoid parkour: Chaining dynamic human skills via motion
  matching.
\newblock \emph{arXiv preprint arXiv:2602.15827}, 2026.

\bibitem[DeepSeek-AI et~al.(2025)DeepSeek-AI, Guo, Yang, Zhang, Song, Zhang,
  Xu, Zhu, Ma, Wang, Bi, Zhang, Yu, Wu, Wu, Gou, Shao, Li, Gao, Liu, Xue, Wang,
  Wu, Feng, Lu, Zhao, Deng, Zhang, Ruan, Dai, Chen, Ji, Li, Lin, Dai, Luo, Hao,
  Chen, Li, Zhang, Bao, Xu, Wang, Ding, Xin, Gao, Qu, Li, Guo, Li, Wang, Chen,
  Yuan, Qiu, Li, Cai, Ni, Liang, Chen, Dong, Hu, Gao, Guan, Huang, Yu, Wang,
  Zhang, Zhao, Wang, Zhang, Xu, Xia, Zhang, Zhang, Tang, Li, Wang, Li, Tian,
  Huang, Zhang, Wang, Chen, Du, Ge, Zhang, Pan, Wang, Chen, Jin, Chen, Lu,
  Zhou, Chen, Ye, Wang, Yu, Zhou, Pan, Li, Zhou, Wu, Ye, Yun, Pei, Sun, Wang,
  Zeng, Zhao, Liu, Liang, Gao, Yu, Zhang, Xiao, An, Liu, Wang, Chen, Nie,
  Cheng, Liu, Xie, Liu, Yang, Li, Su, Lin, Li, Jin, Shen, Chen, Sun, Wang,
  Song, Zhou, Wang, Shan, Li, Wang, Wei, Zhang, Xu, Li, Zhao, Sun, Wang, Yu,
  Zhang, Shi, Xiong, He, Piao, Wang, Tan, Ma, Liu, Guo, Ou, Wang, Gong, Zou,
  He, Xiong, Luo, You, Liu, Zhou, Zhu, Xu, Huang, Li, Zheng, Zhu, Ma, Tang,
  Zha, Yan, Ren, Ren, Sha, Fu, Xu, Xie, Zhang, Hao, Ma, Yan, Wu, Gu, Zhu, Liu,
  Li, Xie, Song, Pan, Huang, Xu, Zhang, and Zhang]{deepseek-ai2025deepseekr1}
DeepSeek-AI, Daya Guo, Dejian Yang, Haowei Zhang, Junxiao Song, Ruoyu Zhang,
  Runxin Xu, Qihao Zhu, Shirong Ma, Peiyi Wang, Xiao Bi, Xiaokang Zhang,
  Xingkai Yu, Yu~Wu, Z.~F. Wu, Zhibin Gou, Zhihong Shao, Zhuoshu Li, Ziyi Gao,
  Aixin Liu, Bing Xue, Bingxuan Wang, Bochao Wu, Bei Feng, Chengda Lu,
  Chenggang Zhao, Chengqi Deng, Chenyu Zhang, Chong Ruan, Damai Dai, Deli Chen,
  Dongjie Ji, Erhang Li, Fangyun Lin, Fucong Dai, Fuli Luo, Guangbo Hao,
  Guanting Chen, Guowei Li, H.~Zhang, Han Bao, Hanwei Xu, Haocheng Wang,
  Honghui Ding, Huajian Xin, Huazuo Gao, Hui Qu, Hui Li, Jianzhong Guo, Jiashi
  Li, Jiawei Wang, Jingchang Chen, Jingyang Yuan, Junjie Qiu, Junlong Li, J.~L.
  Cai, Jiaqi Ni, Jian Liang, Jin Chen, Kai Dong, Kai Hu, Kaige Gao, Kang Guan,
  Kexin Huang, Kuai Yu, Lean Wang, Lecong Zhang, Liang Zhao, Litong Wang, Liyue
  Zhang, Lei Xu, Leyi Xia, Mingchuan Zhang, Minghua Zhang, Minghui Tang, Meng
  Li, Miaojun Wang, Mingming Li, Ning Tian, Panpan Huang, Peng Zhang, Qiancheng
  Wang, Qinyu Chen, Qiushi Du, Ruiqi Ge, Ruisong Zhang, Ruizhe Pan, Runji Wang,
  R.~J. Chen, R.~L. Jin, Ruyi Chen, Shanghao Lu, Shangyan Zhou, Shanhuang Chen,
  Shengfeng Ye, Shiyu Wang, Shuiping Yu, Shunfeng Zhou, Shuting Pan, S.~S. Li,
  Shuang Zhou, Shaoqing Wu, Shengfeng Ye, Tao Yun, Tian Pei, Tianyu Sun,
  T.~Wang, Wangding Zeng, Wanjia Zhao, Wen Liu, Wenfeng Liang, Wenjun Gao,
  Wenqin Yu, Wentao Zhang, W.~L. Xiao, Wei An, Xiaodong Liu, Xiaohan Wang,
  Xiaokang Chen, Xiaotao Nie, Xin Cheng, Xin Liu, Xin Xie, Xingchao Liu, Xinyu
  Yang, Xinyuan Li, Xuecheng Su, Xuheng Lin, X.~Q. Li, Xiangyue Jin, Xiaojin
  Shen, Xiaosha Chen, Xiaowen Sun, Xiaoxiang Wang, Xinnan Song, Xinyi Zhou,
  Xianzu Wang, Xinxia Shan, Y.~K. Li, Y.~Q. Wang, Y.~X. Wei, Yang Zhang,
  Yanhong Xu, Yao Li, Yao Zhao, Yaofeng Sun, Yaohui Wang, Yi~Yu, Yichao Zhang,
  Yifan Shi, Yiliang Xiong, Ying He, Yishi Piao, Yisong Wang, Yixuan Tan,
  Yiyang Ma, Yiyuan Liu, Yongqiang Guo, Yuan Ou, Yuduan Wang, Yue Gong, Yuheng
  Zou, Yujia He, Yunfan Xiong, Yuxiang Luo, Yuxiang You, Yuxuan Liu, Yuyang
  Zhou, Y.~X. Zhu, Yanhong Xu, Yanping Huang, Yaohui Li, Yi~Zheng, Yuchen Zhu,
  Yunxian Ma, Ying Tang, Yukun Zha, Yuting Yan, Z.~Z. Ren, Zehui Ren, Zhangli
  Sha, Zhe Fu, Zhean Xu, Zhenda Xie, Zhengyan Zhang, Zhewen Hao, Zhicheng Ma,
  Zhigang Yan, Zhiyu Wu, Zihui Gu, Zijia Zhu, Zijun Liu, Zilin Li, Ziwei Xie,
  Ziyang Song, Zizheng Pan, Zhen Huang, Zhipeng Xu, Zhongyu Zhang, and Zhen
  Zhang.
\newblock Deepseek-r1: Incentivizing reasoning capability in llms via
  reinforcement learning.
\newblock \emph{arXiv preprint arXiv: 2501.12948}, 2025.

\bibitem[Rajeswaran et~al.(2018)Rajeswaran, Kumar, Gupta, Vezzani, Schulman,
  Todorov, and Levine]{rajeswaran2018learning}
Aravind Rajeswaran, Vikash Kumar, Abhishek Gupta, Giulia Vezzani, John
  Schulman, Emanuel Todorov, and Sergey Levine.
\newblock Learning complex dexterous manipulation with deep reinforcement
  learning and demonstrations.
\newblock \emph{arXiv preprint arXiv:1709.10087}, 2018.

\bibitem[Flash and Hogan(1985)]{flash1985coordination}
Tamar Flash and Neville Hogan.
\newblock The coordination of arm movements: An experimentally confirmed
  mathematical model.
\newblock \emph{Journal of neuroscience}, 5\penalty0 (7):\penalty0 1688--1703,
  1985.

\bibitem[Harris and Wolpert(1998)]{harris1998signal}
Christopher~M Harris and Daniel~M Wolpert.
\newblock Signal-dependent noise determines motor planning.
\newblock \emph{Nature}, 394\penalty0 (6695):\penalty0 780--784, 1998.

\bibitem[Mysore et~al.(2021)Mysore, Mabsout, Mancuso, and
  Saenko]{mysore2021regularizing}
Siddharth Mysore, Bassel Mabsout, Renato Mancuso, and Kate Saenko.
\newblock Regularizing action policies for smooth control with reinforcement
  learning.
\newblock In \emph{2021 IEEE International Conference on Robotics and
  Automation (ICRA)}, pages 1810--1816. IEEE, 2021.

\bibitem[Balasubramanian et~al.(2015)Balasubramanian, Melendez-Calderon,
  Roby-Brami, and Burdet]{balasubramanian2015analysis}
Sivakumar Balasubramanian, Alejandro Melendez-Calderon, Agnes Roby-Brami, and
  Etienne Burdet.
\newblock On the analysis of movement smoothness.
\newblock \emph{Journal of neuroengineering and rehabilitation}, 12\penalty0
  (1):\penalty0 112, 2015.

\bibitem[Rohrer et~al.(2002)Rohrer, Fasoli, Krebs, Hughes, Volpe, Frontera,
  Stein, and Hogan]{rohrer2002movement}
Brandon Rohrer, Susan Fasoli, Hermano~Igo Krebs, Richard Hughes, Bruce Volpe,
  Walter~R Frontera, Joel Stein, and Neville Hogan.
\newblock Movement smoothness changes during stroke recovery.
\newblock \emph{Journal of neuroscience}, 22\penalty0 (18):\penalty0
  8297--8304, 2002.

\bibitem[Hogan and Sternad(2009)]{hogan2009sensitivity}
Neville Hogan and Dagmar Sternad.
\newblock Sensitivity of smoothness measures to movement duration, amplitude,
  and arrests.
\newblock \emph{Journal of motor behavior}, 41\penalty0 (6):\penalty0 529--534,
  2009.

\bibitem[Ivanenko et~al.(2004)Ivanenko, Poppele, and
  Lacquaniti]{ivanenko2004five}
Yuri~P Ivanenko, Richard~E Poppele, and Francesco Lacquaniti.
\newblock Five basic muscle activation patterns account for muscle activity
  during human locomotion.
\newblock \emph{The Journal of physiology}, 556\penalty0 (1):\penalty0
  267--282, 2004.

\bibitem[Todorov and Ghahramani(2004)]{todorov2004analysis}
Emanuel Todorov and Zoubin Ghahramani.
\newblock Analysis of the synergies underlying complex hand manipulation.
\newblock In \emph{The 26th Annual International Conference of the {{IEEE}}
  Engineering in Medicine and Biology Society}, volume~2, pages 4637--4640.
  IEEE, 2004.

\bibitem[Wang et~al.(2023)Wang, Basu, Durandau, and Sartori]{wang2023wearable}
Huawei Wang, Akash Basu, Guillaume Durandau, and Massimo Sartori.
\newblock A wearable real-time kinetic measurement sensor setup for human
  locomotion.
\newblock \emph{Wearable technologies}, 4:\penalty0 e11, 2023.

\bibitem[{Mussa-Ivaldi} et~al.(1994){Mussa-Ivaldi}, Giszter, and
  Bizzi]{mussa1994linear}
Ferdinando~A {Mussa-Ivaldi}, Simon~F Giszter, and Emilio Bizzi.
\newblock Linear combinations of primitives in vertebrate motor control.
\newblock \emph{Proceedings of the National Academy of Sciences}, 91\penalty0
  (16):\penalty0 7534--7538, 1994.

\bibitem[{d'Avella} and Tresch(2001)]{davella2001modularity}
Andrea {d'Avella} and {\relax MMCM}~Tresch.
\newblock Modularity in the motor system: Decomposition of muscle patterns as
  combinations of time-varying synergies.
\newblock \emph{Advances in neural information processing systems}, 14, 2001.

\bibitem[{d'Avella} et~al.(2003){d'Avella}, Saltiel, and
  Bizzi]{d2003combinations}
Andrea {d'Avella}, Philippe Saltiel, and Emilio Bizzi.
\newblock Combinations of muscle synergies in the construction of a natural
  motor behavior.
\newblock \emph{Nature neuroscience}, 6\penalty0 (3):\penalty0 300--308, 2003.

\bibitem[Tresch and Jarc(2009)]{tresch2009case}
Matthew~C Tresch and Anthony Jarc.
\newblock The case for and against muscle synergies.
\newblock \emph{Current opinion in neurobiology}, 19\penalty0 (6):\penalty0
  601--607, 2009.

\bibitem[Alessandro et~al.(2013)Alessandro, Delis, Nori, Panzeri, and
  Berret]{alessandro2013muscle}
Cristiano Alessandro, Ioannis Delis, Francesco Nori, Stefano Panzeri, and
  Bastien Berret.
\newblock Muscle synergies in neuroscience and robotics: From input-space to
  task-space perspectives.
\newblock \emph{Frontiers in computational neuroscience}, 7:\penalty0 43, 2013.

\bibitem[Loeb(2021)]{loeb2021learning}
Gerald~E Loeb.
\newblock Learning to use muscles.
\newblock \emph{Journal of human kinetics}, 76\penalty0 (1):\penalty0 9--33,
  2021.

\bibitem[Kidzi{\'n}ski et~al.(2018{\natexlab{a}})Kidzi{\'n}ski, Mohanty, Ong,
  Hicks, Carroll, Levine, Salath{\'e}, and Delp]{kidzinski2018learning}
{\L}ukasz Kidzi{\'n}ski, Sharada~P. Mohanty, Carmichael~F. Ong, Jennifer~L.
  Hicks, Sean~F. Carroll, Sergey Levine, Marcel Salath{\'e}, and Scott~L. Delp.
\newblock Learning to {{Run Challenge}}: {{Synthesizing Physiologically
  Accurate Motion Using Deep Reinforcement Learning}}.
\newblock In Sergio Escalera and Markus Weimer, editors, \emph{The {{NIPS}} '17
  {{Competition}}: {{Building Intelligent Systems}}}, pages 101--120, Cham,
  2018{\natexlab{a}}. Springer International Publishing.
\newblock ISBN 978-3-319-94042-7.
\newblock \doi{10.1007/978-3-319-94042-7_6}.

\bibitem[Kidzi{\'n}ski et~al.(2018{\natexlab{b}})Kidzi{\'n}ski, Mohanty, Ong,
  Huang, Zhou, Pechenko, Stelmaszczyk, Jarosik, Pavlov, Kolesnikov, Plis, Chen,
  Zhang, Chen, Shi, Zheng, Yuan, Lin, Michalewski, Mi{\l}o{\'s}, Osi{\'n}ski,
  Melnik, Schilling, Ritter, Carroll, Hicks, Levine, Salath{\'e}, and
  Delp]{kidzinski2018learninga}
{\L}ukasz Kidzi{\'n}ski, Sharada~Prasanna Mohanty, Carmichael Ong, Zhewei
  Huang, Shuchang Zhou, Anton Pechenko, Adam Stelmaszczyk, Piotr Jarosik,
  Mikhail Pavlov, Sergey Kolesnikov, Sergey Plis, Zhibo Chen, Zhizheng Zhang,
  Jiale Chen, Jun Shi, Zhuobin Zheng, Chun Yuan, Zhihui Lin, Henryk
  Michalewski, Piotr Mi{\l}o{\'s}, B{\l}a{\.z}ej Osi{\'n}ski, Andrew Melnik,
  Malte Schilling, Helge Ritter, Sean Carroll, Jennifer Hicks, Sergey Levine,
  Marcel Salath{\'e}, and Scott Delp.
\newblock Learning to run challenge solutions: Adapting reinforcement learning
  methods for neuromusculoskeletal environments.
\newblock \emph{arXiv preprint arXiv:1804.00361}, 2018{\natexlab{b}}.

\bibitem[Kidzi{\'n}ski et~al.(2019)Kidzi{\'n}ski, Ong, Mohanty, Hicks, Carroll,
  Zhou, Zeng, Wang, Lian, Tian, Ja{\'s}kowski, Andersen, Lykkeb{\o}, Toklu,
  Shyam, Srivastava, Kolesnikov, Hrinchuk, Pechenko, Ljungstr{\"o}m, Wang, Hu,
  Hu, Qiu, Huang, Shpilman, Sosin, Svidchenko, Malysheva, Kudenko, Rane, Bhatt,
  Wang, Qi, Yu, Peng, Yuan, Li, Tian, Yang, Ma, Khadka, Majumdar, Dwiel, Liu,
  Tumer, Watson, Salath{\'e}, Levine, and Delp]{kidzinski2019artificial}
{\L}ukasz Kidzi{\'n}ski, Carmichael Ong, Sharada~Prasanna Mohanty, Jennifer
  Hicks, Sean~F. Carroll, Bo~Zhou, Hongsheng Zeng, Fan Wang, Rongzhong Lian,
  Hao Tian, Wojciech Ja{\'s}kowski, Garrett Andersen, Odd~Rune Lykkeb{\o},
  Nihat~Engin Toklu, Pranav Shyam, Rupesh~Kumar Srivastava, Sergey Kolesnikov,
  Oleksii Hrinchuk, Anton Pechenko, Mattias Ljungstr{\"o}m, Zhen Wang, Xu~Hu,
  Zehong Hu, Minghui Qiu, Jun Huang, Aleksei Shpilman, Ivan Sosin, Oleg
  Svidchenko, Aleksandra Malysheva, Daniel Kudenko, Lance Rane, Aditya Bhatt,
  Zhengfei Wang, Penghui Qi, Zeyang Yu, Peng Peng, Quan Yuan, Wenxin Li,
  Yunsheng Tian, Ruihan Yang, Pingchuan Ma, Shauharda Khadka, Somdeb Majumdar,
  Zach Dwiel, Yinyin Liu, Evren Tumer, Jeremy Watson, Marcel Salath{\'e},
  Sergey Levine, and Scott Delp.
\newblock Artificial intelligence for prosthetics - challenge solutions.
\newblock \emph{arXiv preprint arXiv:1902.02441}, 2019.

\bibitem[Kidzi{\'n}ski et~al.(2020)Kidzi{\'n}ski, Ong, Mohanty, Hicks, Carroll,
  Zhou, Zeng, Wang, Lian, Tian, et~al.]{kidzinski2020artificial}
{\L}ukasz Kidzi{\'n}ski, Carmichael Ong, Sharada~Prasanna Mohanty, Jennifer
  Hicks, Sean Carroll, Bo~Zhou, Hongsheng Zeng, Fan Wang, Rongzhong Lian, Hao
  Tian, et~al.
\newblock Artificial intelligence for prosthetics: {{Challenge}} solutions.
\newblock In \emph{The {{NeurIPS}}'18 Competition: {{From}} Machine Learning to
  Intelligent Conversations}, pages 69--128. Springer, 2020.

\bibitem[Wei et~al.(2026)Wei, Zuo, and Sui]{wei2026scalable}
Yunyue Wei, Chenhui Zuo, and Yanan Sui.
\newblock Scalable exploration for high-dimensional continuous control via
  value-guided flow.
\newblock \emph{arXiv preprint arXiv:2601.19707}, 2026.

\bibitem[Yao et~al.(2022)Yao, Song, Chen, and Liu]{yao2022controlvae}
Heyuan Yao, Zhenhua Song, Baoquan Chen, and Libin Liu.
\newblock Controlvae: Model-based learning of generative controllers for
  physics-based characters.
\newblock \emph{{ACM} Trans. Graph.}, 41\penalty0 (6):\penalty0 183:1--183:16,
  2022.
\newblock \doi{10.1145/3550454.3555434}.
\newblock URL \url{https://doi.org/10.1145/3550454.3555434}.

\bibitem[Li et~al.(2026{\natexlab{b}})Li, Wang, Ziliotto, Simos, Kovecses,
  Durandau, and Mathis]{li2026towards}
Chengkun Li, Cheryl Wang, Bianca Ziliotto, Merkourios Simos, Jozsef Kovecses,
  Guillaume Durandau, and Alexander Mathis.
\newblock Towards embodied ai with musclemimic: Unlocking full-body
  musculoskeletal motor learning at scale.
\newblock \emph{arXiv preprint arXiv:2603.25544}, 2026{\natexlab{b}}.

\bibitem[Hansen et~al.(2024)Hansen, Su, and Wang]{hansen2024td}
Nicklas Hansen, Hao Su, and Xiaolong Wang.
\newblock Td-mpc2: Scalable, robust world models for continuous control.
\newblock In \emph{The Twelfth International Conference on Learning
  Representations (ICLR)}, 2024.
\newblock URL \url{https://openreview.net/forum?id=vP6vSGi5WH}.

\bibitem[Rusu et~al.(2016)Rusu, Colmenarejo, Gulcehre, Desjardins, Kirkpatrick,
  Pascanu, Mnih, Kavukcuoglu, and Hadsell]{rusu2016policy}
Andrei~A. Rusu, Sergio~Gomez Colmenarejo, Caglar Gulcehre, Guillaume
  Desjardins, James Kirkpatrick, Razvan Pascanu, Volodymyr Mnih, Koray
  Kavukcuoglu, and Raia Hadsell.
\newblock Policy distillation.
\newblock \emph{arXiv preprint arXiv:1511.06295}, 2016.

\bibitem[Parisotto et~al.(2016)Parisotto, Ba, and
  Salakhutdinov]{parisotto2016actormimic}
Emilio Parisotto, Jimmy~Lei Ba, and Ruslan Salakhutdinov.
\newblock Actor-mimic: Deep multitask and transfer reinforcement learning.
\newblock \emph{arXiv preprint arXiv:1511.06342}, 2016.

\bibitem[Teh et~al.(2017)Teh, Bapst, Czarnecki, Quan, Kirkpatrick, Hadsell,
  Heess, and Pascanu]{teh2017distral}
Yee Teh, Victor Bapst, Wojciech~M. Czarnecki, John Quan, James Kirkpatrick,
  Raia Hadsell, Nicolas Heess, and Razvan Pascanu.
\newblock Distral: {{Robust}} multitask reinforcement learning.
\newblock In \emph{Advances in {{Neural Information Processing Systems}}},
  volume~30. Curran Associates, Inc., 2017.

\bibitem[Bagatella et~al.(2024)Bagatella, Hübotter, Martius, and
  Krause]{bagatella2024active}
Marco Bagatella, Jonas Hübotter, Georg Martius, and Andreas Krause.
\newblock Active fine-tuning of generalist policies.
\newblock \emph{arXiv preprint arXiv:2410.05026}, 2024.

\bibitem[Xu et~al.(2024)Xu, Li, Luo, and Levine]{xu2024rldg}
Charles Xu, Qiyang Li, Jianlan Luo, and Sergey Levine.
\newblock {RLDG: Robotic generalist policy distillation via reinforcement
  learning}.
\newblock \emph{arXiv preprint arXiv:2412.09858}, 2024.

\bibitem[Furuta et~al.(2022)Furuta, Iwasawa, Matsuo, and Gu]{furuta2023system}
Hiroki Furuta, Yusuke Iwasawa, Yutaka Matsuo, and S.~Gu.
\newblock A system for morphology-task generalization via unified
  representation and behavior distillation.
\newblock \emph{International Conference on Learning Representations}, 2022.
\newblock \doi{10.48550/arXiv.2211.14296}.

\bibitem[Jia et~al.(2022)Jia, Li, Ling, Liu, Wu, and Su]{jia2022improving}
Zhiwei Jia, Xuanlin Li, Zhan Ling, Shuang Liu, Yiran Wu, and Hao Su.
\newblock Improving {{Policy Optimization}} with {{Generalist-Specialist
  Learning}}.
\newblock In \emph{Proceedings of the 39th {{International Conference}} on
  {{Machine Learning}}}, pages 10104--10119. PMLR, June 2022.

\bibitem[Chen et~al.(2021)Chen, Lu, Rajeswaran, Lee, Grover, Laskin, Abbeel,
  Srinivas, and Mordatch]{chen2021decision}
Lili Chen, Kevin Lu, Aravind Rajeswaran, Kimin Lee, Aditya Grover, Misha
  Laskin, Pieter Abbeel, Aravind Srinivas, and Igor Mordatch.
\newblock Decision transformer: Reinforcement learning via sequence modeling.
\newblock \emph{Advances in neural information processing systems},
  34:\penalty0 15084--15097, 2021.

\bibitem[Reed et~al.(2022)Reed, Zolna, Parisotto, Colmenarejo, Novikov,
  {Barth-Maron}, Gimenez, Sulsky, Kay, Springenberg, Eccles, Bruce, Razavi,
  Edwards, Heess, Chen, Hadsell, Vinyals, Bordbar, and
  de~Freitas]{reed2022generalist}
Scott Reed, Konrad Zolna, Emilio Parisotto, Sergio~Gomez Colmenarejo, Alexander
  Novikov, Gabriel {Barth-Maron}, Mai Gimenez, Yury Sulsky, Jackie Kay,
  Jost~Tobias Springenberg, Tom Eccles, Jake Bruce, Ali Razavi, Ashley Edwards,
  Nicolas Heess, Yutian Chen, Raia Hadsell, Oriol Vinyals, Mahyar Bordbar, and
  Nando de~Freitas.
\newblock A generalist agent.
\newblock \emph{arXiv preprint arXiv:2205.06175}, 2022.

\bibitem[Zargarbashi et~al.(2024)Zargarbashi, Cheng, Kang, Sumner, and
  Coros]{zargarbashi2024robotkeyframing}
Fatemeh Zargarbashi, Jin Cheng, Dongho Kang, Robert Sumner, and Stelian Coros.
\newblock Robotkeyframing: Learning locomotion with high-level objectives via
  mixture of dense and sparse rewards.
\newblock \emph{Conference on Robot Learning}, 2024.
\newblock \doi{10.48550/arXiv.2407.11562}.

\bibitem[Team et~al.(2024)Team, Ghosh, Walke, {Karl Pertsch}, Black, Mees,
  Dasari, Hejna, Kreiman, Xu, Luo, Tan, Chen, Sanketi, Vuong, Xiao, Sadigh,
  Finn, and Levine]{team2024octo}
Octo~Model Team, Dibya Ghosh, Homer Walke, {Karl Pertsch}, Kevin Black, Oier
  Mees, Sudeep Dasari, Joey Hejna, Tobias Kreiman, Charles Xu, Jianlan Luo,
  You~Liang Tan, Lawrence~Yunliang Chen, Pannag Sanketi, Quan Vuong, Ted Xiao,
  Dorsa Sadigh, Chelsea Finn, and Sergey Levine.
\newblock Octo: An open-source generalist robot policy.
\newblock \emph{arXiv preprint arXiv:2405.12213}, 2024.

\bibitem[Trabucco et~al.(2022)Trabucco, Phielipp, and
  Berseth]{trabucco2022anymorph}
Brandon Trabucco, Mariano Phielipp, and Glen Berseth.
\newblock {{AnyMorph}}: {{Learning}} transferable polices by inferring agent
  morphology.
\newblock In \emph{International conference on Machine learning}, International
  {{Conference}} on {{Machine Learning}}, pages 21677--21691. PMLR, 2022.

\bibitem[Wang et~al.(2024)Wang, Chen, Zhao, and He]{wang2024scaling}
Lirui Wang, Xinlei Chen, Jialiang Zhao, and Kaiming He.
\newblock Scaling proprioceptive-visual learning with heterogeneous pre-trained
  transformers.
\newblock In \emph{The Thirty-eighth Annual Conference on Neural Information
  Processing Systems}, 2024.
\newblock URL \url{https://openreview.net/forum?id=Pf7kdIjHRf}.

\bibitem[Sferrazza et~al.(2024)Sferrazza, Huang, Liu, Lee, and
  Abbeel]{sferrazza2024body}
Carmelo Sferrazza, Dun-Ming Huang, Fangchen Liu, Jongmin Lee, and Pieter
  Abbeel.
\newblock Body transformer: Leveraging robot embodiment for policy learning.
\newblock \emph{arXiv preprint arXiv:2408.06316}, 2024.

\bibitem[Brohan et~al.(2022)Brohan, Brown, Carbajal, Chebotar, Dabis, Finn,
  Gopalakrishnan, Hausman, Herzog, Hsu, Ibarz, Ichter, Irpan, Jackson,
  Jesmonth, Joshi, Julian, Kalashnikov, Kuang, Leal, Lee, Levine, Lu, Malla,
  Manjunath, Mordatch, Nachum, Parada, Peralta, Perez, Pertsch, Quiambao, Rao,
  Ryoo, Salazar, Sanketi, Sayed, Singh, Sontakke, Stone, Tan, Tran, Vanhoucke,
  Vega, Vuong, Xia, Xiao, Xu, Xu, Yu, and Zitkovich]{brohan2023rt1}
Anthony Brohan, Noah Brown, Justice Carbajal, Yevgen Chebotar, Joseph Dabis,
  Chelsea Finn, K.~Gopalakrishnan, Karol Hausman, Alexander Herzog, Jasmine
  Hsu, Julian Ibarz, Brian Ichter, A.~Irpan, Tomas Jackson, Sally Jesmonth,
  Nikhil~J. Joshi, Ryan~C. Julian, Dmitry Kalashnikov, Yuheng Kuang, Isabel
  Leal, Kuang-Huei Lee, S.~Levine, Yao Lu, U.~Malla, D.~Manjunath, Igor
  Mordatch, Ofir Nachum, Carolina Parada, Jodilyn Peralta, Emily Perez, Karl
  Pertsch, Jornell Quiambao, Kanishka Rao, M.~Ryoo, Grecia Salazar, Pannag~R.
  Sanketi, Kevin Sayed, Jaspiar Singh, S.~Sontakke, Austin Stone, Clayton Tan,
  Huong Tran, Vincent Vanhoucke, Steve Vega, Q.~Vuong, F.~Xia, Ted Xiao, Peng
  Xu, Sichun Xu, Tianhe Yu, and Brianna Zitkovich.
\newblock Rt-1: Robotics transformer for real-world control at scale.
\newblock \emph{arXiv preprint arXiv:2212.06817}, 2022.

\bibitem[Kim et~al.(2024)Kim, Pertsch, Karamcheti, Xiao, Balakrishna, Nair,
  Rafailov, Foster, Lam, Sanketi, Vuong, Kollar, Burchfiel, Tedrake, Sadigh,
  Levine, Liang, and Finn]{kim2024openvla}
Moo~Jin Kim, Karl Pertsch, Siddharth Karamcheti, Ted Xiao, Ashwin Balakrishna,
  Suraj Nair, Rafael Rafailov, Ethan Foster, Grace Lam, Pannag Sanketi, Quan
  Vuong, Thomas Kollar, Benjamin Burchfiel, Russ Tedrake, Dorsa Sadigh, Sergey
  Levine, Percy Liang, and Chelsea Finn.
\newblock {OpenVLA: An open-source vision-language-action model}.
\newblock \emph{arXiv preprint arXiv:2406.09246}, 2024.

\bibitem[Mathis(2024)]{mathis2024adaptive}
Mackenzie~Weygandt Mathis.
\newblock Adaptive intelligence: leveraging insights from adaptive behavior in
  animals to build flexible ai systems.
\newblock \emph{arXiv preprint arXiv:2411.15234}, 2024.

\bibitem[Yan et~al.(2020)Yan, Goodman, Moore, Solla, and
  Bensmaia]{yan2020unexpected}
Yuke Yan, James~M Goodman, Dalton~D Moore, Sara~A Solla, and Sliman~J Bensmaia.
\newblock Unexpected complexity of everyday manual behaviors.
\newblock \emph{Nature communications}, 11\penalty0 (1):\penalty0 3564, 2020.

\bibitem[Johnston and Fusi(2023)]{johnston2023abstract}
W.~Jeffrey Johnston and Stefano Fusi.
\newblock Abstract representations emerge naturally in neural networks trained
  to perform multiple tasks.
\newblock \emph{Nature Communications}, 14\penalty0 (1):\penalty0 1040,
  February 2023.
\newblock ISSN 2041-1723.
\newblock \doi{10.1038/s41467-023-36583-0}.

\bibitem[Driscoll et~al.(2024)Driscoll, Shenoy, and
  Sussillo]{driscoll2024flexible}
Laura~N. Driscoll, Krishna Shenoy, and David Sussillo.
\newblock Flexible multitask computation in recurrent networks utilizes shared
  dynamical motifs.
\newblock \emph{Nature Neuroscience}, 27\penalty0 (7):\penalty0 1349--1363,
  July 2024.
\newblock ISSN 1546-1726.
\newblock \doi{10.1038/s41593-024-01668-6}.

\bibitem[Riveland and Pouget(2024)]{riveland2024natural}
Reidar Riveland and Alexandre Pouget.
\newblock Natural language instructions induce compositional generalization in
  networks of neurons.
\newblock \emph{Nature Neuroscience}, 27\penalty0 (5):\penalty0 988--999, May
  2024.
\newblock ISSN 1546-1726.
\newblock \doi{10.1038/s41593-024-01607-5}.

\bibitem[Shenoy et~al.(2013)Shenoy, Sahani, and Churchland]{shenoy2013cortical}
Krishna~V Shenoy, Maneesh Sahani, and Mark~M Churchland.
\newblock Cortical control of arm movements: a dynamical systems perspective.
\newblock \emph{Annual review of neuroscience}, 36\penalty0 (1):\penalty0
  337--359, 2013.

\bibitem[Vaswani et~al.(2017)Vaswani, Shazeer, Parmar, Uszkoreit, Jones, Gomez,
  ukasz Kaiser, and Polosukhin]{vaswani2017attention}
Ashish Vaswani, Noam Shazeer, Niki Parmar, Jakob Uszkoreit, Llion Jones,
  Aidan~N Gomez, {\L}~ukasz Kaiser, and Illia Polosukhin.
\newblock Attention is {{All}} you {{Need}}.
\newblock In \emph{Advances in {{Neural Information Processing Systems}}},
  volume~30. Curran Associates, Inc., 2017.

\bibitem[Devlin et~al.(2019)Devlin, Chang, Lee, and Toutanova]{devlin2019bert}
Jacob Devlin, Ming-Wei Chang, Kenton Lee, and Kristina Toutanova.
\newblock Bert: Pre-training of deep bidirectional transformers for language
  understanding.
\newblock \emph{Naacl}, 2019.

\bibitem[Shaw et~al.(2018)Shaw, Uszkoreit, and Vaswani]{shaw2018selfattention}
Peter Shaw, Jakob Uszkoreit, and Ashish Vaswani.
\newblock Self-attention with relative position representations.
\newblock \emph{arXiv preprint arXiv:1803.02155}, 2018.

\bibitem[Su et~al.(2023)Su, Lu, Pan, Murtadha, Wen, and Liu]{su2023roformer}
Jianlin Su, Yu~Lu, Shengfeng Pan, Ahmed Murtadha, Bo~Wen, and Yunfeng Liu.
\newblock {RoFormer: Enhanced transformer with rotary position embedding}.
\newblock \emph{arXiv preprint arXiv:2104.09864}, 2023.

\bibitem[Virtanen et~al.(2020)Virtanen, Gommers, Oliphant, Haberland, Reddy,
  Cournapeau, Burovski, Peterson, Weckesser, Bright, {van der Walt}, Brett,
  Wilson, Millman, Mayorov, Nelson, Jones, Kern, Larson, Carey, Polat, Feng,
  Moore, {VanderPlas}, Laxalde, Perktold, Cimrman, Henriksen, Quintero, Harris,
  Archibald, Ribeiro, Pedregosa, {van Mulbregt}, and {SciPy 1.0
  Contributors}]{2020SciPy-NMeth}
Pauli Virtanen, Ralf Gommers, Travis~E. Oliphant, Matt Haberland, Tyler Reddy,
  David Cournapeau, Evgeni Burovski, Pearu Peterson, Warren Weckesser, Jonathan
  Bright, St{\'e}fan~J. {van der Walt}, Matthew Brett, Joshua Wilson, K.~Jarrod
  Millman, Nikolay Mayorov, Andrew R.~J. Nelson, Eric Jones, Robert Kern, Eric
  Larson, C~J Carey, {\.I}lhan Polat, Yu~Feng, Eric~W. Moore, Jake
  {VanderPlas}, Denis Laxalde, Josef Perktold, Robert Cimrman, Ian Henriksen,
  E.~A. Quintero, Charles~R. Harris, Anne~M. Archibald, Ant{\^o}nio~H. Ribeiro,
  Fabian Pedregosa, Paul {van Mulbregt}, and {SciPy 1.0 Contributors}.
\newblock {{SciPy} 1.0: Fundamental Algorithms for Scientific Computing in
  Python}.
\newblock \emph{Nature Methods}, 17:\penalty0 261--272, 2020.
\newblock \doi{10.1038/s41592-019-0686-2}.

\end{thebibliography}

}

\clearpage
\appendix
\begin{center}
     {\Large \textbf{Appendix}}    
\end{center}

\setcounter{figure}{0}
\setcounter{table}{0}
\renewcommand{\thefigure}{S\arabic{figure}}
\renewcommand{\thetable}{S\arabic{table}}

\begin{figure}[H]
    \centering
    \includegraphics[width=1.0\linewidth]{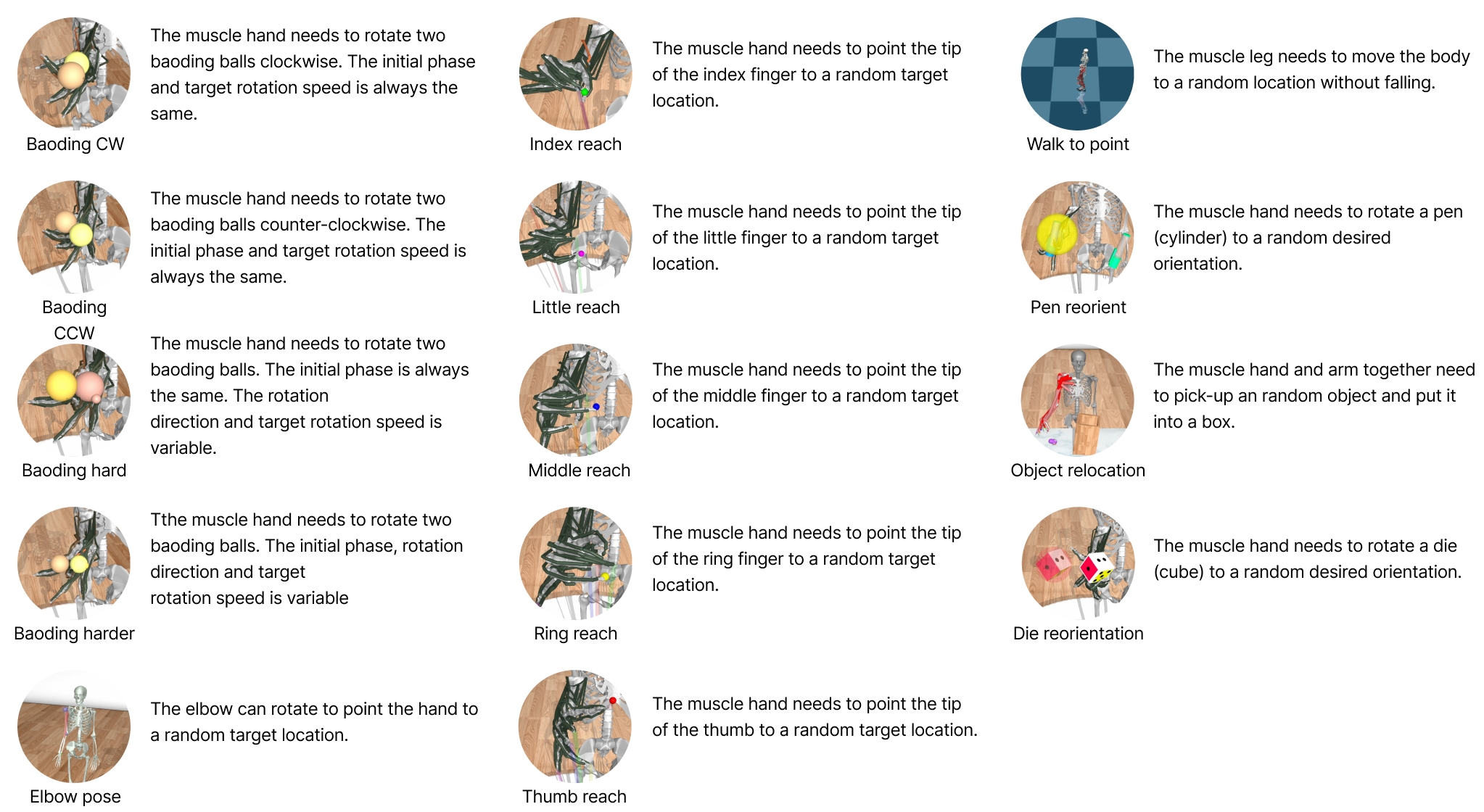}
    \caption{\textbf{Visualizations and descriptions of tasks.} Task descriptions of the 14 MyoSuite tasks that we consider for Arnold.}
    \label{fig:fulltask_description}
\end{figure}

\begin{figure}[h]
    \centering
    \includegraphics[width=1\linewidth]{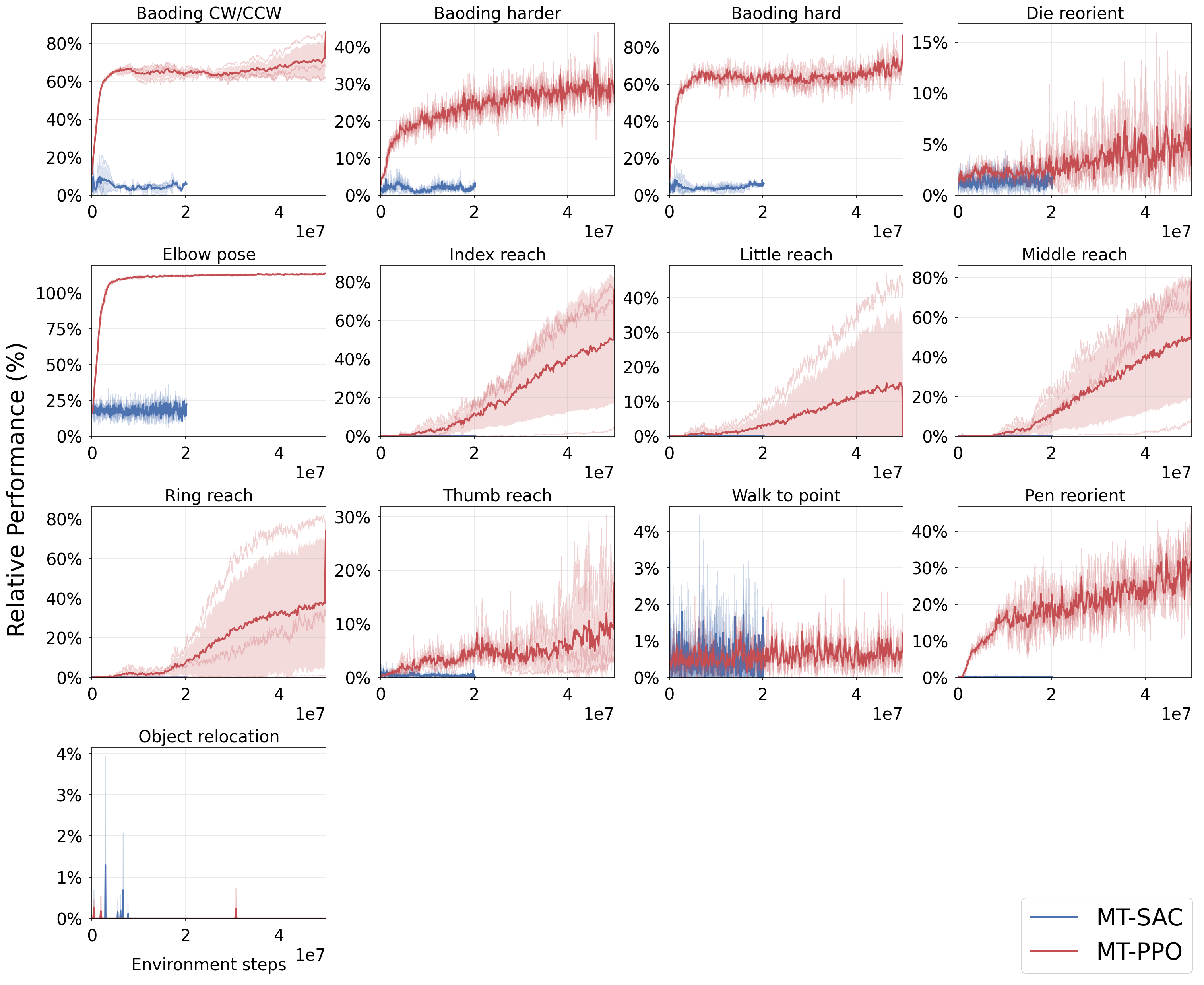}
    \caption{\textbf{Multi-task reinforcement learning with task-aware MLP on combined observation and action space.} Per-task \emph{relative performance} of Multi-task PPO and Multi-task SAC compared to the respective expert policy, plotted across 50M training steps. MT-SAC is stopped early due to the poor performance. Shaded area denotes the standard deviation over 3 training seeds, lighter curve denotes each seed, and the dark curve is the mean over the 3 training seeds. The Baoding CW and Baoding CCW tasks are evaluated jointly during training, and therefore share the same learning curve. All the performances are measured with the \emph{relative performance} defined in the \emph{Problem formulation} section. MT-PPO showed promising improvements despite the limited peak performance on manipulation tasks, while MT-SAC failed to improve beyond the initial performance.}
    \label{fig:mt_ppo_curves}
\end{figure}

\begin{figure}[ht!]
    \centering
    \includegraphics[width=0.95\linewidth]{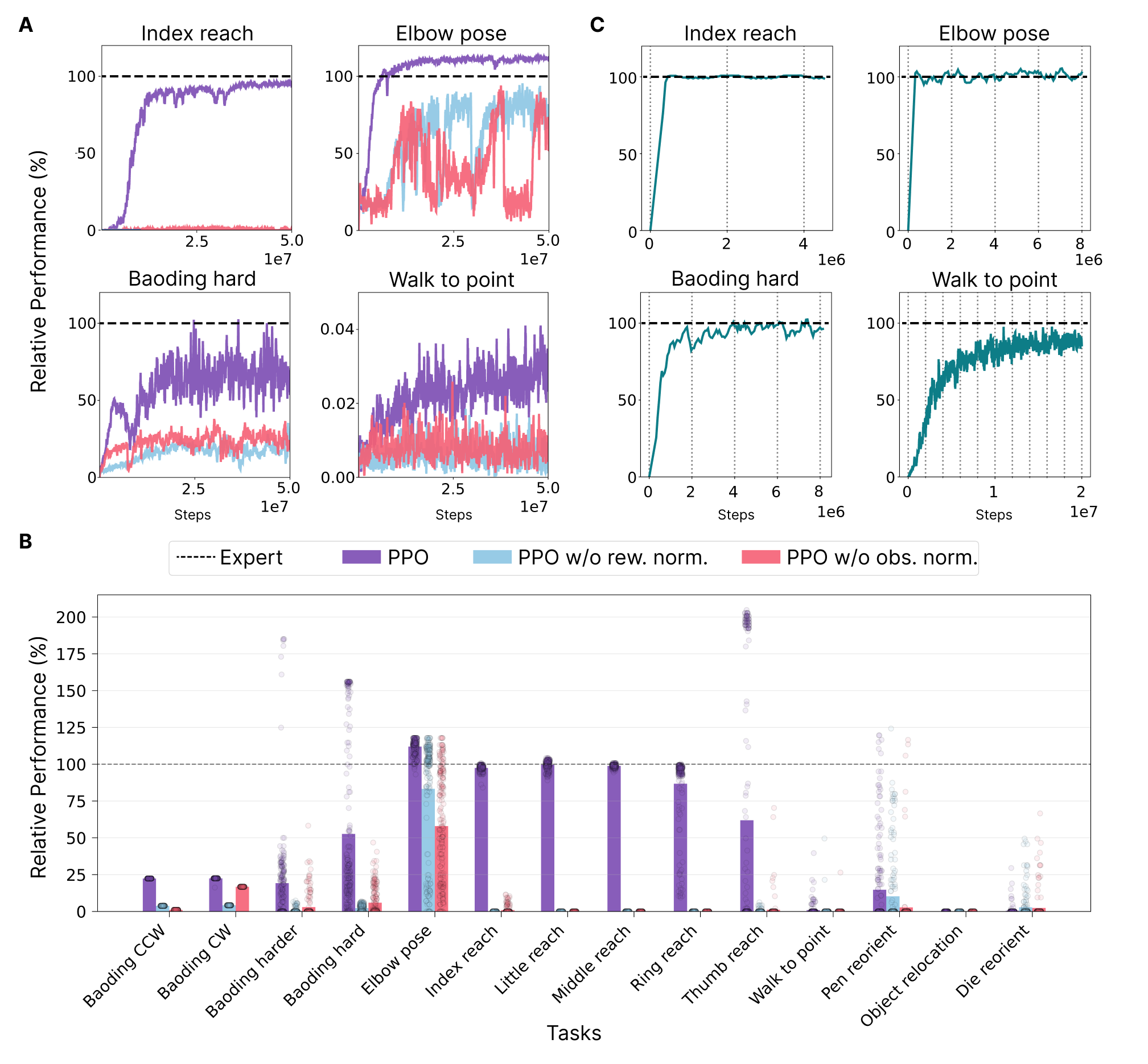}
    \caption{\textbf{Multi-task from scratch and single-task imitation learning performance.} \textbf{A:} Example learning curves of a muscle transformer policy trained on 14 tasks using PPO (purple). Reward standardization (rew. norm., blue) and per-component observation standardization (obs norm, pink) are ablated as variants of PPO. All training curves are shown in Figure \ref{fig:muscle_transfomer_ppo_curves}. \textbf{B:} Per-task performance comparison of muscle transformer-based PPO and the two standardization ablations. Individual points correspond to the \emph{relative performance} of different episodes; bars denote the mean across 200 episodes. Several tasks have a binary nature where the episode is solved either quickly or not at all; this leads to strongly bimodal distributions, such as the one shown in Thumb reach. \textbf{C:} Example learning curves of a muscle transformer policy trained on single-task imitation learning with BC. The student policy successfully converges to expert-level performance across all tasks, with the number of required training steps reflecting the complexity of the individual tasks. All training curves are shown in Figure \ref{fig:single_task_lc}.}
    \label{fig:ablations}
\end{figure}

\begin{figure}[h]
    \centering
    \includegraphics[width=1\linewidth]{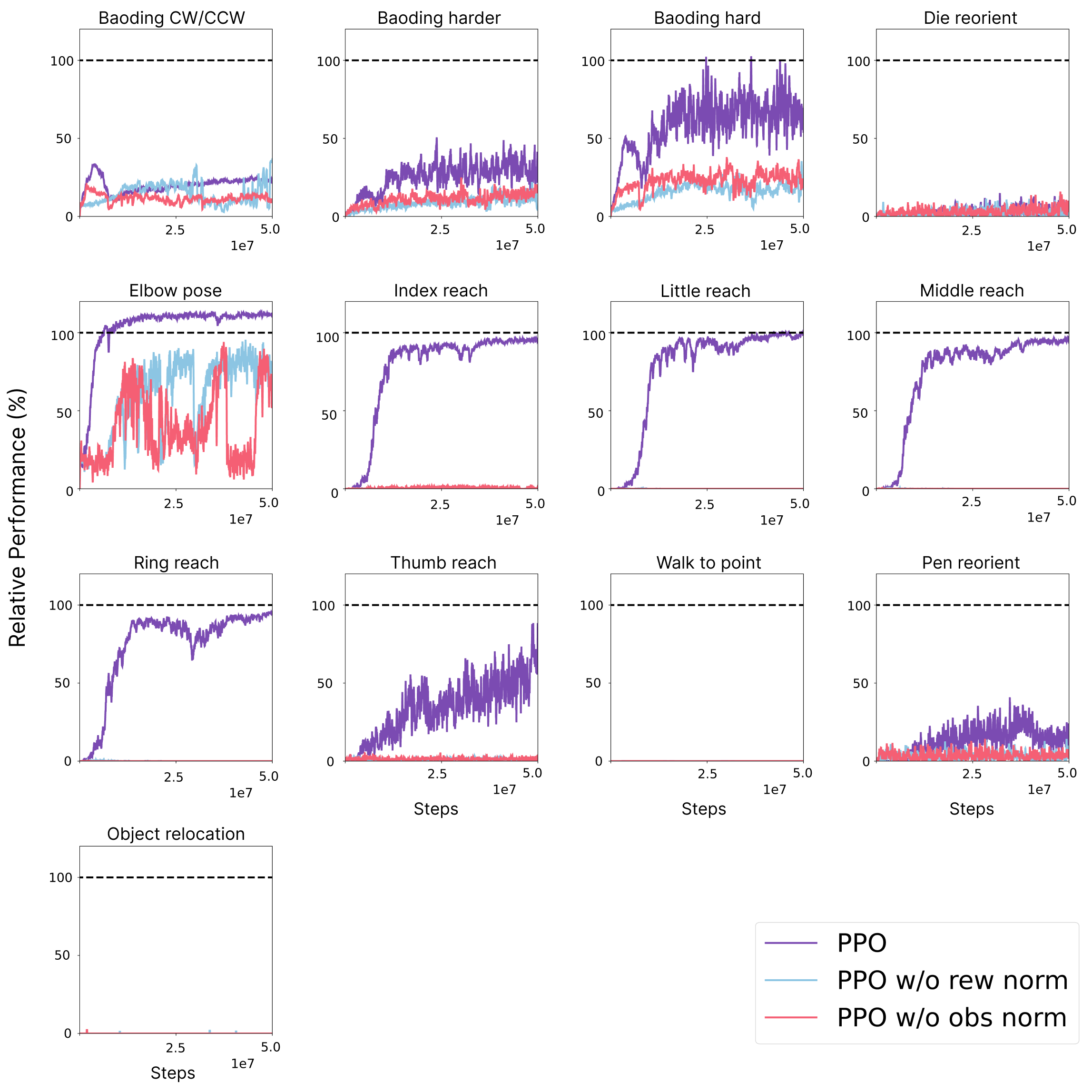}
    \caption{\textbf{Multi-task learning with the muscle transformer} Per-task \emph{relative performance} compared to the respective expert policy, plotted across 50M training steps. The Baoding CW and Baoding CCW tasks are evaluated jointly during training, and therefore share the same learning curve. All the performances are measured with \emph{relative performance} defined in the \emph{Problem formulation} section.}
    \label{fig:muscle_transfomer_ppo_curves}
\end{figure}

\begin{figure}[ht!]
    \centering
    \includegraphics[width=\linewidth]{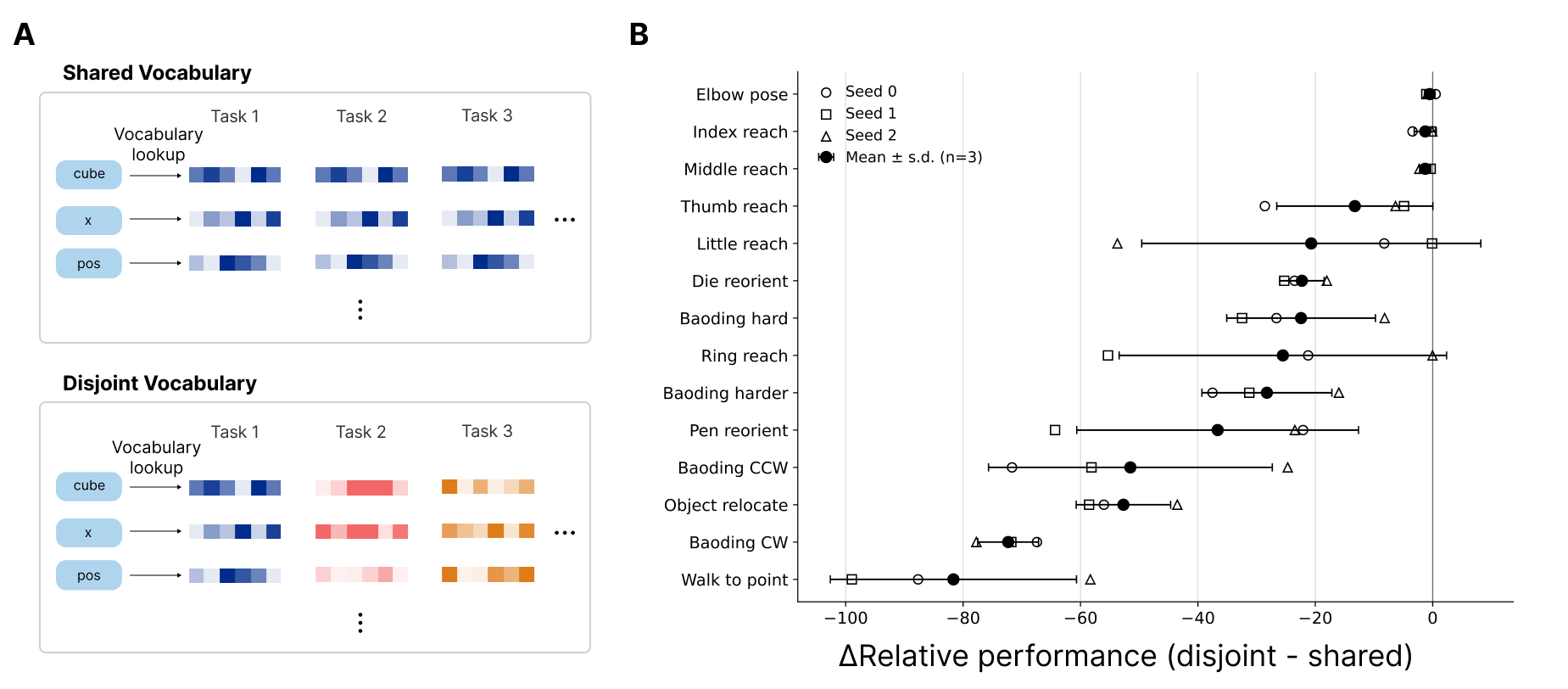}
    \caption{\textbf{Sensorimotor vocabulary explanation and analysis.} \textbf{A:} Schematic of the vocabulary-sharing ablation. In the original shared vocabulary, semantically identical tokens are associated with the same embedding across tasks. In the ablated task-specific (disjoint) vocabulary, each task maintains its own token embeddings, removing cross-task sharing while preserving the same token identities. \textbf{B:} Per-task performance change of the task-specific vocabulary model relative to the original shared-vocabulary model after OBC training (3 seeds). Symbols denote individual seeds; error bars indicate mean $\pm$ s.d. across seeds.}
    \label{fig:overview_with_vocabulary}
\end{figure}

\begin{figure}[ht!]
    \centering
    \includegraphics[width=0.7\linewidth]{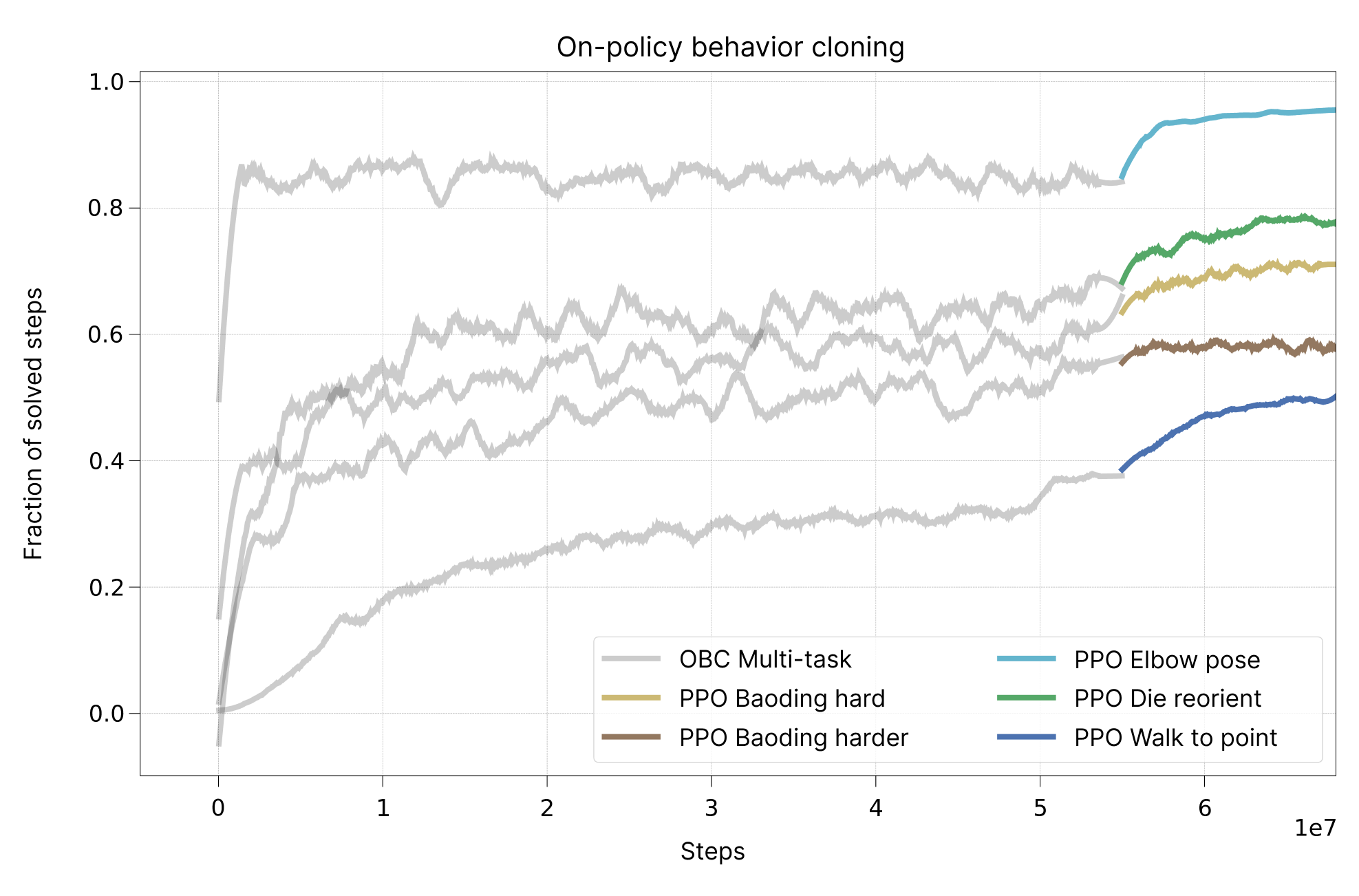}
    \caption{\textbf{Learning curves of Arnold.} Learning curves of multi-task OBC (50M steps plus 5M steps at reduced learning rate) followed by RL fine-tuning on single tasks. The performance is relative to those tasks where RL fine-tuning leads to a performance improvement. The performance is measured as \emph{solved fraction}, different from the metric \emph{relative performance} used in other figures. The policies resulting from the fine-tuning experiments were used as new experts for Arnold.}
    \label{fig:rl_fine_tuning}
\end{figure}

\begin{figure}[h]
    \centering
    \includegraphics[width=1\linewidth]{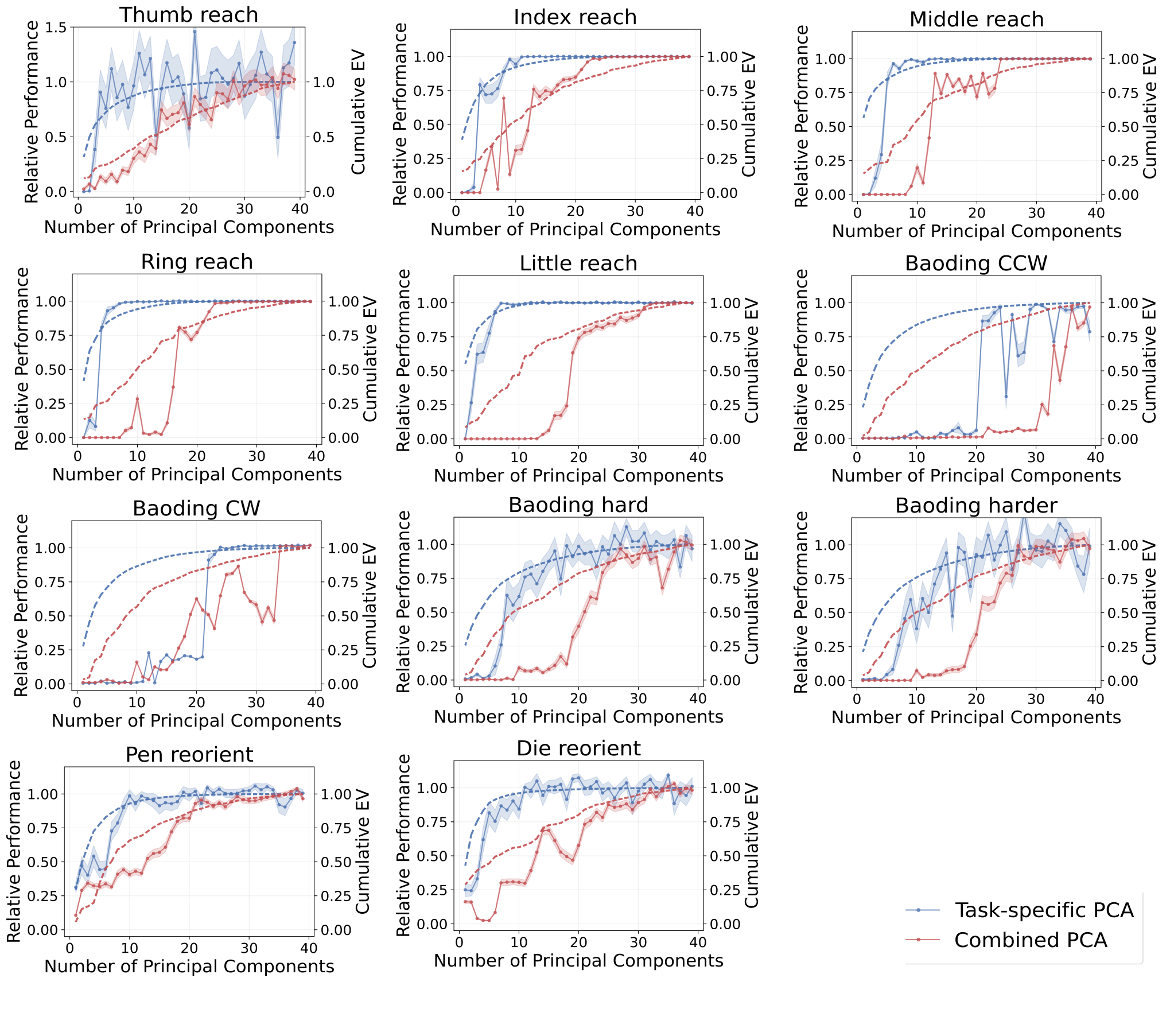}
    \caption{\textbf{Control space dimensionality of tasks (PCA).} Comparison between CSI and cumulative explained variance (EV) as measures of control space dimensionality. CSI results are shown in solid lines for task-specific (blue) and task-shared (red) principal components, and EV results are respectively shown in dashed lines. All the relative performances are measured with \emph{relative performance} defined in the \emph{Problem formulation} section.}
    \label{fig:ev_v_csi}
\end{figure}

\begin{figure}[h]
    \centering
    \includegraphics[width=1\linewidth]{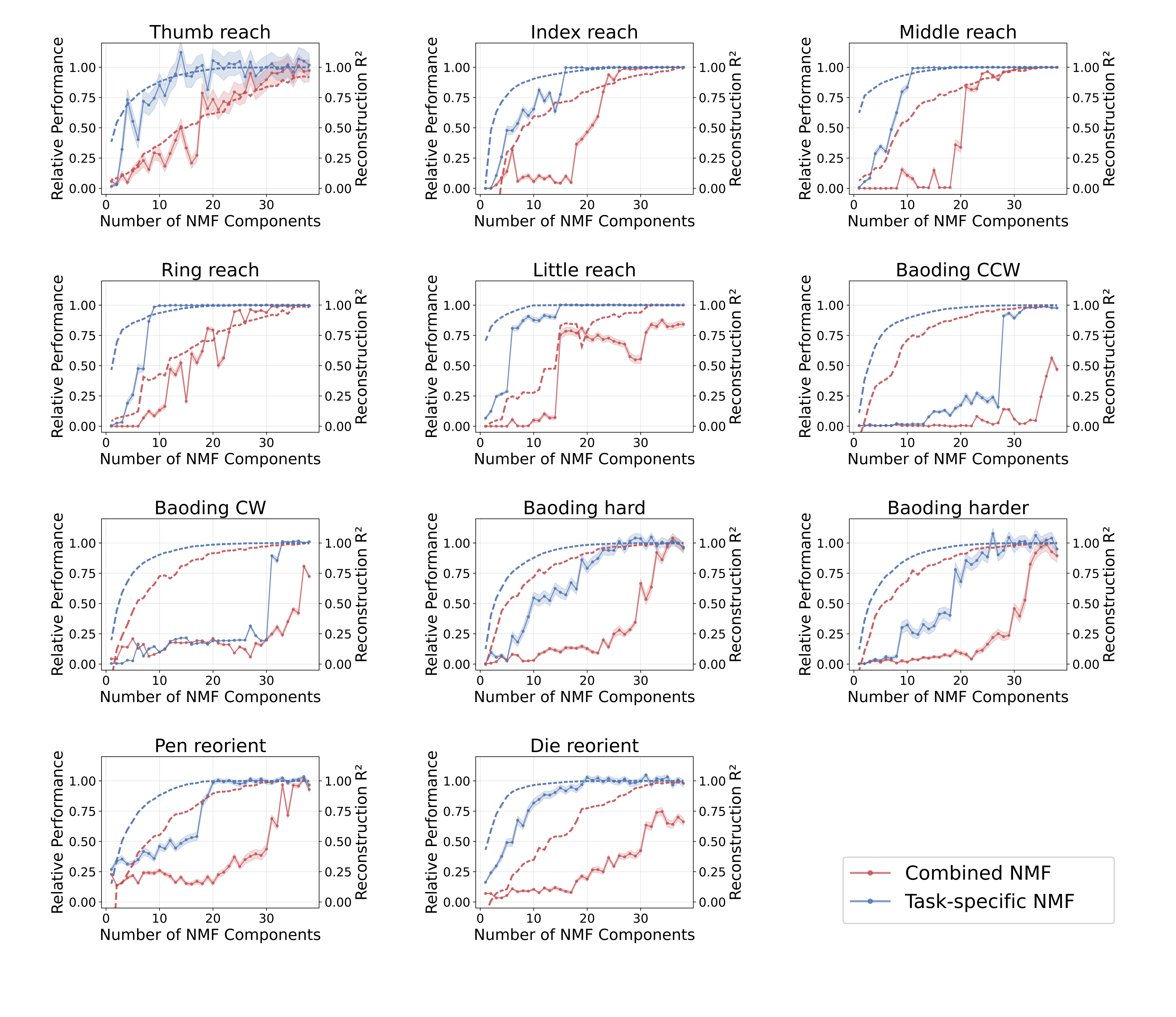}
    \caption{\textbf{Control space dimensionality of tasks (NMF).} Comparison between CSI and cumulative reconstruction ($\text{R}^2$) as measures of control space dimensionality. CSI results are shown in solid lines for task-specific (blue) and task-shared (red) NMF components, and $\text{R}^2$ results are respectively shown in dashed lines. All the performances are measured with \emph{relative performance} defined in the \emph{Problem formulation} section.}
    \label{fig:ev_v_csi_nmf}
\end{figure}

\begin{figure}[h]
    \centering
    \includegraphics[width=1\linewidth]{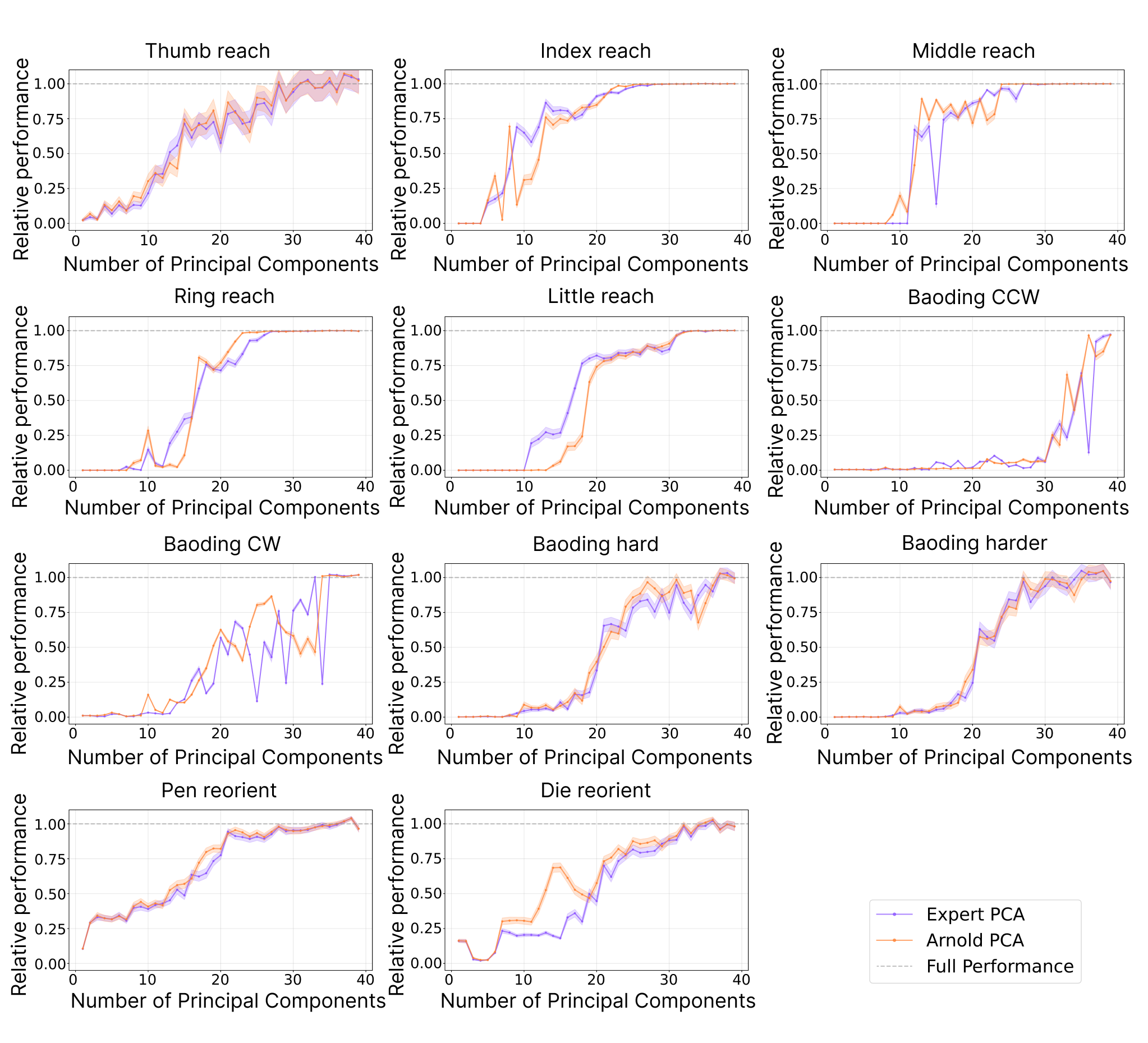}
    \caption{\textbf{Control space dimensionality of experts and Arnold.}  Comparison between the performance for varying action space dimensionality, when the control subspace is defined by principal components extracted from actions generated by Arnold (orange) or single-task specialists (violet), with actions from all tasks combined. All the performances are measured with \emph{relative performance} defined in the \emph{Problem formulation} section.}
    \label{fig:arnold_v_expert}
\end{figure}

\begin{figure}
    \centering
    \includegraphics[width=1\linewidth]{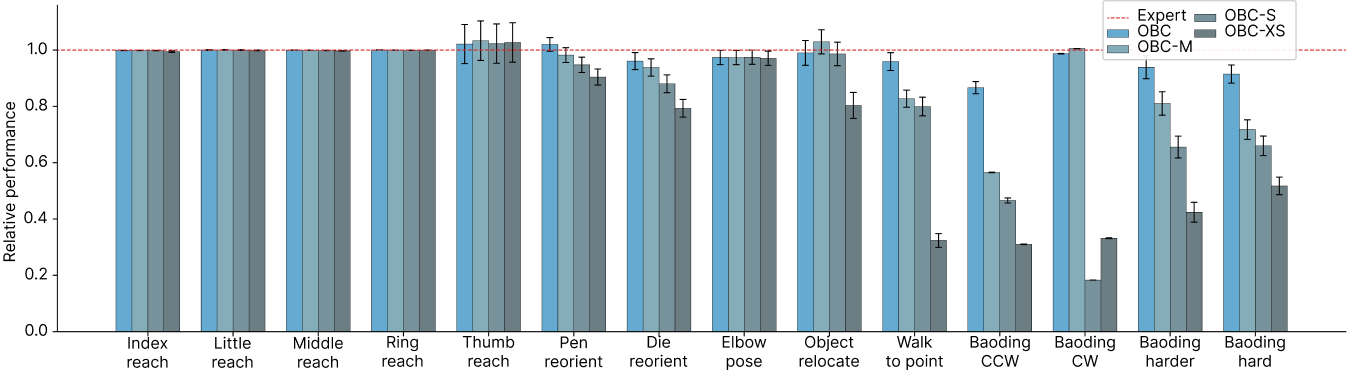}
    \caption{\textbf{Performance comparison across network size.} OBC network capacity variants compared across 14 benchmark tasks (200 episodes each), normalized by per-task expert performance (dashed red line = 1.0) using \emph{relative performance}. Error bars show $\pm$1 SEM.}
    \label{fig:capacity_full_tasks}
\end{figure}

\begin{figure}[h]
    \centering
    \includegraphics[width=1\linewidth]{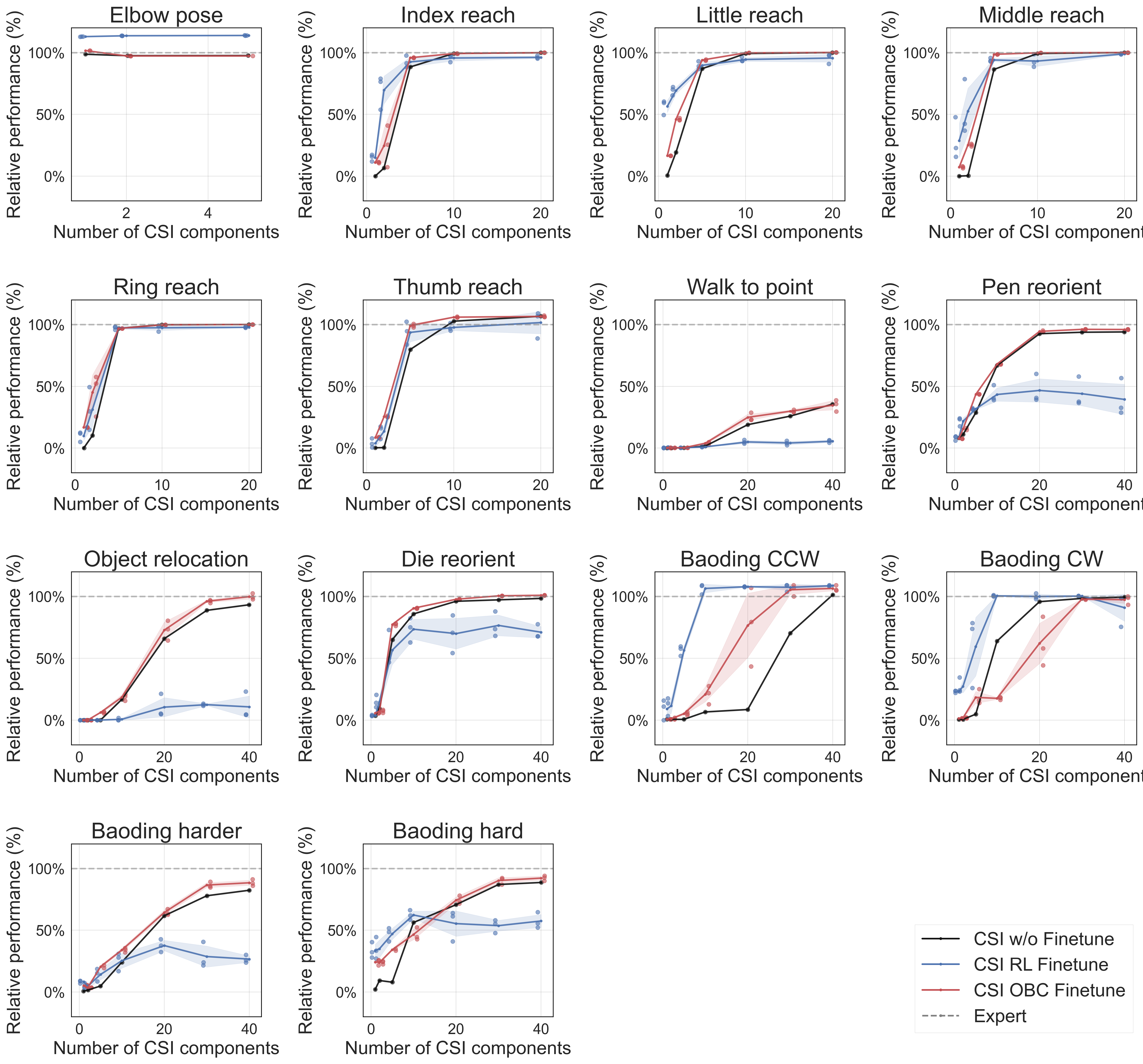}
    \caption{\textbf{Fine-tuning based CSI analysis.} Comparison between fine-tuned CSI policy and frozen CSI policy. MLP policies are trained with OBC for 5M steps on single tasks, and the action space is subsequently constrained to a specific dimensionality. The constrained policies are then fine-tuned with PPO (or OBC) for 5M steps with the dimensionality constraint applied. All experiments are run for 3 different seeds to get the standard deviation of the curves (shaded area), the scattered dots are the performance for the 3 different seeds. All the performances are measured with \emph{relative performance} defined in the \emph{Problem formulation} section.}
    \label{fig:csi_all_tasks}
\end{figure}

\begin{figure}[ht!]
    \centering
    \includegraphics[width=.8\linewidth]{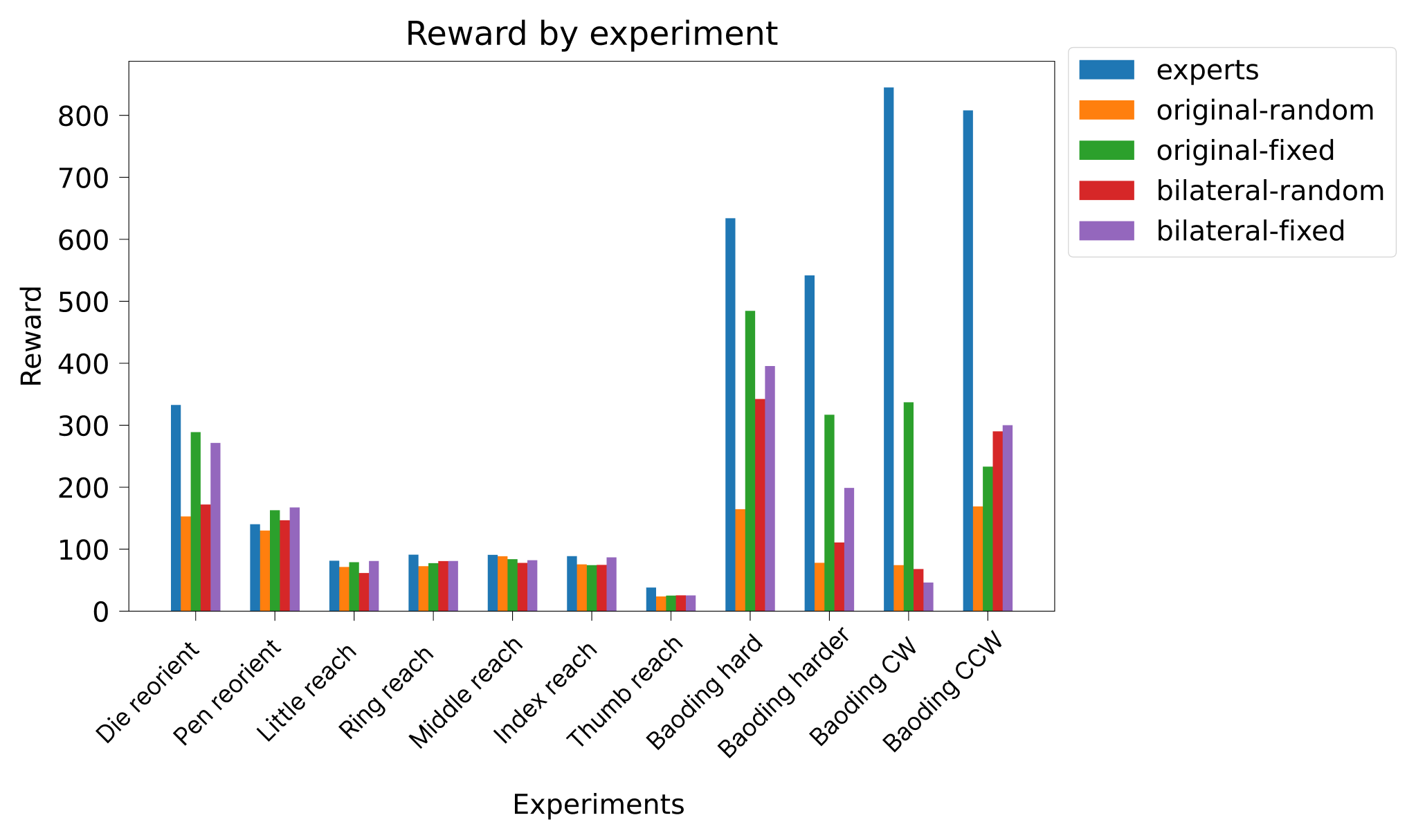}
    \caption{\textbf{Ablation on observation time horizon.} Experiment conducted to compare the performance of original model and a modified model which runs on longer history through self-attention on the time axis. \texttt{original-random} is training our original method on a randomized expert, \texttt{original-fixed} is instead trained on a deterministic expert. \texttt{bilateral-random} and \texttt{bilateral-fixed} is the same experiments done with the modified transformer.}
    \label{fig:bilateral}
\end{figure}

\begin{figure}[ht]
    \centering
    \includegraphics[width=\linewidth]{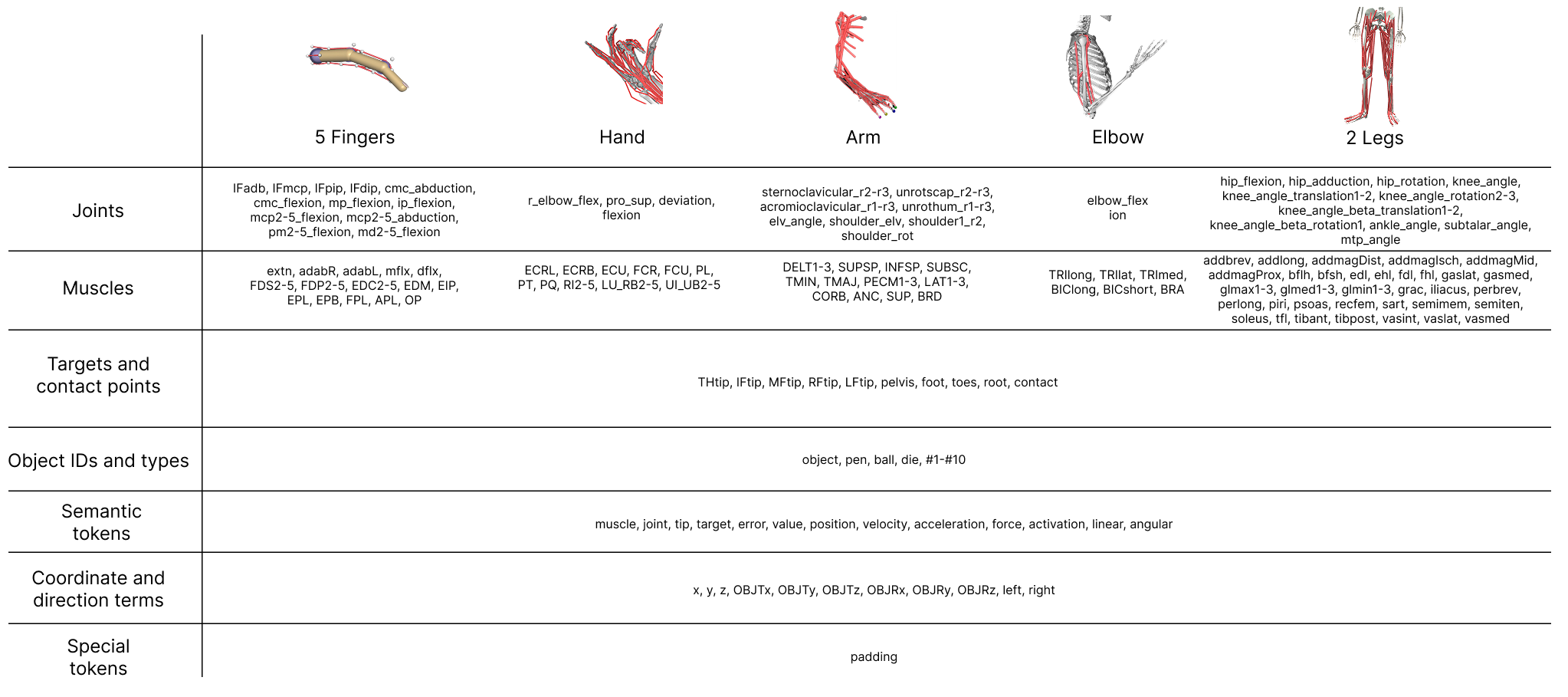}
    \caption{\textbf{List of \textit{sensorimotor vocabulary}.} All input sensory modalities are encoded with a sequence of tokens from the \emph{sensorimotor vocabulary}. In this figure, tokens that belong to fingers are separated from those of the hand due to space limitations.}
    \label{fig:vocabulary_explained}
\end{figure}

\begin{figure}[h]
    \centering
    \includegraphics[width=1\linewidth]{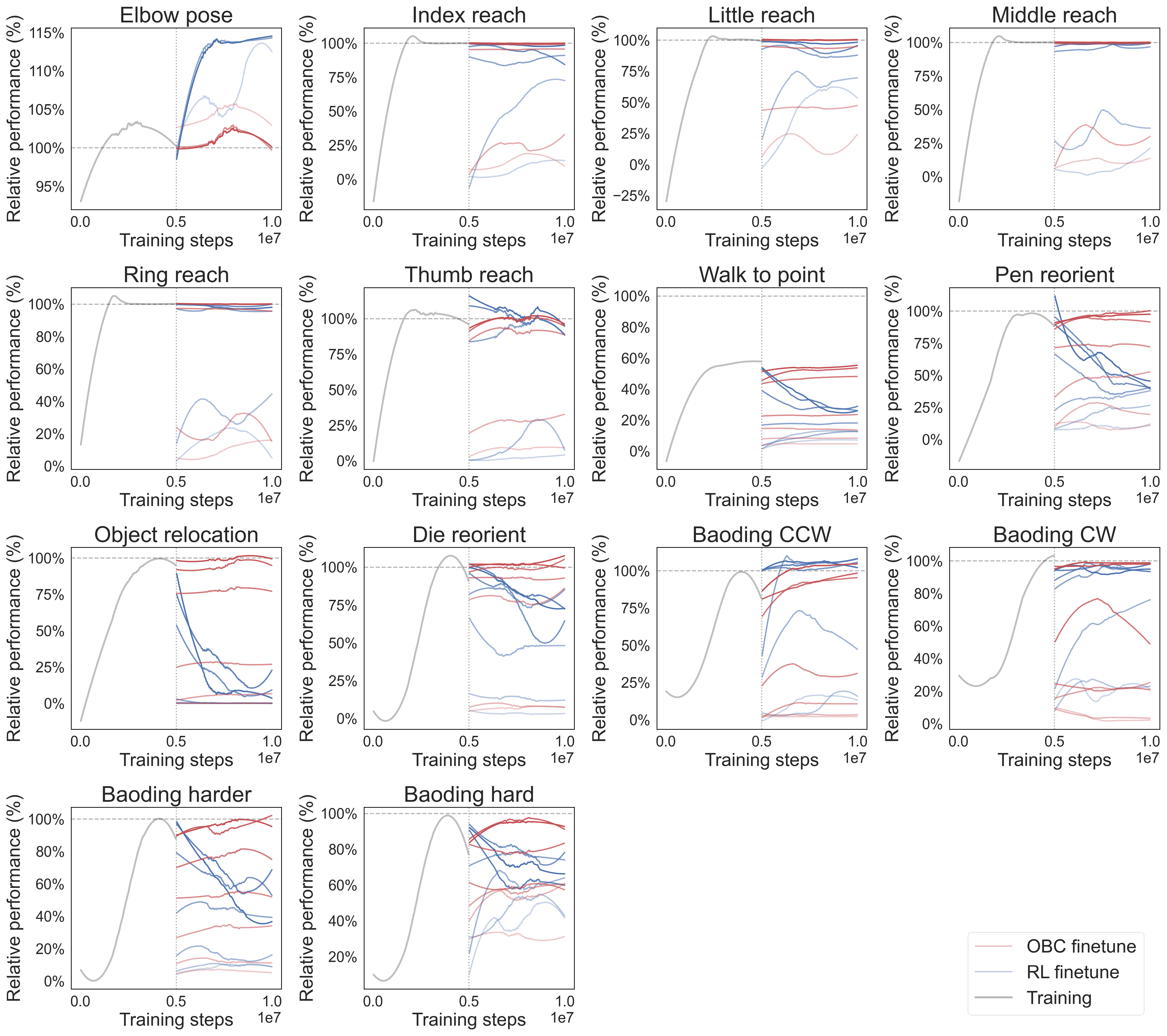}
    \caption{\textbf{Learning curves for fine-tuning based CSI analysis.} Learning curves of CSI analysis experiments. MLP policies are trained with OBC for 5M steps on single tasks, and then the action space is constrained to a specific dimensionality and continued training on the same task for 5M steps with either OBC or RL. Lighter color means lower dimensionality of action space. All the performances are measured with \emph{relative performance} defined in the \emph{Problem formulation} section.}
    \label{fig:csi_learning_curves}
\end{figure}

\begin{figure}[h]
    \centering
    \includegraphics[width=\linewidth]{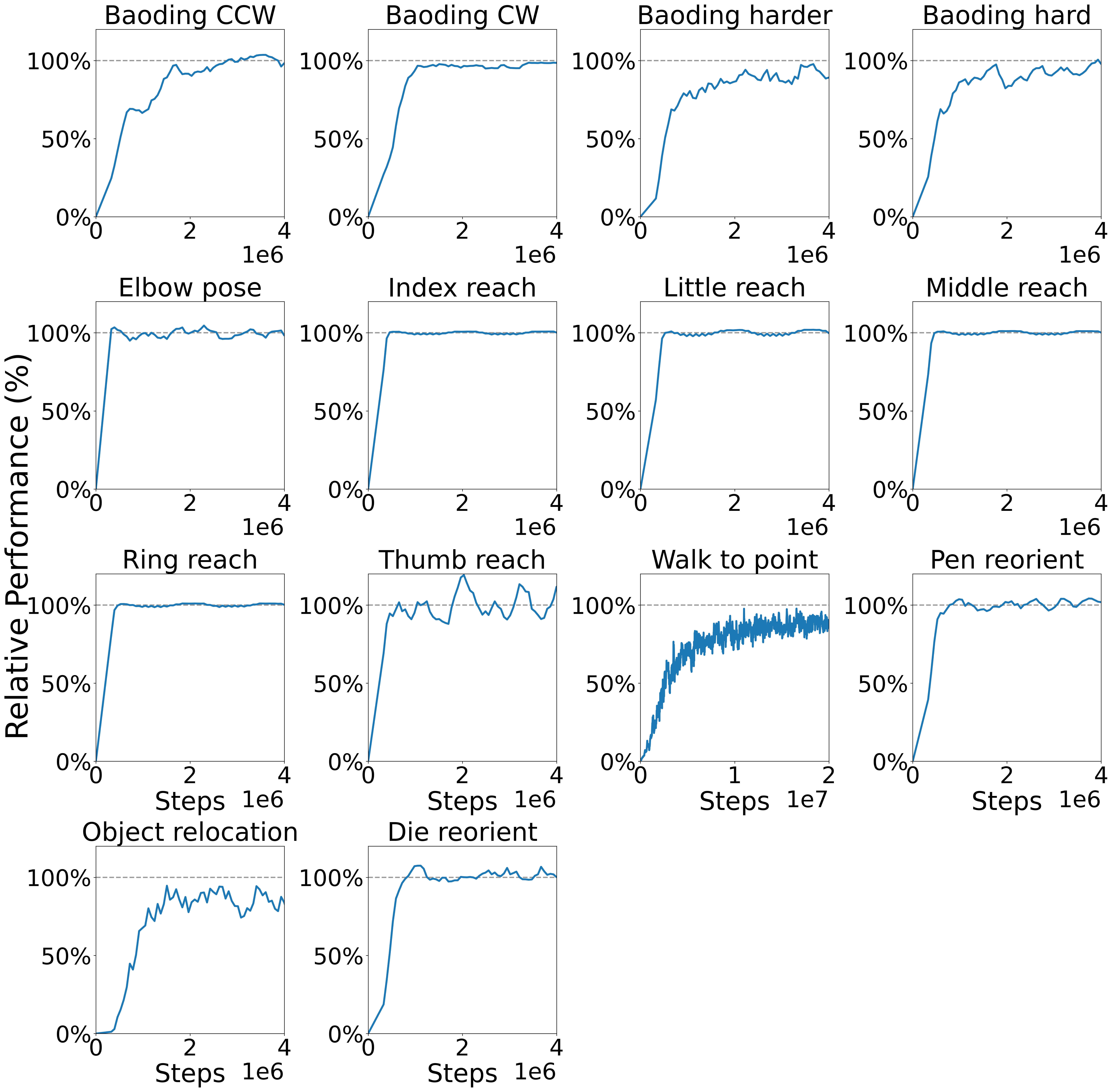}
    \caption{\textbf{Single-task imitation learning curves} We trained Arnold's policy network with BC on single tasks. As can be seen, certain tasks require longer training than the others to reach the expert's performance. We used these graphs as a metric to allocate more parallel environments to tasks that required more training steps in a single-task setting.}
    \label{fig:single_task_lc}
\end{figure}

\begin{sidewaystable}[p]
\centering\small\setlength{\tabcolsep}{3pt}\renewcommand{\arraystretch}{1.2}
\resizebox{\textheight}{!}{%
\begin{tabular}{l cccccccccccccc}
\toprule
\textbf{Method} & {Index reach} & {Little reach} & {Middle reach} & {Ring reach} & {Thumb reach} & {Elbow pose} & {Baoding CW} & {Baoding CCW} & {Baoding harder} & {Baoding hard} & {Die reorient} & {Pen reorient} & {Object relocation} & {Walk to point} \\
\midrule
MT-PPO$^{*}$ (3 seeds) & $57.8 \pm 17.4$ & $18.4 \pm 18.4$ & $47.9 \pm 16.5$ & $41.6 \pm 22.4$ & $9.1 \pm 4.8$ & $113.5 \pm 0.1$ & $61.2 \pm 4.0$ & $19.2 \pm 3.0$ & $5.6 \pm 0.4$ & $29.1 \pm 2.0$ & $2.4 \pm 0.4$ & $26.5 \pm 4.1$ & $0.0 \pm 0.0$ & $0.4 \pm 0.2$ \\
MT-SAC$^{*}$ (3 seeds) & $0.0 \pm 0.0$ & $0.0 \pm 0.0$ & $0.0 \pm 0.0$ & $0.0 \pm 0.0$ & $0.3 \pm 0.0$ & $20.1 \pm 3.6$ & $6.4 \pm 3.6$ & $6.7 \pm 3.7$ & $1.8 \pm 0.5$ & $3.5 \pm 1.9$ & $0.8 \pm 0.0$ & $0.0 \pm 0.0$ & $0.0 \pm 0.0$ & $0.3 \pm 0.2$ \\
PPO + T $^{*}$  (3 seeds) & $4.2 \pm 4.2$ & $0.0 \pm 0.0$ & $0.0 \pm 0.0$ & $0.0 \pm 0.0$ & $5.3 \pm 4.6$ & $18.3 \pm 1.2$ & $6.0 \pm 4.9$ & $6.4 \pm 5.6$ & $4.7 \pm 4.6$ & $3.0 \pm 2.7$ & $3.0 \pm 1.9$ & $1.8 \pm 1.3$ & $0.0 \pm 0.0$ & $0.3 \pm 0.2$ \\
PPO + T + SV $^{*}$ (3 seeds)& $93.2 \pm 0.6$ & $94.2 \pm 1.9$ & $90.0 \pm 2.5$ & $91.8 \pm 1.0$ & $44.6 \pm 2.4$ & $110.1 \pm 0.6$ & $20.3 \pm 0.6$ & $25.9 \pm 3.8$ & $21.7 \pm 1.8$ & $49.1 \pm 9.7$ & $3.1 \pm 0.2$ & $15.6 \pm 2.8$ & $0.0 \pm 0.0$ & $0.5 \pm 0.1$ \\
BC (offline) $^{*}$ (3 seeds) & $99.7 \pm 0.1$ & $99.1 \pm 0.3$ & $99.6 \pm 0.3$ & $99.3 \pm 0.3$ & $92.1 \pm 2.5$ & $97.2 \pm 0.1$ & $61.9 \pm 2.8$ & $34.4 \pm 2.7$ & $82.0 \pm 1.7$ & $90.6 \pm 1.3$ & $84.2 \pm 7.1$ & $102.3 \pm 0.4$ & $86.1 \pm 1.7$ & $7.1 \pm 0.8$ \\
OBC + Task-SV $^{*}$  (3 seeds)& $98.6 \pm 1.1$ & $79.3 \pm 16.7$ & $98.7 \pm 0.6$ & $74.5 \pm 16.1$ & $89.1 \pm 7.9$ & $96.8 \pm 0.5$ & $27.3 \pm 3.0$ & $47.6 \pm 13.7$ & $67.7 \pm 6.0$ & $72.8 \pm 6.9$ & $75.9 \pm 2.1$ & $65.0 \pm 14.2$ & $46.9 \pm 4.6$ & $16.8 \pm 11.1$ \\
OBC w/o Obs. Norm.$^{*}$  (5 seeds)& $99.8 \pm 0.0$ & $99.6 \pm 0.3$ & $99.9 \pm 0.0$ & $99.8 \pm 0.1$ & $95.9 \pm 3.1$ & $97.6 \pm 0.5$ & $71.1 \pm 13.4$ & $49.0 \pm 5.4$ & $75.6 \pm 1.7$ & $78.4 \pm 2.0$ & $89.6 \pm 0.8$ & $97.6 \pm 1.4$ & $74.4 \pm 3.9$ & $63.0 \pm 6.0$ \\
OBC $^{*}$  (5 seeds) & $99.9 \pm 0.0$ & $100.0 \pm 0.0$ & $100.0 \pm 0.0$ & $100.0 \pm 0.0$ & $102.7 \pm 0.7$ & $97.3 \pm 0.0$ & $98.6 \pm 0.2$ & $98.1 \pm 3.6$ & $94.3 \pm 2.2$ & $93.9 \pm 1.3$ & $97.6 \pm 0.5$ & $102.5 \pm 0.4$ & $99.1 \pm 1.4$ & $91.4 \pm 3.4$ \\
OBC-PPO $^{*}$  (5 seeds)& $99.9 \pm 0.0$ & $100.0 \pm 0.0$ & $100.0 \pm 0.0$ & $100.0 \pm 0.0$ & $103.2 \pm 0.1$ & $97.3 \pm 0.1$ & $97.2 \pm 1.1$ & $66.7 \pm 14.7$ & $91.6 \pm 4.0$ & $87.7 \pm 4.8$ & $94.6 \pm 3.4$ & $100.2 \pm 1.3$ & $96.4 \pm 2.3$ & $89.2 \pm 2.0$ \\
\midrule
\textbf{Arnold} (5 seeds)& $99.9 \pm 0.0$ & $100.0 \pm 0.0$ & $100.0 \pm 0.0$ & $100.0 \pm 0.0$ & $103.3 \pm 0.4$ & $113.9 \pm 0.1$ & $98.9 \pm 0.4$ & $102.9 \pm 1.5$ & $106.1 \pm 0.8$ & $107.1 \pm 1.0$ & $110.1 \pm 0.3$ & $102.2 \pm 0.4$ & $99.6 \pm 1.1$ & $121.3 \pm 1.3$ \\
\bottomrule
\end{tabular}}
\caption{\textbf{Per-task ablation results (supplementary).}
Solved fraction relative to the single-task expert policies (\%), reported per
task as the mean $\pm$ s.e.m.\ across random seeds of each model's expert-
normalized average solved-step fraction. Columns are the 14 individual tasks
(no family aggregation). Asterisks ($^{*}$) on the method name denote a
statistically significant decrease in overall performance relative to the full
Arnold model (two-sided Wilcoxon signed-rank test over the $N=14$ tasks,
Holm--Bonferroni corrected across methods, $p<0.05$). MT-SAC training was terminated early (20M interactions) due to low performance.}
\label{tab:ablations_per_task}
\end{sidewaystable}

\begin{table}[h]
    \centering
    \begin{tabular}{lccccc}
    \toprule
    Task & Expert & Arnold & Improvement (\%) & $t$ & $p$ (Holm) \\
    \midrule
    Little reach & $0.921$ & $0.921 \pm 0.000$ & $+0.04 \pm 0.00$ & $12.78$ & $0.0012$ \\
    Index reach & $0.963$ & $0.962 \pm 0.000$ & $-0.10 \pm 0.01$ & $-19.803$ & $1.0000$ \\
    Middle reach & $0.953$ & $0.953 \pm 0.000$ & $+0.01 \pm 0.01$ & $1.129$ & $0.6816$ \\
    Ring reach & $0.956$ & $0.956 \pm 0.000$ & $+0.02 \pm 0.02$ & $1.271$ & $0.6816$ \\
    Thumb reach & $0.483$ & $0.499 \pm 0.002$ & $\mathbf{+3.30 \pm 0.39}$ & $8.424$ & $\mathbf{0.0054^{*}}$ \\
    Die reorient & $0.701$ & $0.772 \pm 0.002$ & $\mathbf{+10.08 \pm 0.33}$ & $30.907$ & $\mathbf{<0.0001^{*}}$ \\
    Pen reorient & $0.652$ & $0.666 \pm 0.003$ & $\mathbf{+2.20 \pm 0.44}$ & $5.023$ & $\mathbf{0.0258^{*}}$ \\
    Baoding CW & $0.955$ & $0.944 \pm 0.003$ & $-1.15 \pm 0.35$ & $-3.263$ & $1.0000$ \\
    Baoding CCW & $0.915$ & $0.942 \pm 0.014$ & $+2.91 \pm 1.50$ & $1.941$ & $0.3727$ \\
    Baoding hard & $0.641$ & $0.687 \pm 0.006$ & $\mathbf{+7.14 \pm 0.96}$ & $7.463$ & $\mathbf{0.0069^{*}}$ \\
    Baoding harder & $0.541$ & $0.574 \pm 0.004$ & $\mathbf{+6.14 \pm 0.79}$ & $7.746$ & $\mathbf{0.0067^{*}}$ \\
    Elbow pose & $0.849$ & $0.967 \pm 0.001$ & $\mathbf{+13.90 \pm 0.12}$ & $111.254$ & $\mathbf{<0.0001^{*}}$ \\
    Object relocation & $0.339$ & $0.337 \pm 0.004$ & $-0.42 \pm 1.06$ & $-0.398$ & $1.0000$ \\
    Walk to point & $0.404$ & $0.489 \pm 0.005$ & $\mathbf{+21.29 \pm 1.34}$ & $15.854$ & $\mathbf{0.0006^{*}}$ \\
    \bottomrule
    \end{tabular}
    \caption{\textbf{Significance test for comparing Arnold with single task expert policies.} Per-task comparison between Arnold and the corresponding single-task experts. For each of the 14 tasks, we evaluated Arnold over $n=5$ independent seeds. We report absolute solved step fraction as the performance metric (columns `Expert' and `Arnold'). We then tested whether Arnold's distribution of \emph{relative performance} improved over the single task experts. The $p$ value corresponds to a parametric one-sided one-sample $t$-test (df = 4) of the null hypothesis that the mean improvement is $\leq 0$; the 14 p-values are Holm–Bonferroni corrected. Little reach was statistically significant but is practically not considered due to the low relative improvement.}
    \label{tab:arnold-expert-significance}
\end{table}

\begin{table}[h]
\centering
\caption{Architecture comparison of OBC model variants.}
\label{tab:obc-architectures}
\begin{tabular}{lrrrr}
\toprule
\textbf{Property} & \textbf{OBC} & \textbf{OBC-M} & \textbf{OBC-S} & \textbf{OBC-XS} \\
\midrule
\texttt{embedding\_size}   & \textbf{128} & 64  & 32  & 16  \\
\texttt{num\_layers}       & \textbf{6}   & 3   & 3   & 3   \\
\texttt{num\_heads}        & \textbf{4}   & 4   & 2   & 2   \\
\texttt{dim\_feedforward}  & \textbf{512} & 512 & 512 & 256 \\
\midrule
\textbf{Total parameters}  & \textbf{4{,}450{,}564} & 890{,}692 & 386{,}212 & 104{,}020 \\
Relative to OBC       & 1.00$\times$ & 0.20$\times$ & 0.087$\times$ & 0.023$\times$ \\
\bottomrule
\end{tabular}
\end{table}

\begin{table}[h]
    \centering
    \begin{tabular}{lc}
        \toprule
        Environment Name & Count \\
        \midrule
        Elbow pose & 2 \\
        Thumb reach & 2 \\
        Index reach & 2 \\
        Middle reach & 2 \\
        Ring reach & 2 \\
        Little reach & 2 \\
        Pen reorient & 2 \\
        Die reorient & 2 \\
        Baoding CW & 4 \\
        Baoding CCW & 4 \\
        Baoding hard & 6 \\
        Baoding harder & 6 \\
        Object relocation & 6 \\
        Walk to point & 12 \\
        \bottomrule
    \end{tabular}
    \caption{\textbf{Number of environments.} Number of environments per task used in Arnold training. We picked these heuristic numbers based on the single-task imitation experiments (Figure~\ref{fig:single_task_lc}).}
    \label{tab:env_counts}
\end{table}

\begin{table}[t]
\centering
\begin{tabular}{lccc}
\toprule
Method & $r_{\mathrm{rb}}$ & $W$ & $p_{\mathrm{Holm}}$ \\
\midrule
MT-SAC           & $+1.000$ & $0$  & $1.1\times10^{-3}$$^{\dagger}$ \\
MT-PPO           & $+1.000$ & $0$  & $1.1\times10^{-3}$$^{\dagger}$ \\
MT-PPO+T         & $+1.000$ & $0$  & $1.1\times10^{-3}$$^{\dagger}$ \\
MT-PPO+T+SV      & $+1.000$ & $0$  & $1.1\times10^{-3}$$^{\dagger}$ \\
BC               & $+0.981$ & $1$  & $1.1\times10^{-3}$ \\
OBC + Task-SV    & $+1.000$ & $0$  & $1.1\times10^{-3}$$^{\dagger}$ \\
OBC w/o Obs.Norm & $+1.000$ & $0$  & $1.1\times10^{-3}$$^{\dagger}$ \\
OBC-PPO          & $+0.981$ & $1$  & $1.1\times10^{-3}$ \\
OBC              & $+0.790$ & $11$ & $6.7\times10^{-3}$ \\
\bottomrule
\end{tabular}
\caption{Arnold vs.\ each ablated method. Matched-pairs rank-biserial effect
size $r_{\mathrm{rb}}$, Wilcoxon signed-rank statistic $W$, and Holm-corrected
two-sided $p$-value over the $N=14$ paired per-task scores. Positive
$r_{\mathrm{rb}}$ favours Arnold. $^{\dagger}$ marks rows at the exact discrete
floor of the test ($p_{\mathrm{raw}} = 2/2^{14} = 1.2\times10^{-4}$).}
\label{tab:ablation-significance}
\end{table}

\begin{table}[t]
\centering
\begin{tabular}{llcc}
\toprule
Metric & Comparison & $W$ & $p_{\mathrm{Holm}}$ \\
\midrule
\multirow{3}{*}{Mean $|\Delta a|$ $\downarrow$}
  & Expert vs.\ Single-task & $19$ & $0.240$ \\
  & Expert vs.\ OBC         & $0$  & $0.0029^{*\dagger}$ \\
  & Single-task vs.\ OBC    & $0$  & $0.0029^{*\dagger}$ \\
\midrule
\multirow{3}{*}{SPARC $\uparrow$}
  & Expert vs.\ Single-task & $31$ & $0.898$ \\
  & Expert vs.\ OBC         & $20$ & $0.557$ \\
  & Single-task vs.\ OBC    & $11$ & $0.161$ \\
\midrule
\multirow{3}{*}{LDLJ $\uparrow$}
  & Expert vs.\ Single-task & $16$ & $0.147$ \\
  & Expert vs.\ OBC         & $1$  & $0.0059^{*}$ \\
  & Single-task vs.\ OBC    & $1$  & $0.0059^{*}$ \\
\midrule
\multirow{3}{*}{NVP $\downarrow$}
  & Expert vs.\ Single-task & $7$  & $0.019^{*}$ \\
  & Expert vs.\ OBC         & $0$  & $0.0029^{*\dagger}$ \\
  & Single-task vs.\ OBC    & $0$  & $0.0029^{*\dagger}$ \\
\bottomrule
\end{tabular}
\caption{Pairwise smoothness comparisons between the per-task PPO expert,
single-task OBC, and multi-task OBC. Two-sided Wilcoxon signed-rank over the
$n=11$ paired per-task means, Holm-corrected across the three comparisons
within each metric. Arrows give the direction in which each metric indicates
greater smoothness. $^{*}$ marks $p_{\mathrm{Holm}} < 0.05$;
$^{\dagger}$ marks comparisons at the exact discrete floor of the test
($p_{\mathrm{raw}} = 2/2^{11} = 9.8\times10^{-4}$).}
\label{tab:smoothness-significance}
\end{table}

\clearpage

\fakesection{Videos}

\subsection*{Video of Arnold's performance}

To visually illustrate the diversity of the tasks and embodiments, we created two videos of Arnold's behavior across the different tasks. We note that number of successful and failed episodes does not reflect the average performance of Arnold on the tasks.

\begin{video}[ht!]
    \centering
    \includegraphics[width=.5\linewidth]{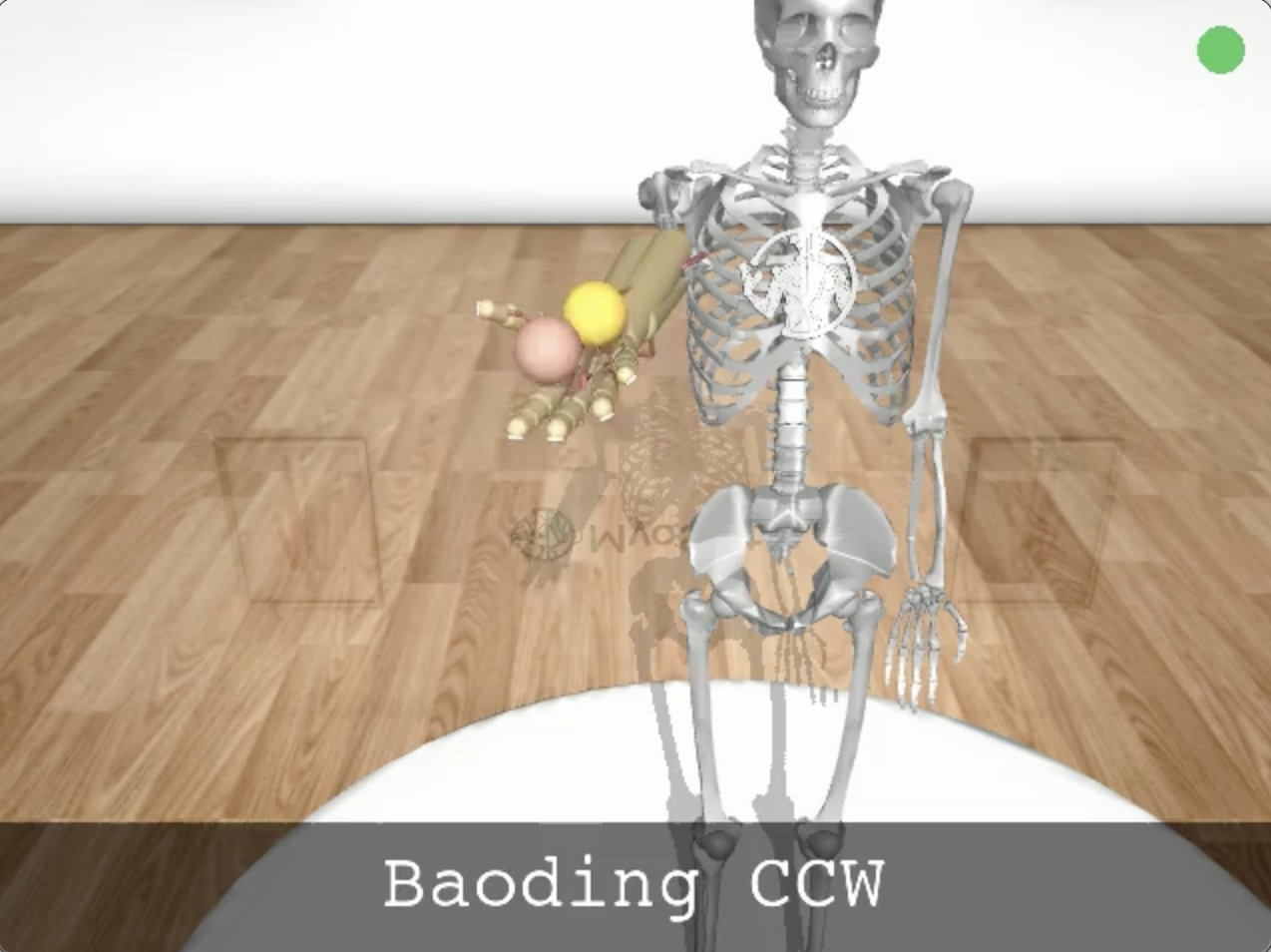}
    \caption{\textbf{Selected successful and unsuccessful trials of Arnold.} The linked video shows successful trials by Arnold on a few tasks. The green dot indicates successful and the red dot unsuccessful trials, respectively. URL:~\url{https://youtu.be/CPw9A0x767E} } %
    \label{fig:video_1}
\end{video}

\begin{video}[ht!]
    \centering
    \includegraphics[width=.5\linewidth]{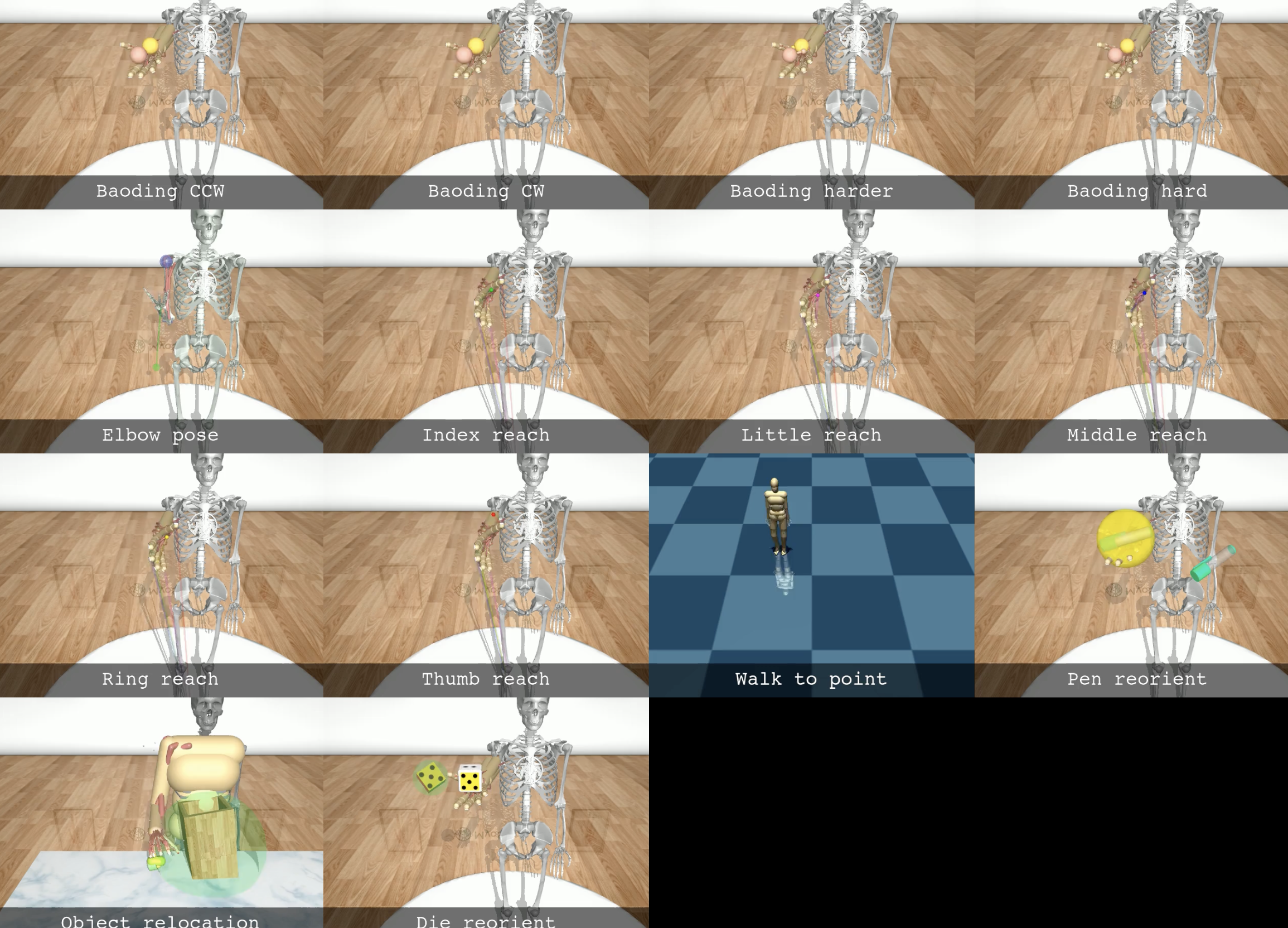}
    \caption{\textbf{Visual illustration of diversity of tasks and embodiments.} The linked video shows a visualization of Arnold carrying out 14 different tasks. In the grid movies, several successful episodes are shown per task. URL:~\url{https://youtu.be/MN3u83K6VC8} } %
    \label{fig:video_2}
\end{video}

\end{document}